%% file: main.tex
\documentclass[sigconf]{acmart}

\copyrightyear{2021}
\acmYear{2021} 
\setcopyright{acmcopyright}\acmConference[MM '21]{Proceedings of the 29th ACM International Conference on Multimedia}{October 20--24, 2021}{Virtual Event, China}
\acmBooktitle{Proceedings of the 29th ACM International Conference on Multimedia (MM '21), October 20--24, 2021, Virtual Event, China}
\acmPrice{15.00}
\acmDOI{10.1145/3474085.3475623}
\acmISBN{978-1-4503-8651-7/21/10}

\usepackage{graphicx}
\usepackage{amsmath}

\usepackage{listings}
\usepackage{array}
\usepackage{makecell}
\usepackage{tabularx}
\usepackage{multirow}
\usepackage{booktabs}
\usepackage{hyperref}
\newcommand{\etal}{\textit{et al.}}
\newcommand{\eg}{\textit{e.g.}}
\newcommand{\ie}{\textit{i.e.}}

\newcolumntype{Y}{>{\centering\arraybackslash}X}
\newcolumntype{P}{>{\centering\arraybackslash}p{.04\textwidth}}
\newcolumntype{Q}{>{\centering\arraybackslash}p{.02\textwidth}}
\newcolumntype{s}{>{\hsize=.3\hsize}X}
\newcolumntype{e}{>{\hsize=.9\hsize}X}
\newcolumntype{S}{>{\hsize=1.5\hsize}X}

\begin{document}
\fancyhead{}
\title{Attention-guided Temporally Coherent Video Object Matting}



\author{Yunke Zhang}
\affiliation{%
  \small
  \institution{Zhejiang University}
  \streetaddress{}
  \city{}
  \country{}
}
\email{yunkezhang@zju.edu.cn}

\author{Chi Wang}
\affiliation{%
  \small
  \institution{Zhejiang University}
  \streetaddress{}
  \city{}
  \country{}
}
\email{wangchi1995@zju.edu.cn}

\author{Miaomiao Cui}
\affiliation{%
  \small
  \institution{Alibaba Group}
  \streetaddress{}
  \city{}
  \country{}
}
\email{miaomiao.cmm@alibaba-inc.com}

\author{Peiran Ren}
\affiliation{%
  \small
  \institution{Alibaba Group}
  \streetaddress{}
  \city{}
  \country{}
}
\email{peiran.rpr@alibaba-inc.com}

\author{Xuansong Xie}
\affiliation{%
  \small
  \institution{Alibaba Group}
  \streetaddress{}
  \city{}
  \country{}
}
\email{xingtong.xxs@taobao.com}

\author{Xian-Sheng Hua}
\affiliation{%
  \small
  \institution{Damo Academy, Alibaba Group}
  \streetaddress{}
  \city{}
  \country{}
}
\email{xiansheng.hxs@alibaba-inc.com}

\author{Hujun Bao}
\affiliation{%
  \small
  \institution{Zhejiang University}
  \streetaddress{}
  \city{}
  \country{}
}
\email{bao@cad.zju.edu.cn}

\author{Qixing Huang}
\affiliation{%
  \small
  \institution{The University of Texas at Austin}
  \streetaddress{}
  \city{}
  \country{}
}
\email{huangqx@cs.utexas.edu}

\author{Weiwei Xu}
\affiliation{%
  \small
  \institution{Zhejiang University}
  \streetaddress{}
  \city{}
  \country{}
}
\email{xww@cad.zju.edu.cn}
\authornote{Corresponding author.}

\renewcommand{\shortauthors}{Zhang and Wang, et al.}

\begin{abstract}
This paper proposes a novel deep learning-based video object matting method that can achieve temporally coherent matting results. Its key component is an attention-based temporal aggregation module that maximizes image matting networks' strength for video matting networks. This module computes temporal correlations for pixels adjacent to each other along the time axis in feature space, which is robust against motion noises. We also design a novel loss term to train the attention weights, which drastically boosts the video matting performance. Besides, we show how to effectively solve the trimap generation problem by fine-tuning a state-of-the-art video object segmentation network with a sparse set of user-annotated keyframes. To facilitate video matting and trimap generation networks' training, we construct a large-scale video matting dataset with 80 training and 28 validation foreground video clips with ground-truth alpha mattes. Experimental results show that our method can generate high-quality alpha mattes for various videos featuring appearance change, occlusion, and fast motion. Our code and dataset can be found at: \url{https://github.com/yunkezhang/TCVOM}
\end{abstract}



\maketitle


\input{01_introduction.tex}
\input{02_related_works.tex}

\input{03_approach.tex}
\input{04_experiments.tex}
\input{05_discussion.tex}
\input{06_conclusion.tex}

\begin{acks}
We would like to thank anonymous reviewers for their constructive comments. Weiwei Xu is partially supported by National Key R\&D Program of China (2017YFB1002600) and NSFC (No.~61732016). Qixing Huang would like to acknowledge the support from NSFIIS-2047677 and NSF HDR TRIPODS-1934932.
\end{acks}

\bibliographystyle{ACM-Reference-Format}
\bibliography{main}

\end{document}

%% file: 01_introduction.tex
\vspace{-10pt}
\section{Introduction}

The task of video object matting is to compute temporally coherent alpha mattes for a foreground video object at each frame. It is a fundamental task for many video editing applications, \eg~compositing the foreground object into new background videos. The resulting alpha mattes represent the fractional opacity (between 0 and 1) of pixels. Such opacity mainly comes from the transparency or the partial coverage of background pixels around the foreground object boundaries. Specifically, the matting problem tries to solve for three types of unknowns at each pixel, i.e., the foreground color $F$, the background color $B$, and the alpha value $\alpha$, based on the measured pixel color $C$, where $C = \alpha F+(1-\alpha)B$. Moreover, to facilitate image and video matting, a trimap~\cite{wang2008image} is usually required to separate an image into the foreground region (FR), the background region (BR), and the unknown region (UR). Here UR covers partial or transparent foreground object boundaries.

\begin{figure}[t]
\begin{center}
    \setlength{\tabcolsep}{1pt}
    \resizebox{0.9\linewidth}{!}{
        \begin{tabular}{cccc}
        \rotatebox{90}{\scriptsize{Frames}} &
        \includegraphics[width=0.3\columnwidth]{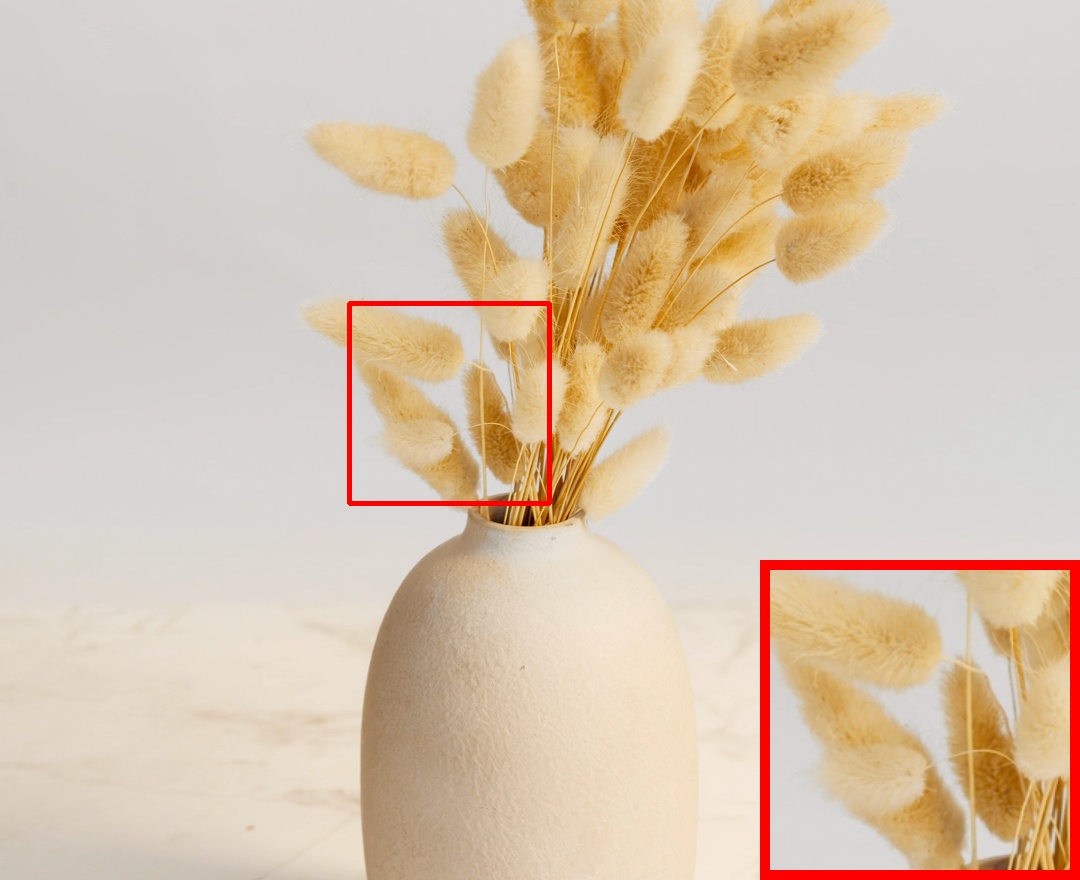} &
        \includegraphics[width=0.3\columnwidth]{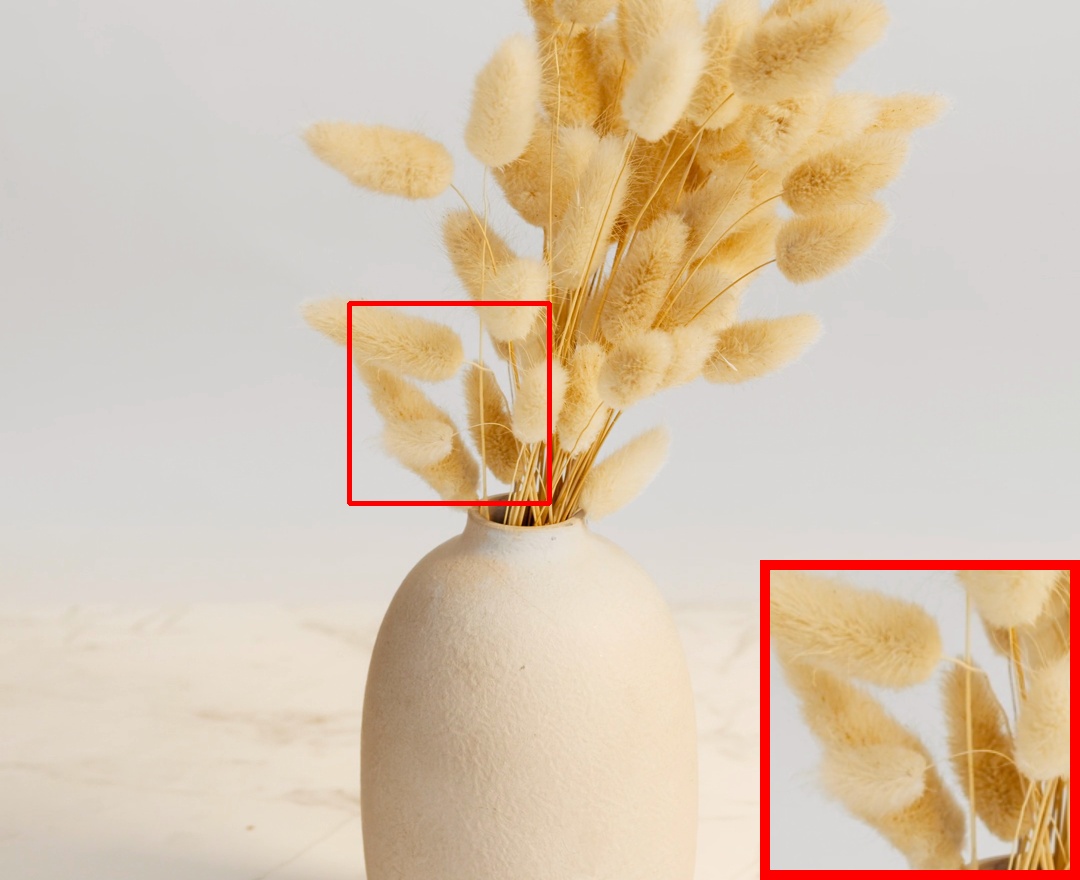} &
		\includegraphics[width=0.3\columnwidth]{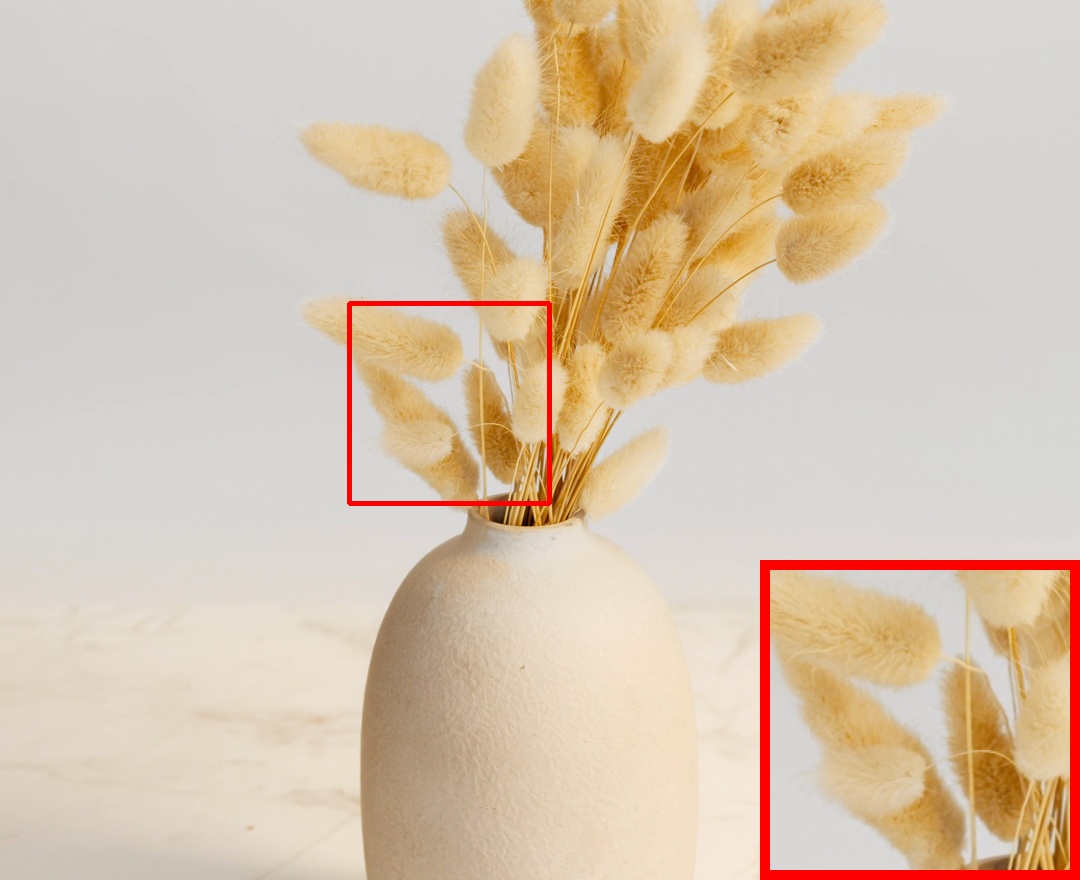} \\
        \rotatebox{90}{\scriptsize{Trimap}} &
		\includegraphics[width=0.3\columnwidth]{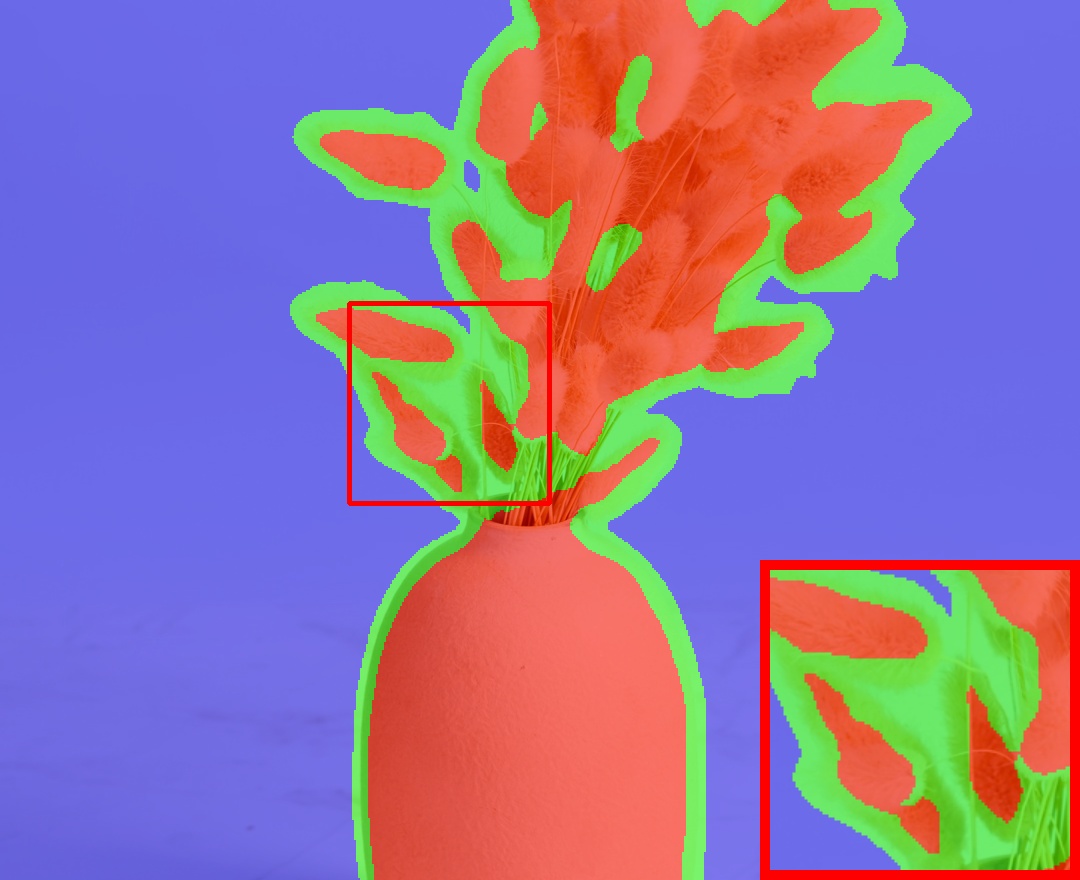} &
        \includegraphics[width=0.3\columnwidth]{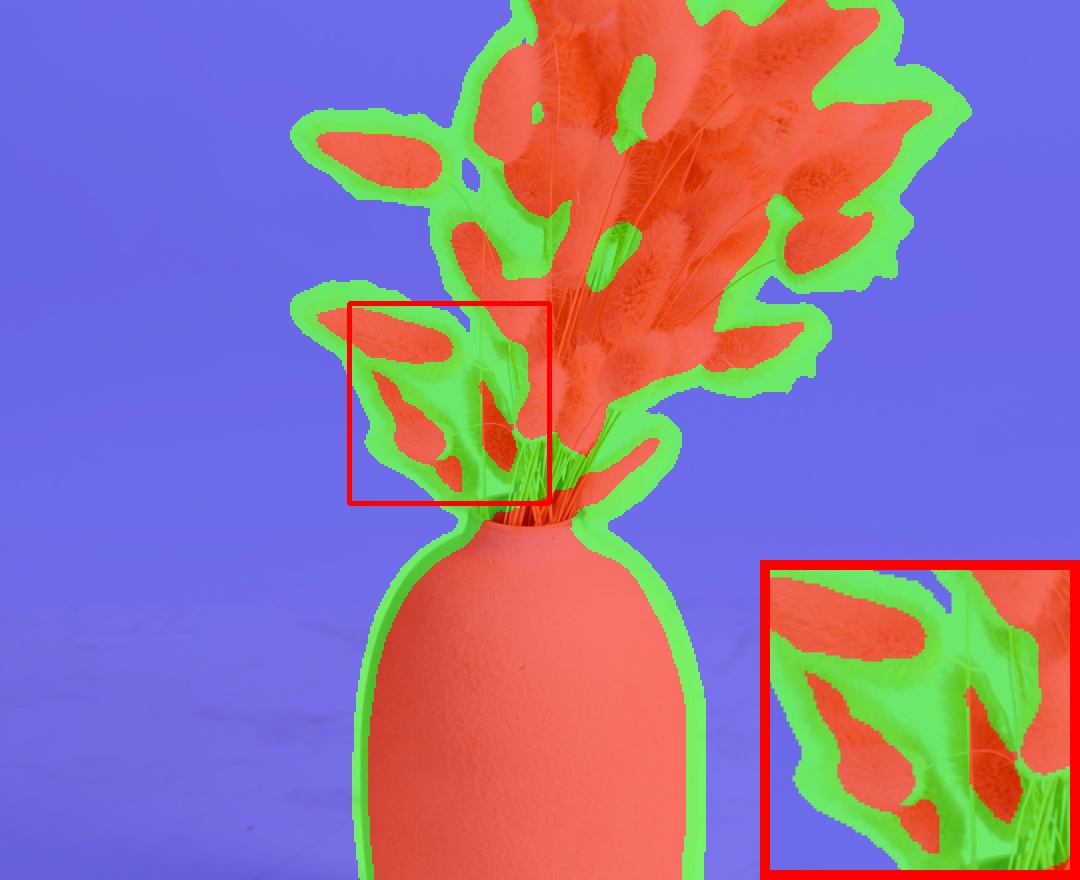} &
		\includegraphics[width=0.3\columnwidth]{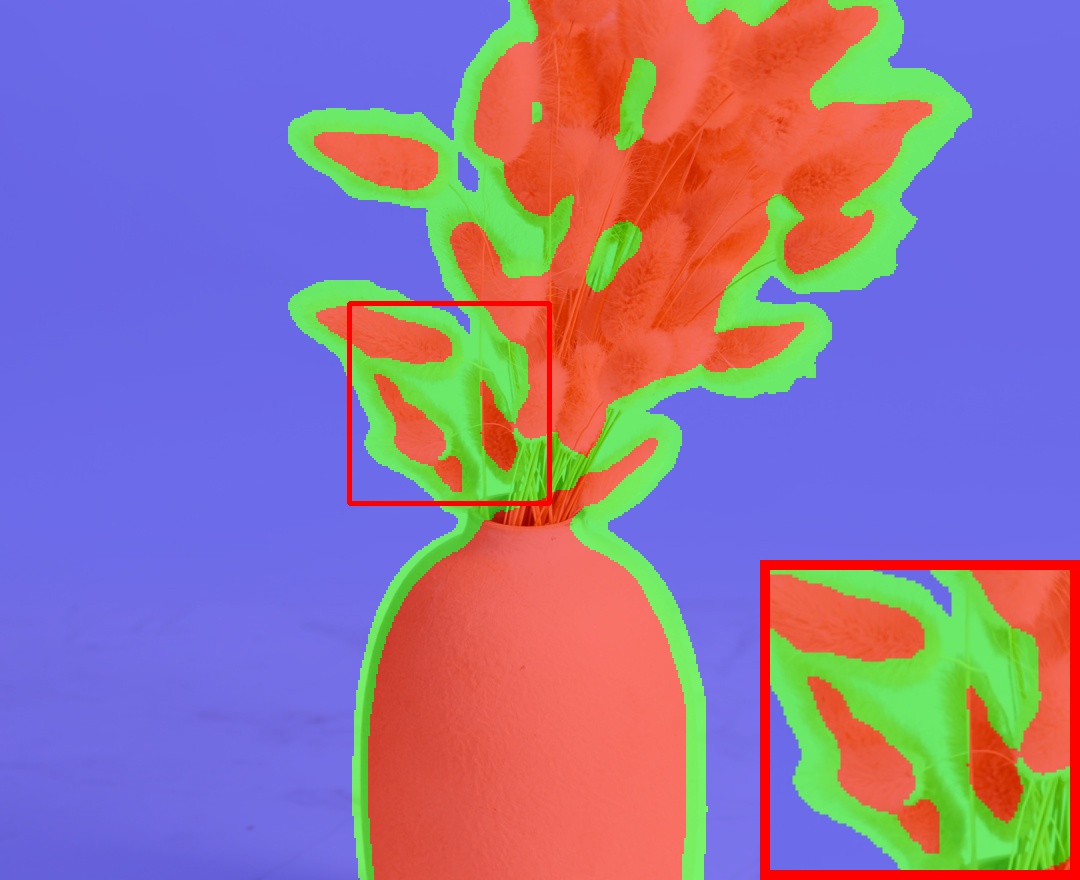} \\
        \rotatebox{90}{\scriptsize{GCA~\cite{li2020natural}}} & 
        \includegraphics[width=0.3\columnwidth]{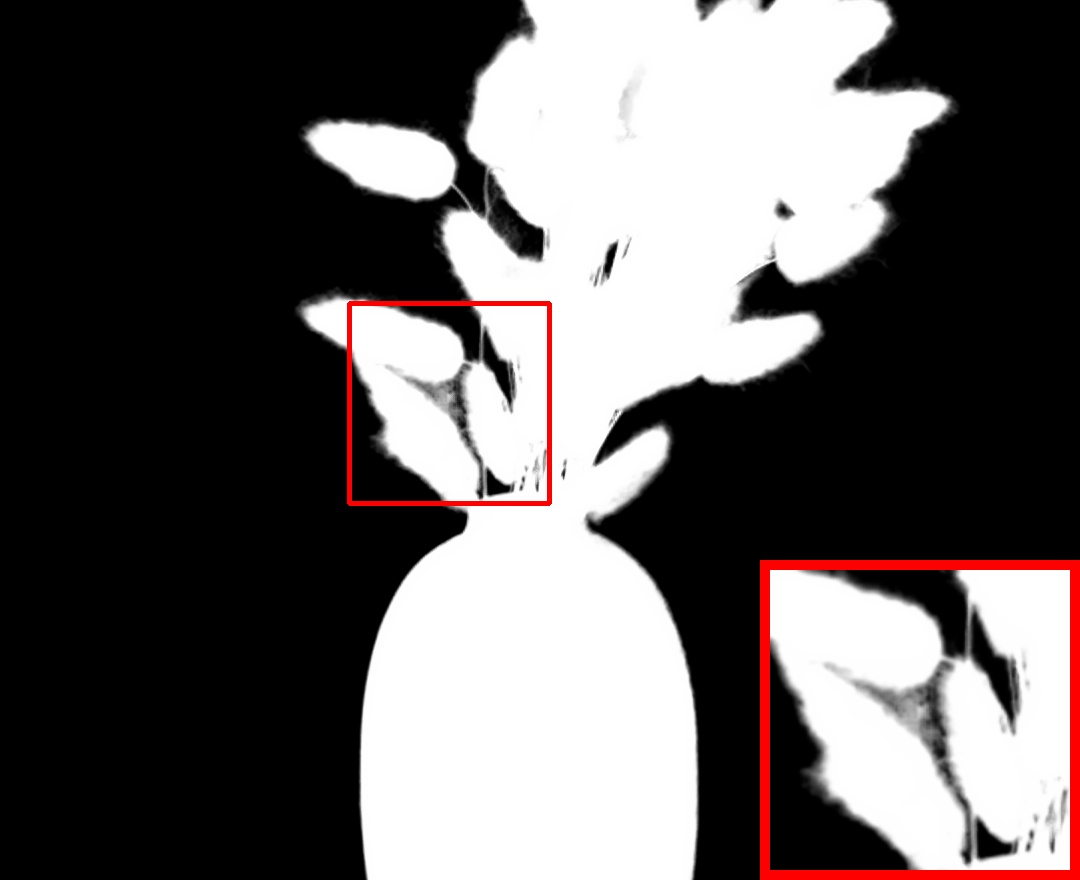} &
        \includegraphics[width=0.3\columnwidth]{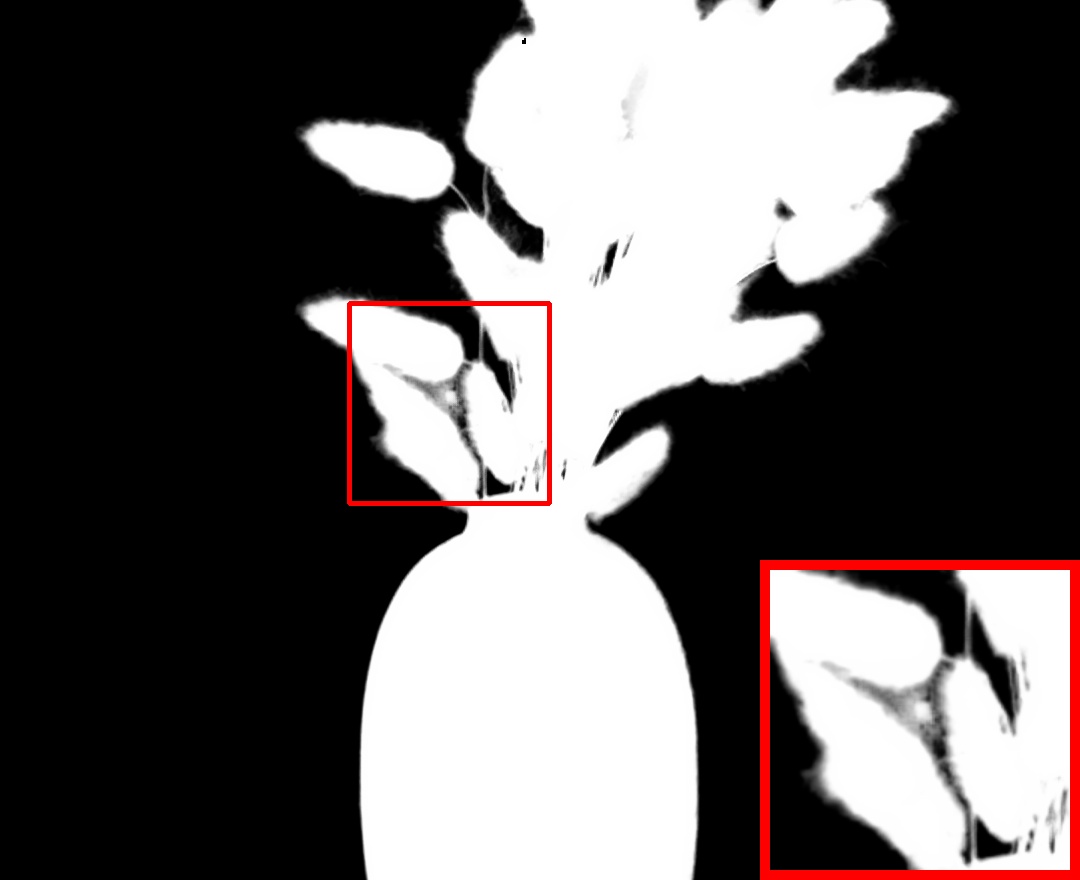} &
		\includegraphics[width=0.3\columnwidth]{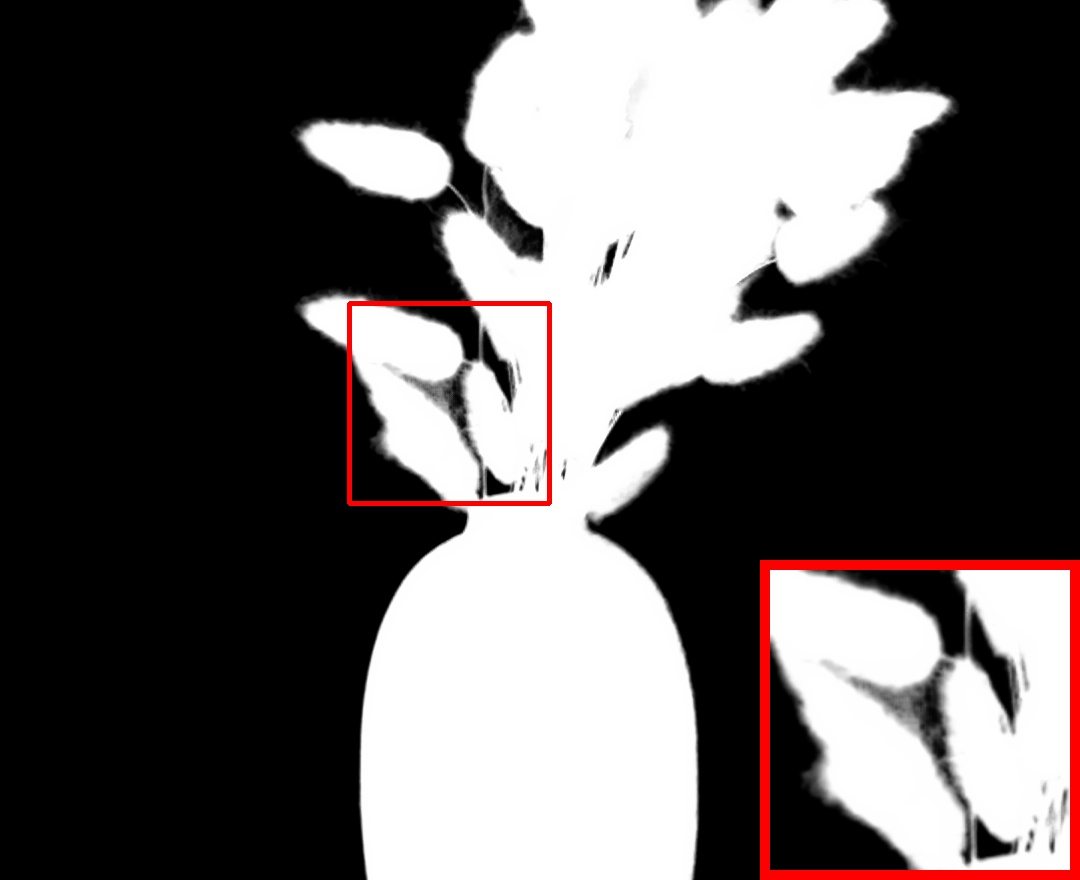} \\
        \rotatebox{90}{\scriptsize{Ours}} & 
        \includegraphics[width=0.3\columnwidth]{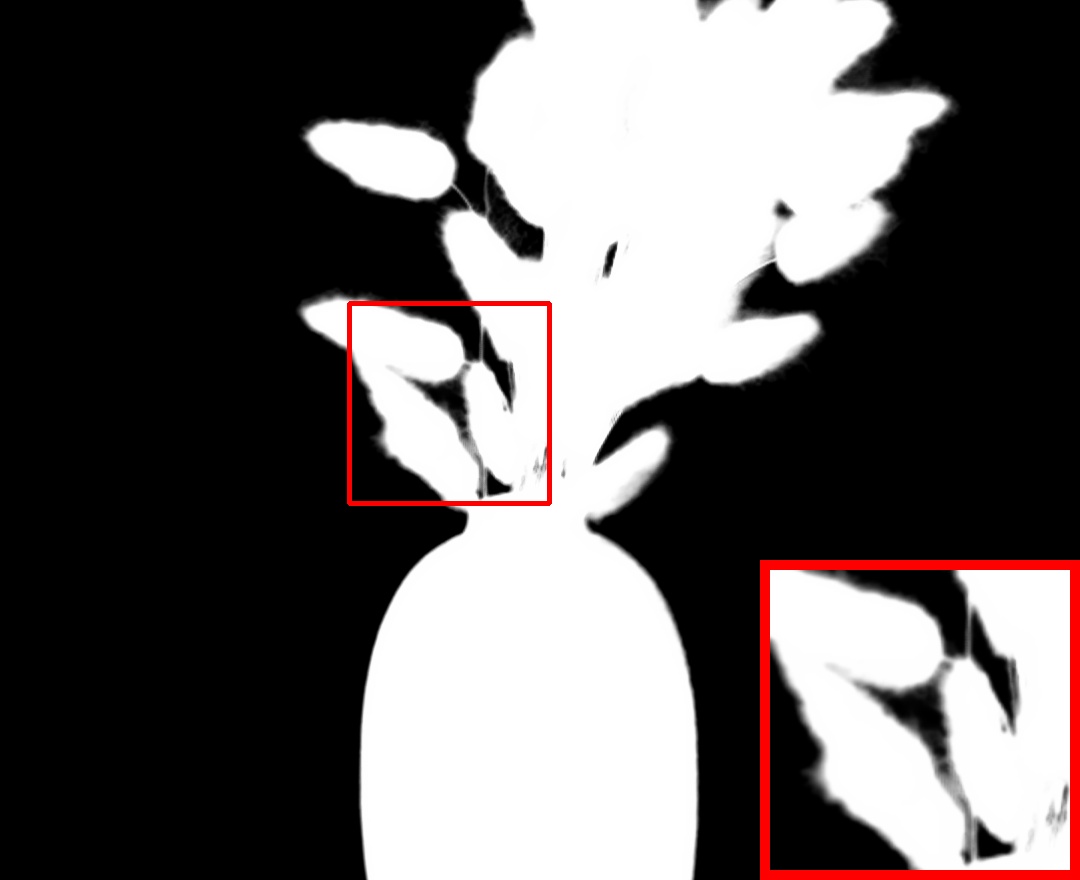} &
        \includegraphics[width=0.3\columnwidth]{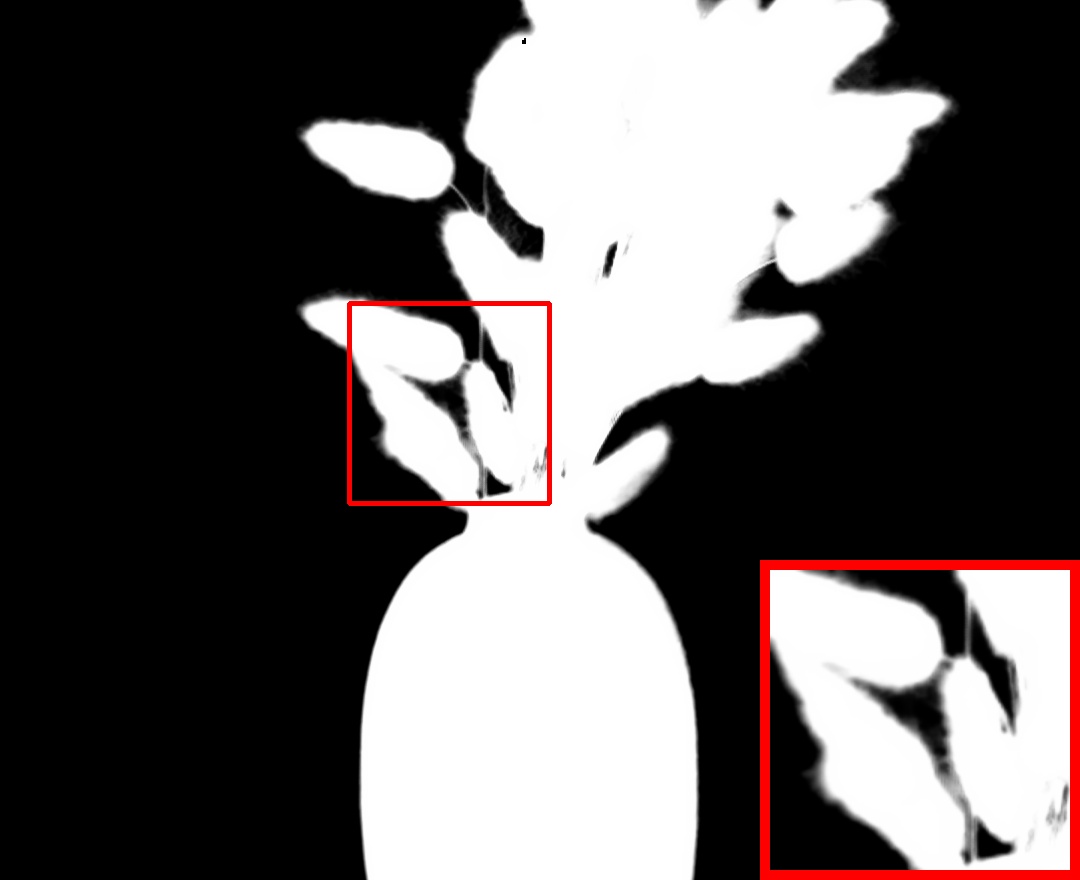} &
		\includegraphics[width=0.3\columnwidth]{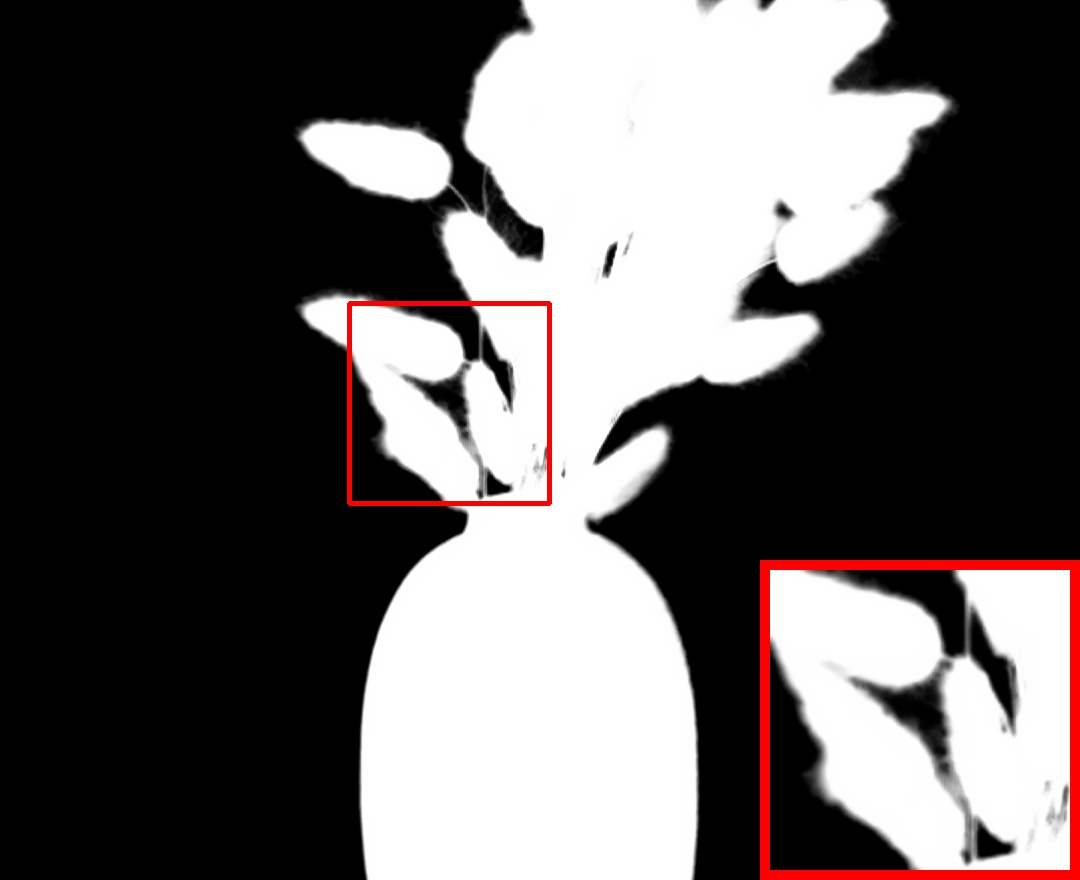} \\
		& \#94 & \#95 & \#96
        \end{tabular}
    }
\end{center}
\vspace{-15pt}
\caption{A video matting result comparison using an Internet video clip ``plant''. Trimaps are generated using our trimap generation method. Red, blue and green color correspond to FR, BR, and UR respectively. ``\#'' denotes the frame number. Our method is capable of generating more temporally coherent result compared to GCA\protect~\cite{li2020natural}, an image matting network. Please see the supplementary video for the complete result.}
\label{fig:teaser}
\vspace{-15pt}
\end{figure}

Video object matting is related to image matting in the sense that each frame of the matting output essentially solves the corresponding image matting problem. The matting problem is challenging since the number of unknowns exceeds the number of measured colors. Thus, it is critical to build priors to constrain the solution space~\cite{aksoy2017designing,chen2013knn,Chuang2001ABA,levin2006closed}. State-of-the-art (SOTA) image matting algorithms typically build on convolutional neural network (CNN). They improve the image matting results significantly by learning multi-scale features to predict alpha values for pixels in the UR~\cite{Cai_2019_ICCV,chen2018tomnet,cho2016natural,Hou_2019_ICCV,DBLP:conf/bmvc/LutzAS18,Tang_2019_CVPR,xu2017deep}.
Given an input video clip and its corresponding trimap for each frame, one can perform video matting with any image matting method by processing each video frame independently. However, this approach may lead to temporal incoherence in the obtained alpha mattes (\eg~flickering, shown in the third row of Figure~\ref{fig:teaser}).
To improve temporal coherence, existing video matting methods exploit temporal correspondence between video frames, such as optical flow, to construct multi-frame alpha or color priors or compute temporal affinities to incorporate motion cues~\cite{Apostoloff2004,Choi2012,Chuang:2002,Li_2013_ICCV,Dongqing2020}. However, they rely on local color distributions as main features and may suffer from motion ambiguities at transparent pixels, resulting in flickering or blocky artifacts in the matting results. 

This paper proposes a novel CNN-based video object matting method to achieve temporally coherent results. Its essential component is a simple yet effective attention-based temporal aggregation module (TAM) that can be seamlessly combined with SOTA image matting networks, such as GCA~\cite{li2020natural}, IndexNet~\cite{Lu_2019_ICCV} (Index) and DIM~\cite{xu2017deep}, extending them into video matting networks. This simple design maximizes image matting networks' strength and yields a substantial performance boost for video matting, especially on temporal-related metrics. We leverage the widely used attention mechanism to compute the temporal attention weights~\cite{vaswani2017attention,wang2018non} for a pair of pixels adjacent to each other along the time axis. Conceptually, these weights are analogous to the non-local, temporal affinity values used in traditional affinity-based video matting methods ~\cite{Choi2012,eisemann2009spectral}. However, the attention weights are computed using high-dimensional features rather than local color and motion features. Moreover, we design a novel target affinity term to supervise the learning of attention weights. This term's ground-truth is automatically derived from the alpha matte and used in a cross-entropy loss to guide the training. Such design significantly improves our method's robustness against noises due to video compression, appearance change and motion. As shown in Figure~\ref{fig:teaser}, our method (the last row) can generate much more temporally coherent result.

Another challenge is generating trimaps for an input video clip to fulfill the task of video object matting. To this end, we propose to train the space-time memory network (STM)~\cite{oh2019video}, which is a semi-supervised video object segmentation (VOS) network, to segment each frame into FR, BR and UR. It only requires the user to annotate trimaps of a target object at several keyframes, usually three to five frames for a video clip of around 200 frames in our experiments, which enhances the efficiency of video matting significantly. To handle the large variations of user-annotated keyframe trimaps, we perform online-finetuning on the STM network. We then ensemble the bidirectional prediction results to improve the quality of generated trimaps. 

In summary, the main contributions of this work are:
\begin{itemize}
\vspace{-5pt}
\setlength{\parskip}{0pt}
\setlength{\itemsep}{0pt}
\item We propose a temporal aggregation module that integrates image matting networks to achieve temporally coherent video matting results. It leverages the attention mechanism to compute temporal affinity values in the feature space, resulting in a robust matting method to handle challenging videos featuring appearance change, occlusion, and fast motion.
\item We propose an STM-based trimap generation method to enhance the efficiency of video matting greatly. The user only needs to annotate trimaps at several keyframes to generate trimap for every video frame.
\item To enable video object matting and trimap generation networks training, we construct a video object matting dataset, termed VideoMatting108, that covers various objects and different types of motions. In total, our dataset has 108 foreground video clips with ground-truth alpha mattes, all in 1080p resolution, averaging in 821 frames per clip. The dataset will be made publicly available.
\end{itemize}

%% file: 02_related_works.tex
\vspace{-10pt}
\section{Related Works}

\noindent \textbf{Image matting.} The sampling-based image matting methods~\cite{Chuang2001ABA,Feng2016ACS,gastal2010shared, he2011global,Ruzon2000Alpha} build the FR and BR color priors using the sampled pixels to infer the alpha values, while the affinity-based methods~\cite{ aksoy2017designing,aksoy2018semantic,Xue2009AGF,chen2013knn,grady2005random,levin2006closed,levin2008spectral,Sun2004Poisson} propagate the alpha values from the known FR and BR pixels to the UR pixels based on affinity score and have proven to be robust when dealing with complex images~\cite{Chuang2001ABA,gastal2010shared,Ruzon2000Alpha}. The deep learning-based matting methods usually train a convolutional encoder-decoder neural network to predict alpha values or foreground/background colors with user-specified trimaps~\cite{Cai_2019_ICCV,chen2018tomnet,cho2016natural,Hou_2019_ICCV,li2020natural,Lu_2019_ICCV,DBLP:conf/bmvc/LutzAS18,Tang_2019_CVPR,zhou2020attention}. Recently, ``trimap-free'' image matting methods also received much attention as they do not require user annotation. Some of the methods use other forms of prior instead of trimaps, \eg~background image~\cite{sengupta2020background}, rough segmentation map or coarse alpha matte~\cite{yu2020mask}. Others do not use any prior at all~\cite{Zhang_2019_CVPR,liu2020boosting,Qiao_2020_CVPR}. 
The most used image matting dataset in this line of research is provided by Xu \etal~\cite{xu2017deep}, and a larger dataset is proposed recently by Qiao \etal~\cite{Qiao_2020_CVPR}.

\begin{figure*}[t]
    \centering
    \includegraphics[width=0.9\textwidth]{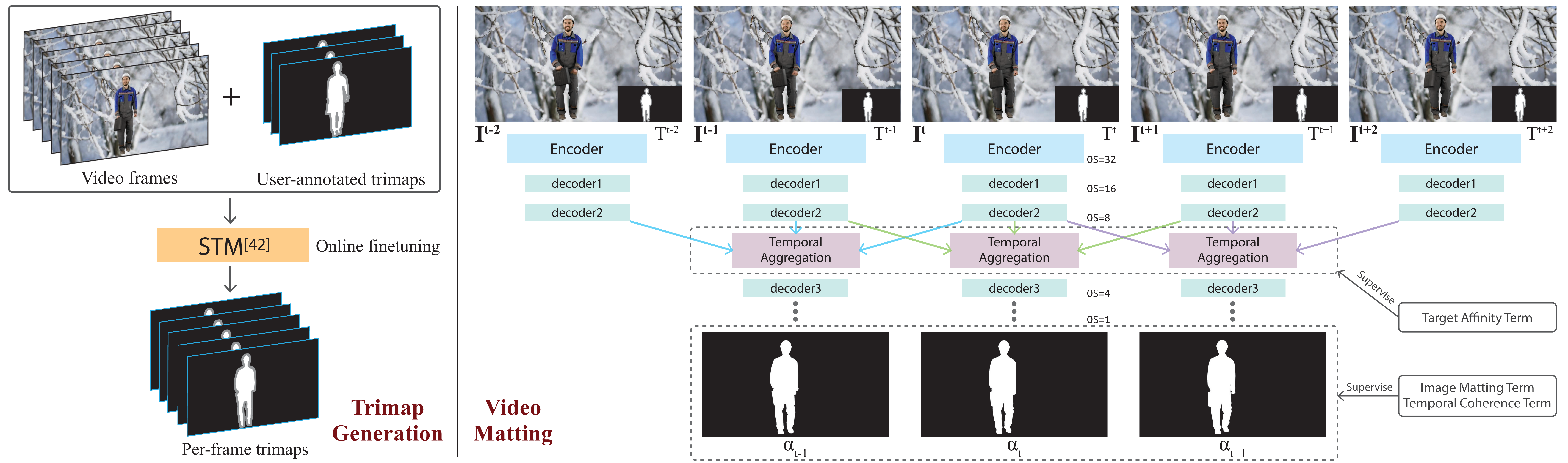}
    \vspace{-10pt}
    \caption{The flowchart of our method during training. ``OS'' denotes output stride. We do not show encoder-decoder skip connections for clarity. All networks and modules share the same weight across different frames.}
    \label{fig:flowchart}
    \vspace{-10pt}
\end{figure*}

\noindent \textbf{Video matting.} The central problem of video matting is how to obtain temporally coherent alpha mattes. Chuang \etal~\cite{Chuang:2002} proposed to interpolate manually specified trimaps at key-frames using optical flow, estimate the background pixels, and then perform Bayesian matting at each frame with the estimated background. Motion cues and prior distributions for alpha values and multi-frame colors are widely used in video matting~\cite{Apostoloff2004,Shum:2004:PLF,Alphaflow2013,Xiao2005Layer}. In~\cite{Choi2012,eisemann2009spectral,LEE201025,Li_2013_ICCV}, spatio-temporal edges between pixels are constructed to compute the alpha mattes for video frames simultaneously. These methods are the extension of affinity-based methods to video matting, which is time-consuming due to the Laplacian matrix's fast-growing size. Zou \etal~\cite{Dongqing2020} proposed to select nonlocal neighbors through sparse coding to constrain pixels having similar features in different frames to get similar alpha values. Besides, depth information can help to construct trimaps and differentiate between pixels of similar colors~\cite{ZhuJointDepthandMatte2009}. Hardware-assisted methods in~\cite{Joshi:2006:NVM,McGuire:2005:DVM}
automatically generate and propagate trimaps in all video
frames and optimize for high-quality alpha mattes. Recently, CNN-based video matting methods gained much attention. Lin \etal~\cite{Lin_2021_CVPR} and Ke \etal~\cite{ke2020green} proposed real-time CNN-based methods for trimap-less human portrait matting. However, both of the methods do not enforce the temporal consistency between video frames during training, which may lead to temporally incoherent results. Our method on the other hand, is a trimap-based video matting method that can handle different types of objects with explicitly supervised temporal consistency. Concurrent to our work, Sun \etal~\cite{Sun_2021_CVPR} also proposed a CNN-based video matting method that focuses on temporal coherency.

\noindent \textbf{Attention mechanisms in segmentation and matting.} Attention mechanism provides an effective way for neural networks to reinforce correlated features and suppress feature noise, leading to a performance boost in segmentation. There are two main variations of this mechanism. One is the channel-wise self-attention pioneered by Hu \etal~\cite{hu2018squeeze}. Given an input feature tensor, it leverages the global average pooling and the fully-connected layer to infer a channel-wise weight vector to modulate feature maps. The other is the non-local block proposed by Wang \etal~\cite{wang2018non}. It computes the spatiotemporal correlation as the attention, reinforcing the consistency of feature maps effectively. The channel-wise attention approach is widely adopted in image segmentation~\cite{yu2018bisenet,yu2018learning,zhang2018context}. Many methods~\cite{fu2019dual,Yu-ECCV-RepGraph-2020,yu2020context,DBLP:conf/eccv/YuanCW20,zhao2018psanet} exploit variants of non-local attention modules to capture the spatial long-range dependency. For image matting tasks, the attention mechanism is mostly used for fusing high and low-level image features. Qiao \etal~\cite{Qiao_2020_CVPR} adopted both channel-wise and spatial attention for trimap-free matting since high-level image features are the key to recognize a foreground object. GCA~\cite{li2020natural} also utilizes high-level image features as the attention map to guide the low-level alpha features, achieving SOTA performance in image matting. We thus employ GCA as one of the base matting network structures. 
Several recent VOS methods also utilize the attention mechanism to fuse features from different video frames for improving temporal consistency~\cite{oh2019fast,Voigtlaender_2019_CVPR}. Oh \etal~\cite{oh2019video} extended the memory network approach used in NLP to VOS, which is also a variation of the spatiotemporal attention mechanism. Yang \etal~\cite{yang2020collaborative} extended this idea by matching both the foreground and background with multi-scale features in those frames, achieving SOTA performance. Our method also leverages the attention mechanism for temporally coherent matting. Nevertheless, our attention module is bi-directional, and we use additional temporal loss terms to supervise the network training.

\noindent \textbf{Temporal coherence.} One standard solution to temporal coherence is the temporal smoothing filter, which considers the spatial and temporal adjacent pixels simultaneously~\cite{chang2007example,Chen_2017_ICCV,Lang:2012:PTC,Paris2008}. Another solution is to impose the temporal coherence in the post-processing, which is blind to image filters~\cite{Bonneel:2015:BVT,Lai_2018_ECCV}. In contrast, our method does not rely on temporal smoothing filter but the feature-space affinity to produce temporally coherent alpha mattes.

%% file: 03_approach.tex
\vspace{-10pt}
\section{Our method}
\label{sec:our_method}
Given an input video, our method first runs trimap generation to propagate the user-annotated trimaps to the other frames. We then run a video matting network, formed by integrating temporal aggregation module (TAM) into an image-based matting network, to obtain a temporally coherent alpha matte at each frame (See Figure~\ref{fig:flowchart}).
When computing an alpha matte for frame $\mathbf{I}^t$ in testing stage, TAM only needs to aggregate the CNN features from three consecutive frames, \ie~$\mathbf{I}^{t-1},\mathbf{I}^t,\mathbf{I}^{t+1}$. The choice of three consecutive frames offers great flexibility in network design while ensures computational efficiency. However, during training, our  network takes five consecutive frames simultaneously as inputs, \ie~$\mathbf{I}^{t-2},...,\mathbf{I}^{t+2}$, and predicts $\alpha^{t-1},\alpha^{t},\alpha^{t+1}$ to facilitate the computation of loss functions. Note that we choose to integrate TAM into the base network at the decoder stage of output stride (OS) 8. It indicates that the resolution of the feature map should be $H/8 \times W/8$, where $H,W$ is the input image resolution. This choice is to balance computational cost and feature level, and we empirically found that OS=8 is a good trade-off (see the supplementary material for the OS experiment). 

In the following, we will first describe the design of TAM (Sec.~\ref{sec:tam}), and proceed to describe the training loss (Sec.~\ref{sec:loss}) and the training strategy of TAM (Sec.~\ref{sec:tam_training}). Finally, we describe the details of trimap generation using STM~\cite{oh2019video} (Sec.~\ref{sec:trimap_generation}) and our video object matting dataset (Sec.~\ref{sec:dataset}).

\vspace{-5pt}
\subsection{Temporal Aggregation Module}
\label{sec:tam}
\begin{figure}[t]
    \begin{center}
    \includegraphics[width=0.9\linewidth]{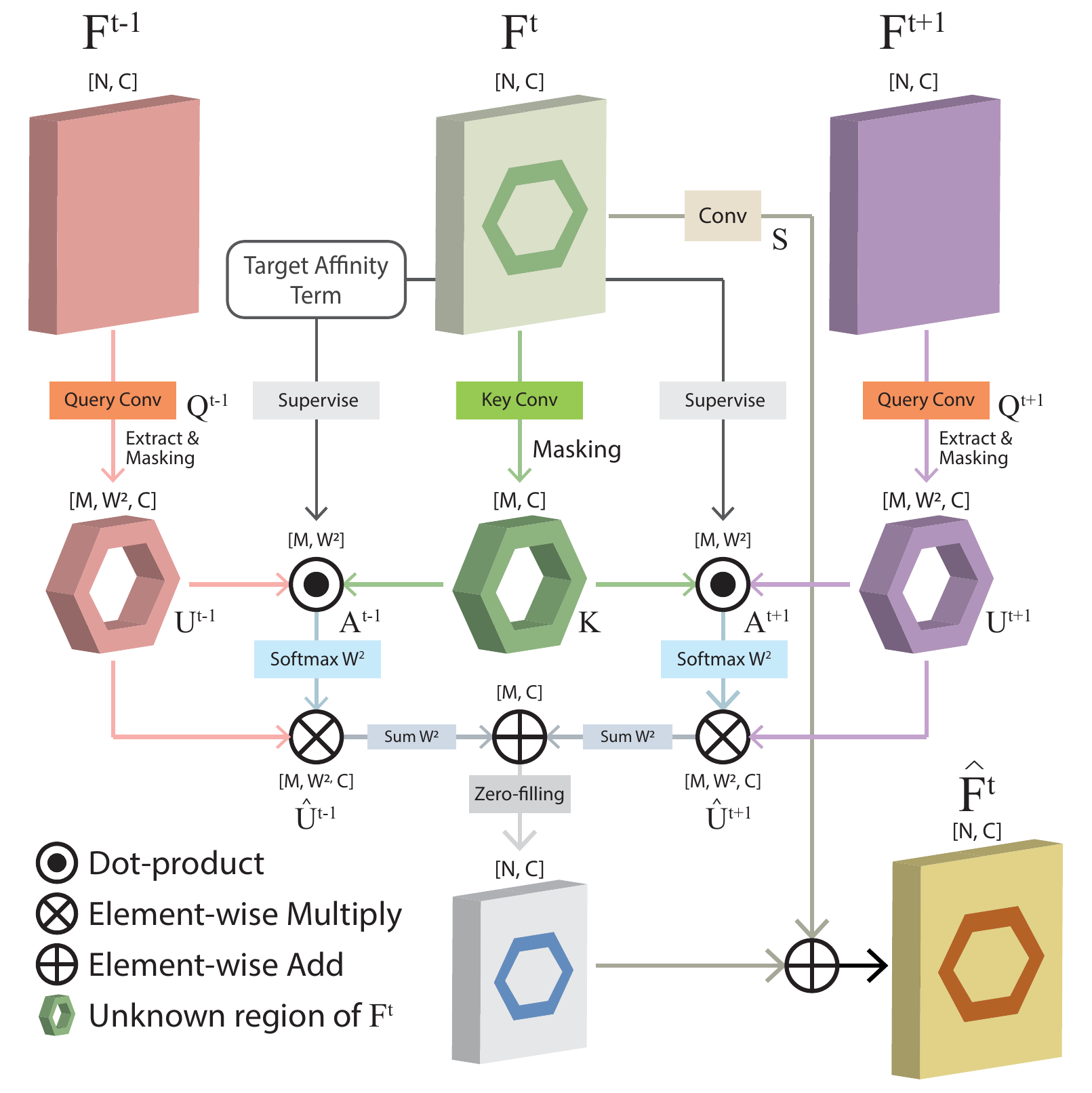}
    \caption{The architecture of our temporal aggregation module.}
    \label{fig:tam}
\end{center}
\vspace{-15pt}
\end{figure}

Figure~\ref{fig:tam} illustrates the structure of TAM. It leverages the attention mechanism to aggregate features from $\mathbf{I}^{t-1}$ and $\mathbf{I}^{t+1}$ with features from $\mathbf{I}^t$ for pixels inside the UR of $\mathbf{I}^t$. This design benefits temporal coherence by encoding temporal correlation in the aggregated features at frame $\mathbf{I}^t$.   

For a video frame $\mathbf{I}^t$, we denote its input feature map as $\mathbf{F}^t \in \mathbb{R}^{N \times C}$. Here $C$ is the total number of channels, and $N$ is the total number of pixels in this feature map. TAM takes three feature maps $\mathbf{F}^{t-1},\mathbf{F}^t,\mathbf{F}^{t+1}$ as inputs.
First, we compute key and query features with separate convolution layers. Specifically, we feed $\mathbf{F}^{t-1},\mathbf{F}^{t+1}$ into a $3 \times 3$ query convolution layer. With $\mathbf{Q}^{t-1},\mathbf{Q}^{t+1}$ we denote the output query feature map. In contrast, we input $\mathbf{F}^t$ into a $3 \times 3$ key convolution layer and denote the output feature map as $\mathbf{K}$. 

Since we only focus on UR of $\mathbf{I}^t$ in TAM, we extract the UR in $\mathbf{K}$ using a mask generated by down-sampling the original trimap to the same resolution. For each pixel $i$ within the UR of $\mathbf{K}$, we extract two $W \times W$ feature patches centered at the same position of pixel $i$ from features $\mathbf{Q}^{t-1}$ and $\mathbf{Q}^{t+1}$, respectively. We denote this window as $\mathbb{W}$ and the extracted patch features as $\mathbf{U}^{t-1},\mathbf{U}^{t+1} \in \mathbb{R}^{M \times W^2 \times C}$, where $M$ is the number of pixels in the UR.

Next, we compute the attention weights between $\mathbf{K}$ and $\mathbf{U}^{t-1}$ and similarly for $\mathbf{K}$ and $\mathbf{U}^{t+1}$, and then use the attention weights as temporal affinity values to modulate the corresponding adjacent frame features. Taking $\mathbf{K}$ and the query features $\mathbf{U}^{t-1}$ of $\mathbf{I}^{t-1}$ as an example, we compute the set of affinity values $\mathbf{A}^{t-1} \in \mathbb{R}^{M \times W^2}$ as follows:
\vspace{-5pt}
\begin{equation}
    \mathbf{A}^{t-1}(i,j) = \frac
    {
        {\rm e}^{\mathbf{K}(i) \cdot \mathbf{U}^{t-1}(i,j)}
    }
    {
        \sum_{k=1}^{W^2}
        {\rm e}^{\mathbf{K}(i) \cdot \mathbf{U}^{t-1}(i,k)}
    }, i \in UR, j \in \mathbb{W}.
\label{eq:TAM_aff}
\vspace{-5pt}
\end{equation}
\noindent where $\mathbf{A}^{t-1}(i,j)$ denotes the affinity value computed for a pixel $i$ of $\mathbf{I}^t$ and a pixel $j$ within the patch at $\mathbf{I}^{t-1}$; note that $\cdot$ denotes the point-wise dot product. This operation effectively matches the key feature $\mathbf{K}(i)$ of pixel $i$ with features from $\mathbf{U}^{t-1}$ inside a corresponding $W \times W$ local patch. While our module only uses a local patch instead of the full feature map in the computation attention weights, empirically, it is enough to capture the temporal correlation since the motion between adjacent frames in a video sequence is relatively small. The patch size is set to $7\times 7$ throughout our experiments if not noted otherwise. Note that we use feature maps of OS=8, which can cover a broad range of motion.

Finally, we formulate feature aggregation for $\mathbf{I}^t$ as follows:
\vspace{-3pt}
\begin{equation}
\begin{array}{rl}
    \mathbf{\hat{F}}^{t}(i)&=\Big\{
    \begin{array}{cl}
        \mathbf{S}(i)+
        \mathbf{\hat{U}}^{t-1}(i)+
        \mathbf{\hat{U}}^{t+1}(i) &,i \in UR\\
        \mathbf{S}(i) &, Otherwise
    \end{array} \\
    \mathbf{\hat{U}}^{f}(i)&=\sum^{W^2}_{j=1} {\mathbf{A}^{f}(i,j)\mathbf{U}^{f}(i,j)}, f=\{t-1,t+1\}
\end{array}
\label{eq:TAM_final}
\vspace{-3pt}
\end{equation}
where the modulated feature $\mathbf{\hat{U}}^{t-1} \in \mathbb{R}^{M \times C}$ is a weighted sum of all features inside the local patch using $\mathbf{A}^{t-1}$, and $\mathbf{S}$ indicates the features output by a $3\times 3$ convolution layer at frame $\mathbf{I}^t$. The features $\mathbf{\hat{F}}^t$ are fed into next convolution layer to continue the workflow of the original network. 

Note that the TAM's design can be seen as sharing the weight between the ``query'' and ``value'' convolution in the conventional ``KQV'' structure. We choose this design because it achieves the best performance among other weight sharing configurations. The influence of the weight sharing will be investigated later in Sec.~\ref{sec:experiment}'s ablation study.

\vspace{-5pt}
\subsection{Loss function}
\label{sec:loss}
The overall loss function is a composition of three different terms:
the image matting term $L_{im}$, the temporal coherence term $L_{tc}$, and the target affinity term $L_{af}$.  In the following, we describe these three terms in details.

\noindent \textbf{Image matting term.} The image matting term $L_{im}$ is directly inherited from the deep learning-based image matting methods. This term only considers the single frame prediction results w.r.t. its corresponding ground truth. Common choices of this term are the combination of $L_1$ alpha matte loss, $L_1$ alpha gradient loss, composition loss~\cite{xu2017deep} and Laplacian loss~\cite{yu2020context}. In all our experiments, we set $L_{im}$ the same as the image matting method we chose to use as the base network. Hence, for DIM~\cite{xu2017deep} and Index~\cite{Lu_2019_ICCV}, we use alpha matte loss, alpha gradient loss, and composition loss; for GCA~\cite{li2020natural}, we only use alpha matte loss.

\noindent \textbf{Temporal coherence term.} We leverage the temporal coherence metric proposed by Erofeev \etal~\cite{Erofeev2015} as the temporal coherence term $L_{tc}$, which can be expressed as follows:
\vspace{-1pt}
\begin{equation}
    L^{p,q}_{tc}(i) = |(\alpha^{t-1}_{i} - \alpha^{t}_{i}) - (\hat{\alpha}^{t-1}_{i} - \hat{\alpha}^{t}_{i})|_1,
\label{eq:L_tg}
\vspace{-1pt}
\end{equation}
\noindent where $\alpha,\hat{\alpha}$ are the predicted and ground-truth alpha matte. $i$ denotes a pixel in the UR of frame $p$. This term is denoted by ``dtSSD'' in~\cite{Erofeev2015}, which penalizes the temporal gradient of alpha mattes in our work. There is another temporal coherence metric, termed as ``MESSDdt'' in~\cite{Erofeev2015}, which augments Eq.~\ref{eq:L_tg} with motion vectors for the purpose of better correspondence between frame $t-1$ and frame $t$. However, we found that even computed with SOTA optical flow algorithms~\cite{teed2020raft}, there are severe motion noises for translucent pixels, which hurts the performance of the network when using ``MESSDdt'' as the temporal coherence loss. Thus, we do not use ``MESSDdt'' in our implementation. The quantitative result of using ``MESSDdt'' as a loss term can be found in the supplementary material.

\noindent \textbf{Target affinity term.} While the attention mechanism works well in many cases, Yu \etal~\cite{yu2020context} demonstrates that providing direct supervision for the attention weights can further boost the performance. Inspired by this work, we thus design the target affinity term $L_{af}$. For a pixel $i \in \text{UR}$ at $\mathbf{I}^t$, the target probability of having a large attention weight between $i$ and a pixel $j \in \mathbb{W}$ inside its local patch at a neighboring frame $f$ can be formulated as:
\vspace{-5pt}
\begin{equation}
\begin{array}{rl}
G^{f}(i,j)&=\Big\{
    \begin{array}{cl}
         1-s, &  |\hat{\alpha}^{t}_i-\hat{\alpha}^{f}_{j}|<\theta \\
         0, &  Otherwise
    \end{array},
\end{array}
\vspace{-5pt}
\end{equation}
\noindent where $f=t-1$ or $t+1$, and $\theta$ is set to be $0.3$. In addition, we follow the label smoothing technique~\cite{szegedy2016rethinking} to avoid over-confident decision by introducing the parameter $s=0.2$. The target probability is computed according to the ground-truth alpha mattes. The goal of this loss term is to make the network learn to assign small affinity values to pixels with large alpha value differences. Therefore, we model the target affinity term between pixel $i$ and $j$ as
\vspace{-3pt}
\begin{equation}
\begin{array}{rl}
L^{f}_{af}(i,j)&=\mathbf{BCE}(\Phi(\mathbf{K}(i) \cdot \mathbf{U}^{f}(i,j)),G^{f}(i,j))
\end{array}
\vspace{-3pt}
\end{equation}
where $\mathbf{BCE}$ denotes the binary cross-entropy function, and $\Phi$ denotes the sigmoid function. During training, the term $L_{af}$ is calculated as $ L_{af}=\frac{1}{2}(L^{t-1}_{af}+L^{t+1}_{af})$.

In summary, our network is trained by the weighted average of these three terms:
\vspace{-5pt}
\begin{equation}
L = w_{im}L_{im} + w_{tc}L_{tc} + w_{af}L_{af}.
\label{eq:full_loss}
\vspace{-5pt}
\end{equation}
\noindent where we set $w_{im} = 1$, $w_{tc} =0.5$, and $w_{af} = 0.25$ in our experiments.

\begin{table*}[t]
    \caption{Result on VideoMatting108 validation set. GCA\protect~\cite{li2020natural}, Index\protect~\cite{Lu_2019_ICCV} and DIM\protect~\cite{xu2017deep} are used as the base image matting network structures to verify the effectiveness of the TAM. The best result is in bold, the second best is underlined. ``+F'' indicates the single image matting method is fine-tuned on our training dataset. ``+TAM'' denotes we add TAM for video matting. ``+TAM$_{share}$'' and ``+TAM$_{sep}$'' denote we share / separate all convolutions in TAM, respectively. ``MSDdt'' denotes ``MESSDdt''.}
    \vspace{-10pt}
    \label{tab:vm108_val}
    \small
    \centering
    \begin{tabularx}{\linewidth}{l|l|YYYYY|YYYYY|YYYYY}
        \hline
        \multirow{2}{*}{Method} & \multirow{2}{*}{Loss}
        & \multicolumn{5}{c|}{Narrow}
        & \multicolumn{5}{c|}{Medium}
        & \multicolumn{5}{c}{Wide}
        \\
        & & SSDA & dtSSD & MSDdt & MSE & SAD
        & SSDA & dtSSD & MSDdt & MSE & SAD
        & SSDA & dtSSD & MSDdt & MSE & SAD
        \\
        \hline
        GCA+F & $L_{im}$ 
        &	49.99	&	27.91	&	1.80	&	8.32	&	46.86	
        &	55.82	&	31.64	&	2.15	&	8.20	&	40.85	
        &	60.69	&	34.83	&	2.50	&	8.41	&	38.59	\\
        +TAM  & $L_{im}$ 
        &	\underline{46.86}	&	26.21	&	1.48	&	\underline{7.68}	&	\underline{44.82}
        &	54.01	&	29.49	&	1.78	&	7.90	&	39.51	
        &	59.09	&	32.55	&	2.07	&	8.18	&	37.41	\\
        +TAM$_{share}$  & $L_{im}$ 
        &	49.71	&	27.49	&	1.68	&	8.34	&	46.45
        &	57.20	&	29.90	&	1.91	&	8.88	&	41.15
        &	62.90 	&	33.13 	&	2.22 	&	9.35 	&	39.31 	\\
        +TAM$_{sep}$  & $L_{im}$ 
        &	54.06	&	27.69	&	1.78	&	10.37	&	48.03
        &	59.13	&	30.75	&	2.00	&	9.84	&	41.56
        &	64.89	&	33.90	&	2.30	&	10.37	&	39.78	\\
        +TAM  & $L_{im}$+$L_{tc}$ 
        &	48.35	&	\underline{25.04}	&	\underline{1.43}	&	8.00	&	45.47	
        &	\underline{52.83}	&	\underline{27.81}	&	\underline{1.60}	&	\underline{7.55}	&	\underline{38.84}
        &	\underline{57.51}	&	\underline{30.34}	&	\underline{1.84}	&	\underline{7.73}	&	\underline{36.57}	\\
        +TAM & $L_{im}$+$L_{af}$ 
        & 46.87 & 25.70 & 1.47 & 7.70 & 45.22
        & 53.00 & 28.97 & 1.72 & 7.73 & 39.47
        & 58.08 & 31.97 & 2.00 & 8.05 & 37.47 \\
        +TAM      & $L_{im}$+$L_{tc}$+$L_{af}$ 
        &	\textbf{45.39}	&	\textbf{24.37}	&	\textbf{1.28}	&	\textbf{7.30}	&	\textbf{44.01}
        &	\textbf{50.41}	&	\textbf{27.28}	&	\textbf{1.48}	&	\textbf{7.07}	&	\textbf{37.65}
        &	\textbf{54.35}	&	\textbf{29.60}	&	\textbf{1.69}	&	\textbf{6.98}	&	\textbf{34.81}	\\
        \hline
        Index+F     & \resizebox{13pt}{5pt}{$L_{im}$}
        & 52.75 & 29.49 & 1.97 & 9.78 & 50.90
        & 58.53 & 33.03 & 2.33 & 9.37 & 43.53
        & 64.49 & 36.39 & 2.73 & 9.73 & 41.22\\
        +TAM      & $L_{im}$+$L_{tc}$+$L_{af}$ 
        & 51.18 & 26.31 & 1.52 & 8.87 & 50.02
        & 57.91 & 29.36 & 1.81 & 8.78 & 43.17
        & 63.56 & 32.09 & 2.10 & 9.21 & 40.97\\
        \hline
        DIM+F     & $L_{im}$ 
        & 56.40 & 31.77 & 2.56 & 10.46 & 51.76
        & 61.85 & 34.55 & 2.82 & 9.99 & 44.38
        & 67.15 & 37.64 & 3.21 & 10.25 & 41.88\\
        +TAM      & $L_{im}$+$L_{tc}$+$L_{af}$ 
        & 53.61 & 27.77 & 1.90 & 9.48 & 50.12
        & 58.94 & 29.89 & 2.06 & 9.02 & 43.28
        & 63.27 & 32.15 & 2.31 & 8.88 & 40.45\\
        \hline

    \end{tabularx}
    \vspace{-10pt}
\end{table*}

\vspace{-5pt}
\subsection{Training strategy}
\label{sec:tam_training}
The training of our video matting network consists a pre-training stage and a main stage. In the following, we denote the video matting network as GCA+TAM, DIM+TAM or Index+TAM which corresponds to base methods GCA~\cite{li2020natural}, DIM~\cite{xu2017deep} and Index~\cite{Lu_2019_ICCV} respectively. For pre-training, we input three frames $\mathbf{I}^{t-1},\mathbf{I}^{t},\mathbf{I}^{t+1}$ along with trimaps and predict the center frame alpha matte $\alpha^t$ using the supervision from $L_{im}$ only. During pre-training, all layers before the TAM in the network are initialized and fixed using the pre-trained weight of a off-the-shelf image matting network. TAM and the rest of the decoder layers are randomly initialized and trained on the DIM dataset~\cite{xu2017deep}. The dataset is augmented with random affine transformations (rotation, translation, and scaling) to generate sequences of three frames. Random flipping and cropping are also conducted for further augmentation. We pre-trained the network for 20 epochs using an input resolution of $512 \times 512$ with the Adam optimizer~\cite{KingmaB14}. For different base image matting methods, we used different batch sizes and learning rates. The batch size is set to $40$ for both DIM+TAM and GCA+TAM, $24$ for Index+TAM. The learning rate is set to $10^{-5}$ for DIM+TAM, $10^{-4}$ for Index+TAM and $4\times10^{-4}$ with ``poly'' decay strategy for GCA+TAM where the decay rate is set to 0.9.

In the main stage, our network takes five consecutive frames $\{\mathbf{I}^{t-2},...,\mathbf{I}^{t+2}\}$ along with their corresponding trimaps as inputs. The motivation comes from the fact that the temporal coherence term requires alpha mattes of three consecutive frames, and each frame needs the features of its two neighboring frames for alpha matte prediction. The predicted $\{\alpha^{t-1},\alpha^{t},\alpha^{t+1}\}$ are used to compute the loss function in Eq.~\ref{eq:full_loss}. We use Adam as the optimizer to train the network on VideoMatting108 for 30 epochs with the same input resolution $512 \times 512$. Our data augmentation strategies include random shape augmentation, such as cropping, flipping and scaling, and random color augmentation, such as hue, saturation, gamma, and JPEG compression~\cite{yu2020context}. We use ``poly'' decay strategy with the base learning rate of $10^{-4}$, $10^{-5}$, $10^{-4}$ and decay rate of 0.9 during main training for GCA+TAM, DIM+TAM and Index+TAM. The batch size is set to 24, 16 and 24 respectively.

\vspace{-5pt}
\subsection{Video Trimap Generation}
\label{sec:trimap_generation}

We leverage the SOTA VOS network STM~\cite{oh2019video} to segment each frame into FR, BR, and UR. That is, we let STM track FR and UR as two different objects in a video and classify the remaining pixels that do not belong to FR and UR as BR. To obtain the ground-truth labels, we label the translucent pixels obtained in the construction of VideoMatting108 without any dilation as UR, and the pixels with alpha value equals one as FR. Additionally, we give UR a higher weight (4.0 in training) to achieve class-balanced cross-entropy loss since UR generally has fewer pixels than FR and BR.

\noindent \textbf{Training parameters.} Same with STM, we also utilize the two-stage training strategy. First, the network is initialized from the weight pre-trained on ImageNet~\cite{deng2009imagenet}. We then use the DIM dataset~\cite{xu2017deep} augmented with random affine transformations (rotation, translation, scaling, and shearing) to pre-train the network. The network is pre-trained for 25 epochs. We then proceed to the main training stage. The only difference is that we use a larger maximum frame skip, which is 75 frames in our implementation since the videos in VideoMatting108 are much longer compared to the VOS dataset like DAVIS~\cite{Perazzi2016DAVIS}. We train the network for 150 epochs, where every epoch consists of 850 iterations. The maximum frame skipping is gradually increased by one every two epochs. We also utilize the ``stair'' learning rate strategy with the base learning rate of $10^{-5}$, $5 \times 10^{-6},10^{-6}$ and $5 \times 10^{-7}$ at 40, 80 and 120 epochs, respectively. We use the batch size of 4, input resolution of $512 \times 512$, and Adam optimizer for all of our experiments.

\noindent \textbf{Inference strategy.} When generating trimaps for a new video that is not present in the training and validation sets, we found that online fine-tuning with the user-annotated trimaps at keyframes drastically improves the performance of the network in our case. During online fine-tuning, we treat the user-annotated trimap as the ground truth and use the same random affine transform technique to generate ``fake'' video sequences. Subsequently, we fine-tune the network on these sequences for 500-800 iterations with a constant learning rate of $10^{-6}$. When there is more than one frame of user-defined trimaps, a bidirectional inference strategy is used to ensemble the prediction results. Please refer to our supplementary material for more details.

\vspace{-5pt}
\subsection{A New Video Matting Dataset}
\label{sec:dataset}
Lacking training data is a massive barrier to deep learning-based video matting methods. For instance, the most commonly used video matting dataset from videomatting.com~\cite{Erofeev2015} has only ten test clips and three clips with ground-truth mattes, which is not enough for network training. To this end, we propose our video matting dataset, \textbf{VideoMatting108}. We rely on green screen video footages to extract ground-truth alpha mattes. First, we collect 68 high-quality (1080p and 4K) footages from the Internet~\cite{Storyblocks}. While these footages have diverse objects, we found that they generally lack several types of objects, such as fur, hair, and semi-transparent objects. Thus, we capture 40 green screen footages for these types of objects ourselves as the supplement. Next, we carefully extract the foreground object's alpha matte and color from the green screen footages using After Effects and BorisFX Primatte Studio~\cite{Primatte_Studio}.

In total, our dataset consists of 108 video clips, all in 1080p resolution. The average length of the video clip is 821 frames, significantly longer than other datasets. The foreground objects cover a wide range of categories, such as human, fluffy toys, cloth (net, lace, and chiffon), smoke and plants. The background footage usually has 3D camera motion or complex scenery, adding more challenge to our dataset. We split the dataset with 80 clips in the training set and 28 clips in the validation set. Trimaps are generated and dilated on the fly with random sized kernel from $1\times1$ to $51\times51$ during training. In the validation set, trimaps are generated by dilating transparent pixels with three different kernel sizes: $11\times11$ for narrow trimaps, $25\times25$ for medium trimaps, and $41\times41$ for wide trimaps.

%% file: 04_experiments.tex
\section{Experimental Results}
\label{sec:experiment}
In this section, we present the evaluation of our approach on the VideoMatting108 dataset. We also evaluate our trimap generation algorithm. Our computing platform for video matting related experiments was a 4 V100 GPU server with 2 Intel 6148 CPUs. Trimap generation related experiments were conducted on a 4 1080Ti GPU server with 2 E5-2678v3 CPUs.

\begin{table}[t]
    \caption{Comparison between GCA+TAM and GCA~\cite{li2020natural} on the 10 test clips from videomatting.com~\cite{Erofeev2015} with different trimaps. The best result is in bold. Please see our supplementary material for quantitative results of each test video clip.}
    \vspace{-10pt}
    \label{tab:videomatting.com}
    \small
    \centering
    \begin{tabularx}{\linewidth}{l|c|YYY}
        \hline
        Method & Trimap
        & SSDA & dtSSD & MESSDdt
        \\
        \hline
        GCA+F & \multirow{2}{*}{Narrow}
        &	39.40	&	30.83	&	1.43 \\
        GCA+TAM & 
        &	\textbf{36.95}	&	\textbf{26.37}	&	\textbf{1.12}  \\
        \hline
        GCA+F & \multirow{2}{*}{Medium}
        &	44.74	&	33.42	&	1.74 \\
        GCA+TAM &
        &	\textbf{42.17}	&	\textbf{28.81}	&	\textbf{1.35}  \\
        \hline
        GCA+F & \multirow{2}{*}{Wide}
        &	50.45	&	36.71	&	2.14 \\
        GCA+TAM &
        &	\textbf{49.23}	&	\textbf{32.94}	&	\textbf{1.76}  \\
        \hline
    \end{tabularx}
    \vspace{-13pt}
\end{table}

\noindent \textbf{Evaluation metrics.} We employ SSDA (average sum of squared difference) and two temporal coherence metrics, namely dtSSD (mean squared difference of direct temporal gradients) and MESSDdt (mean squared difference between the warped temporal gradient) from~\cite{Erofeev2015} to evaluate the accuracy of the predicted video alpha mattes and their temporal coherence. Besides, we also report ``MSE'' (mean squared error) and ``SAD'' (sum of absolute difference) to verify the pixel-wise accuracy of alpha values at each frame. Lower evaluation metrics correspond to better video matting results.

\begin{figure*}[ht]
\begin{center}
    \setlength{\tabcolsep}{1pt}
    \resizebox{0.925\linewidth}{!}{
        \begin{tabular}{cccccccccc}
        \rotatebox{90}{\scriptsize{Frames}} &
		\includegraphics[width=0.15\textwidth]{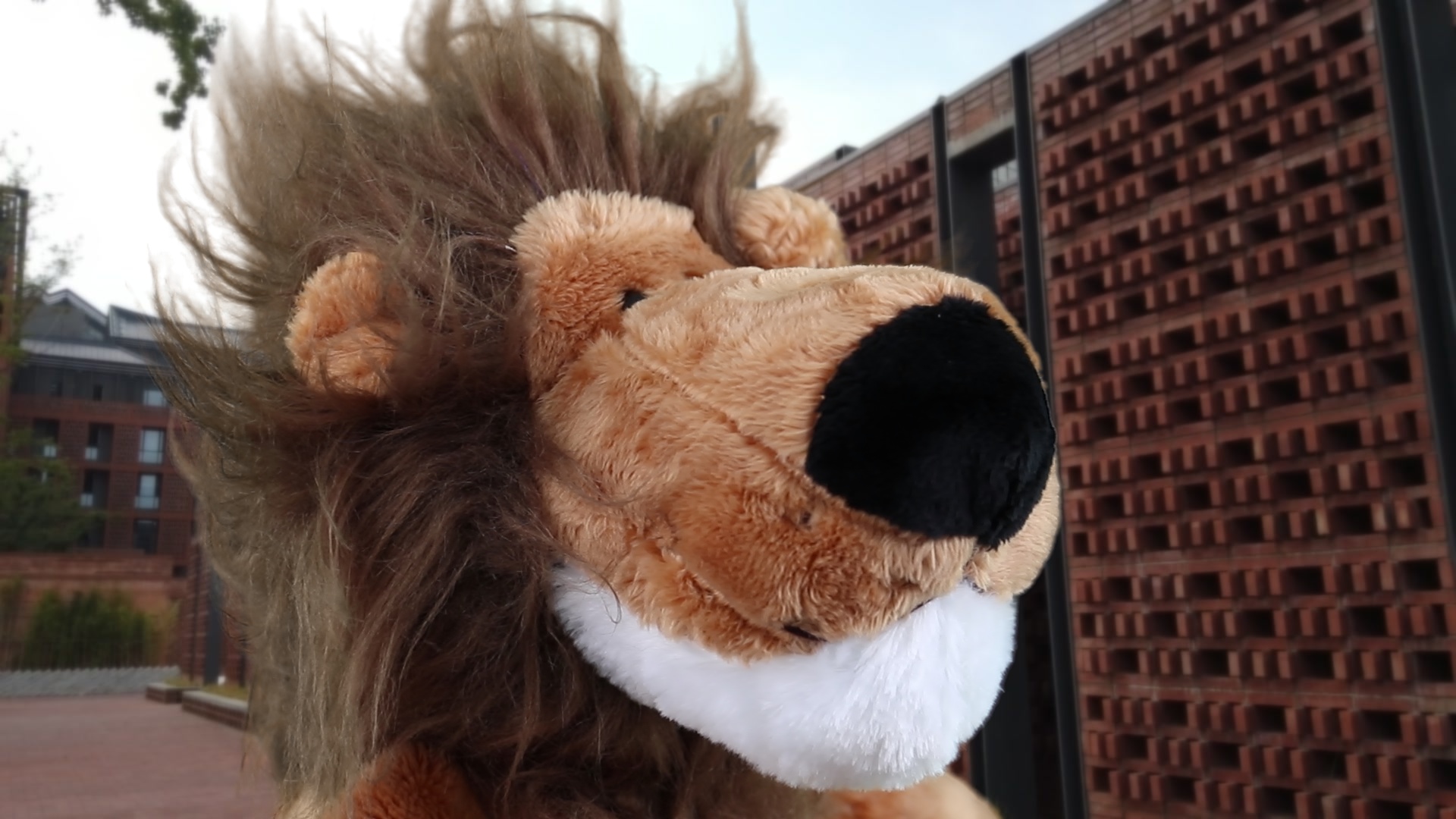} &
        \includegraphics[width=0.15\textwidth]{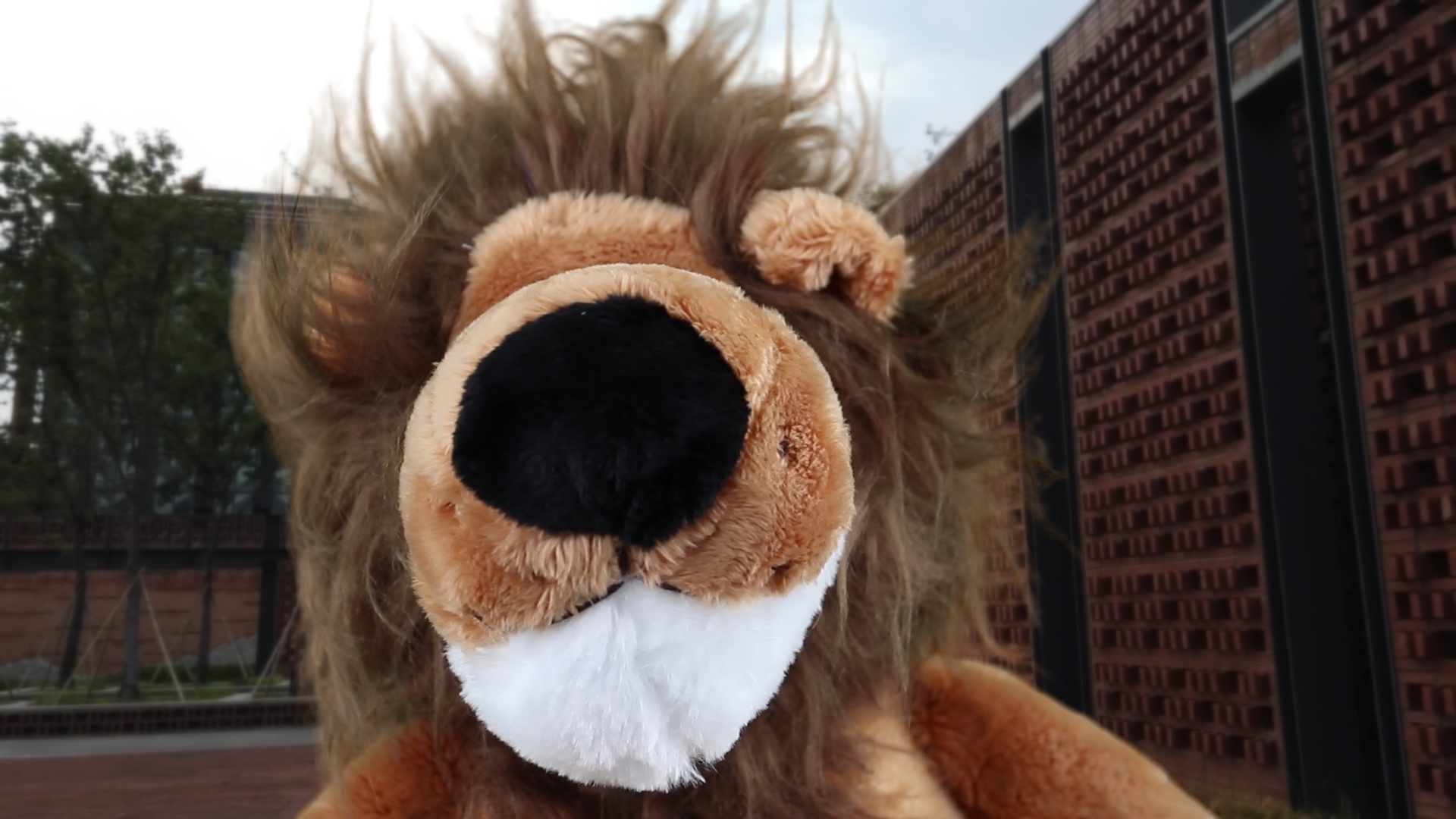} &
        \rotatebox{90}{\scriptsize{Frames}} &
        \includegraphics[width=0.15\textwidth]{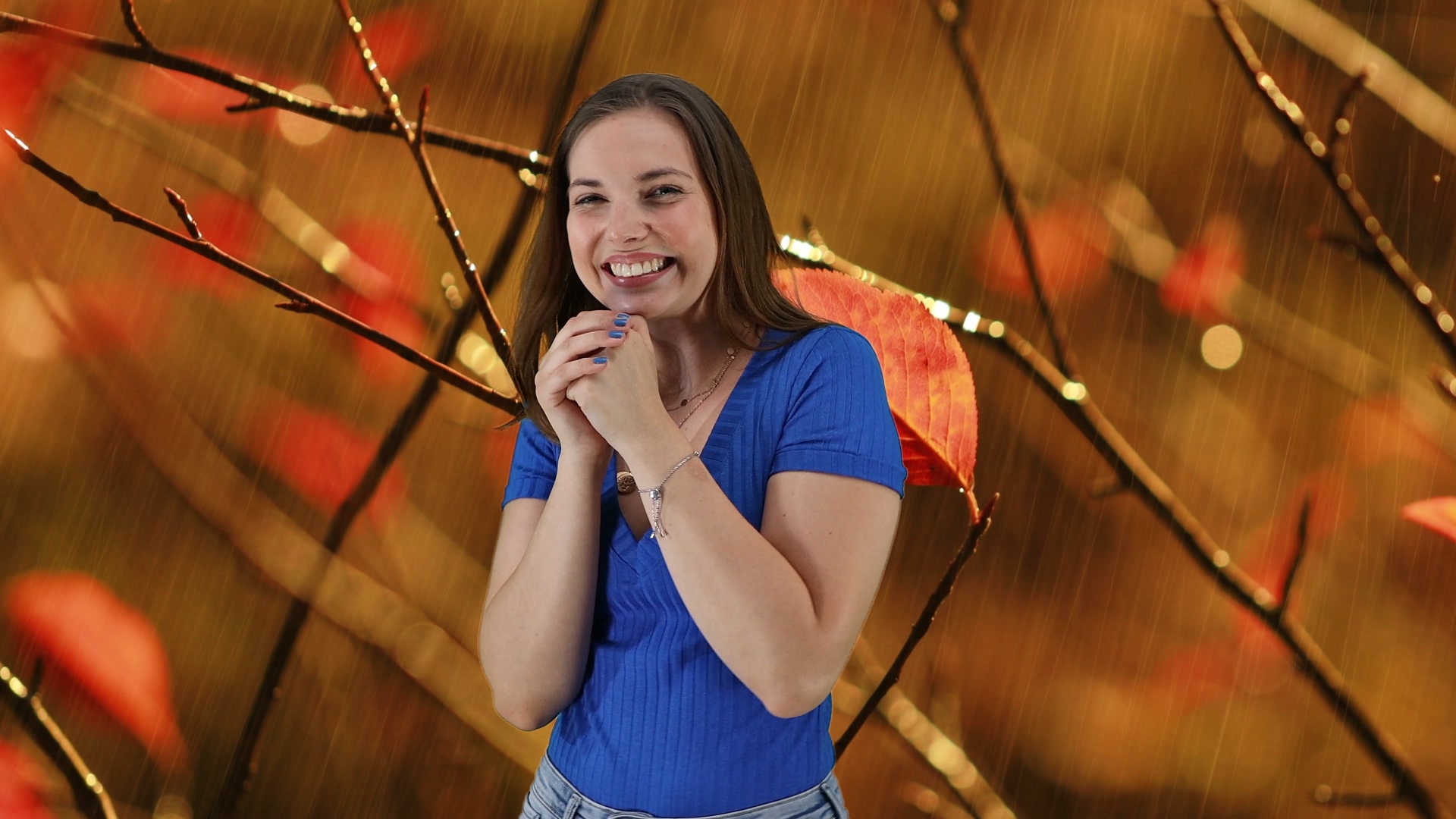} &
        \includegraphics[width=0.15\textwidth]{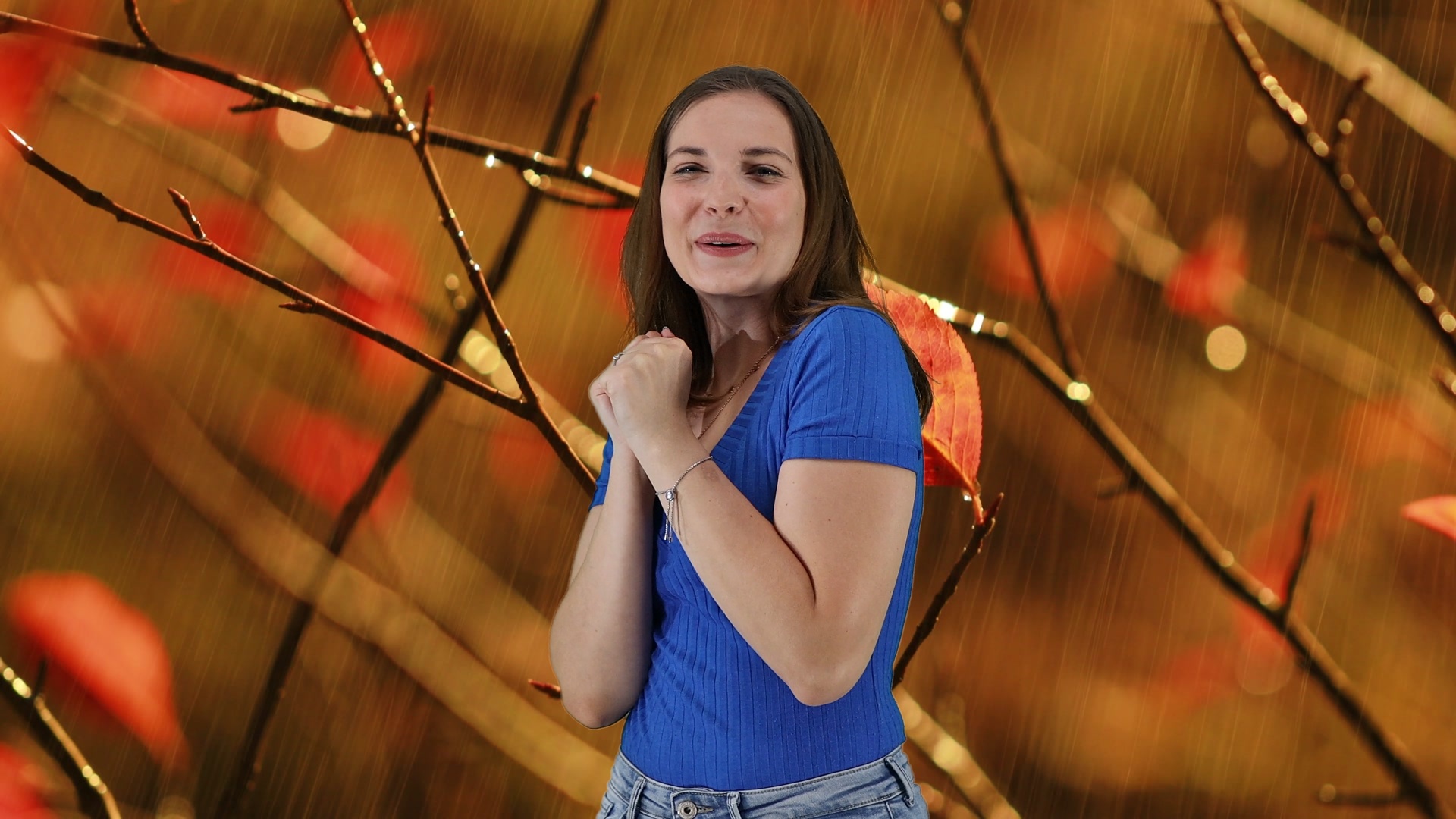} &
		\rotatebox{90}{\scriptsize{Frames}} &
		\includegraphics[width=0.15\textwidth]{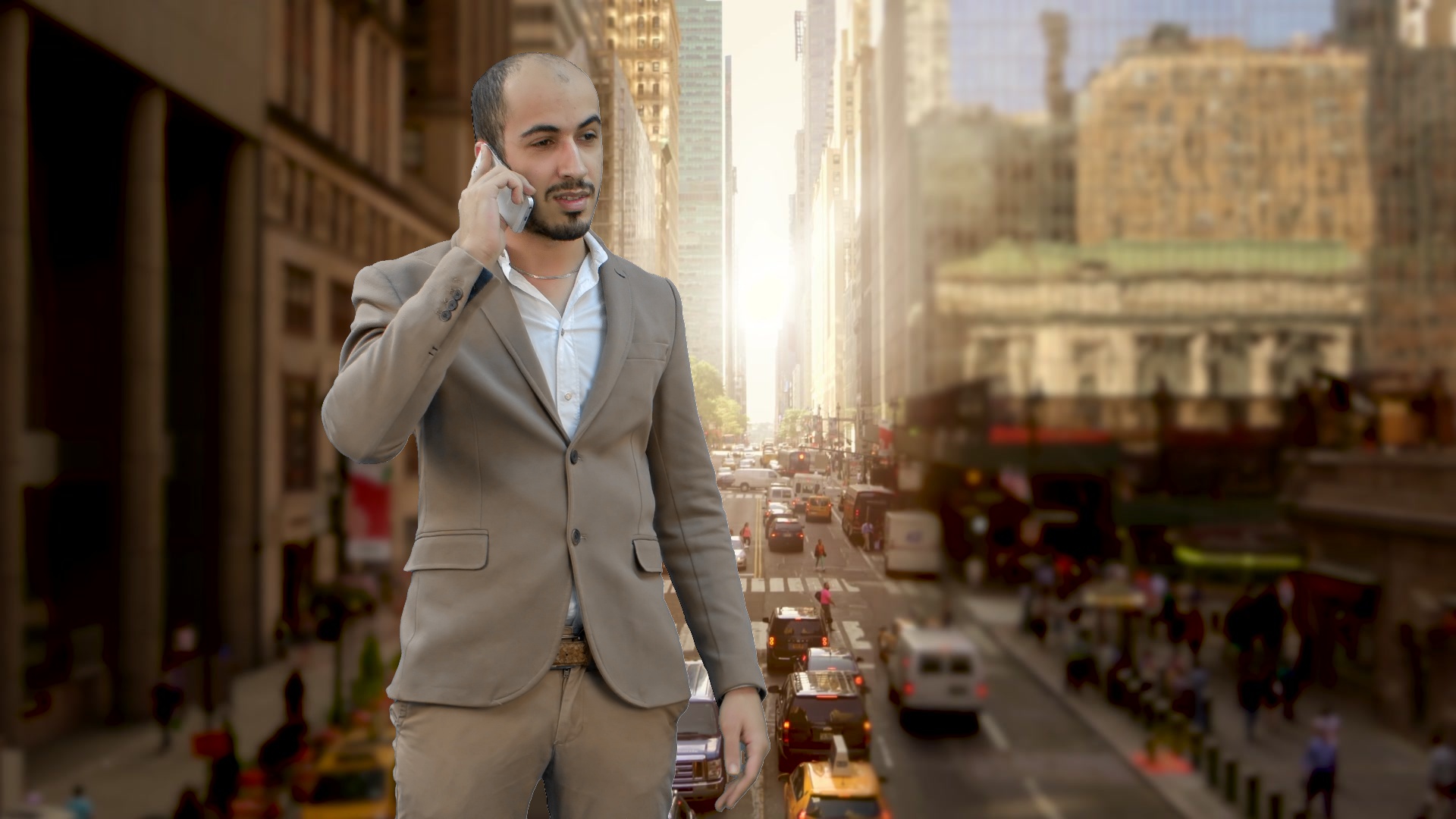} &
        \includegraphics[width=0.15\textwidth]{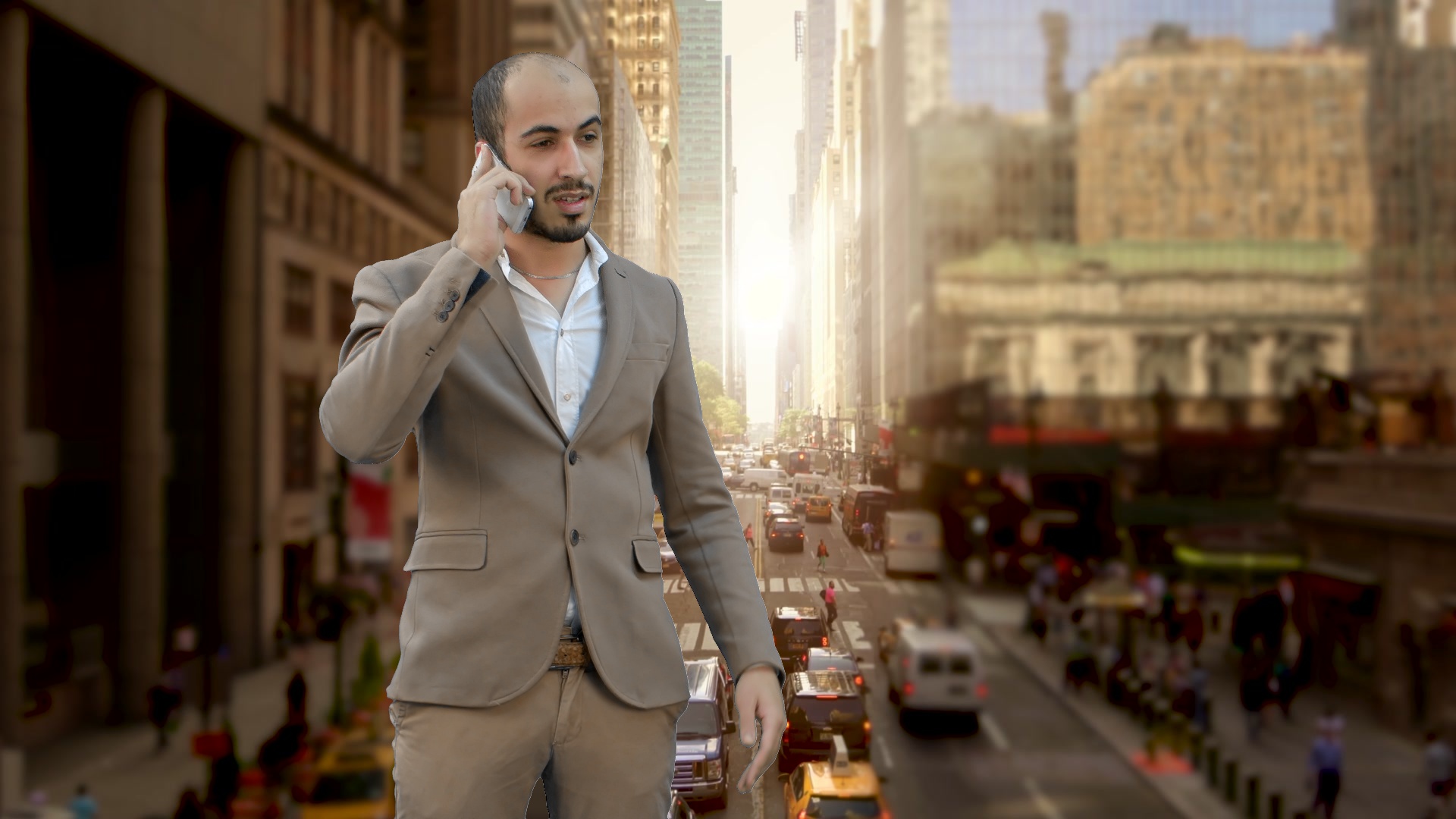} \\
		
        \rotatebox{90}{\scriptsize{GT}} &
		\includegraphics[width=0.15\textwidth]{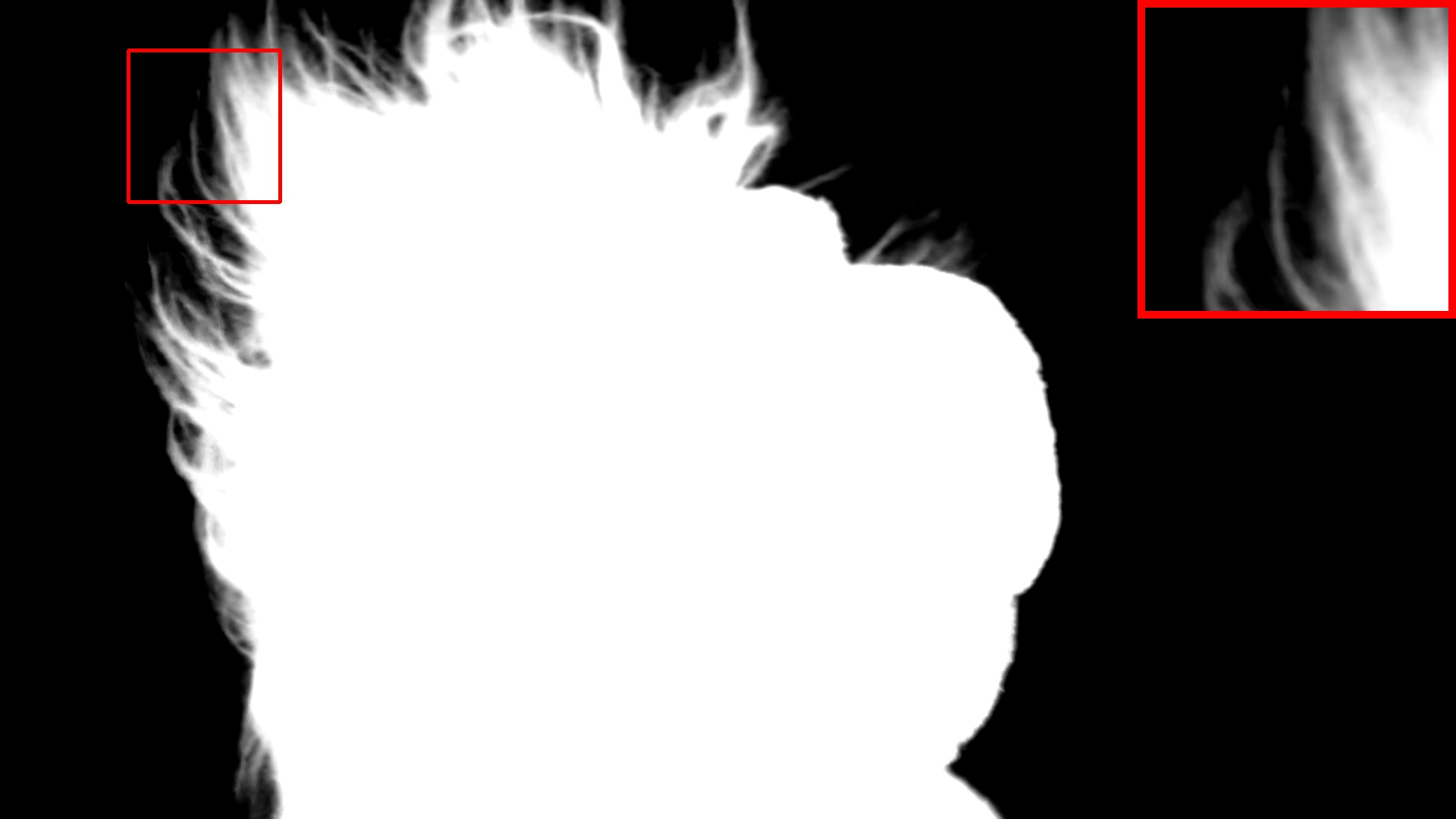} &
        \includegraphics[width=0.15\textwidth]{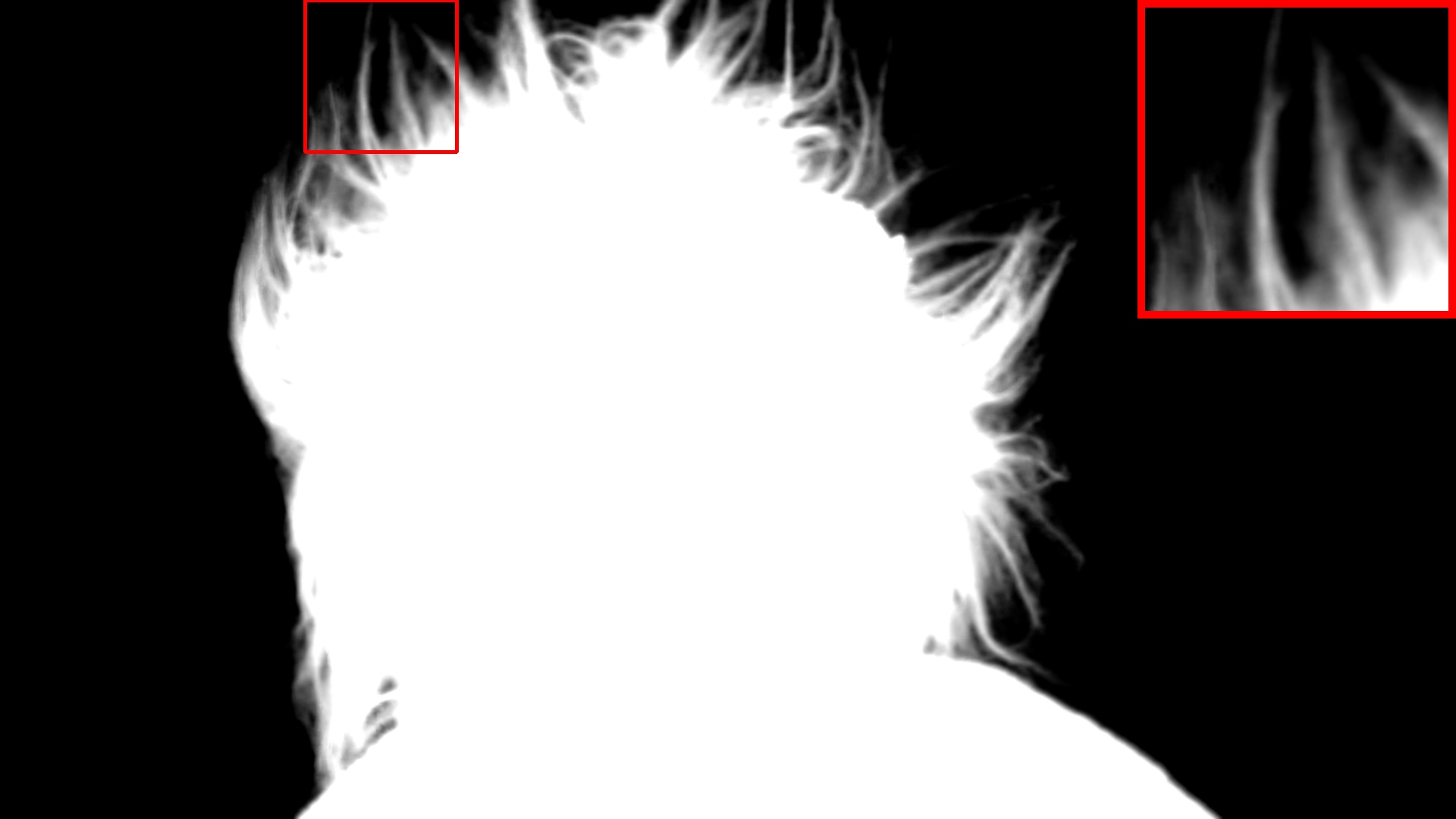} &
        \rotatebox{90}{\scriptsize{GT}} &
        \includegraphics[width=0.15\textwidth]{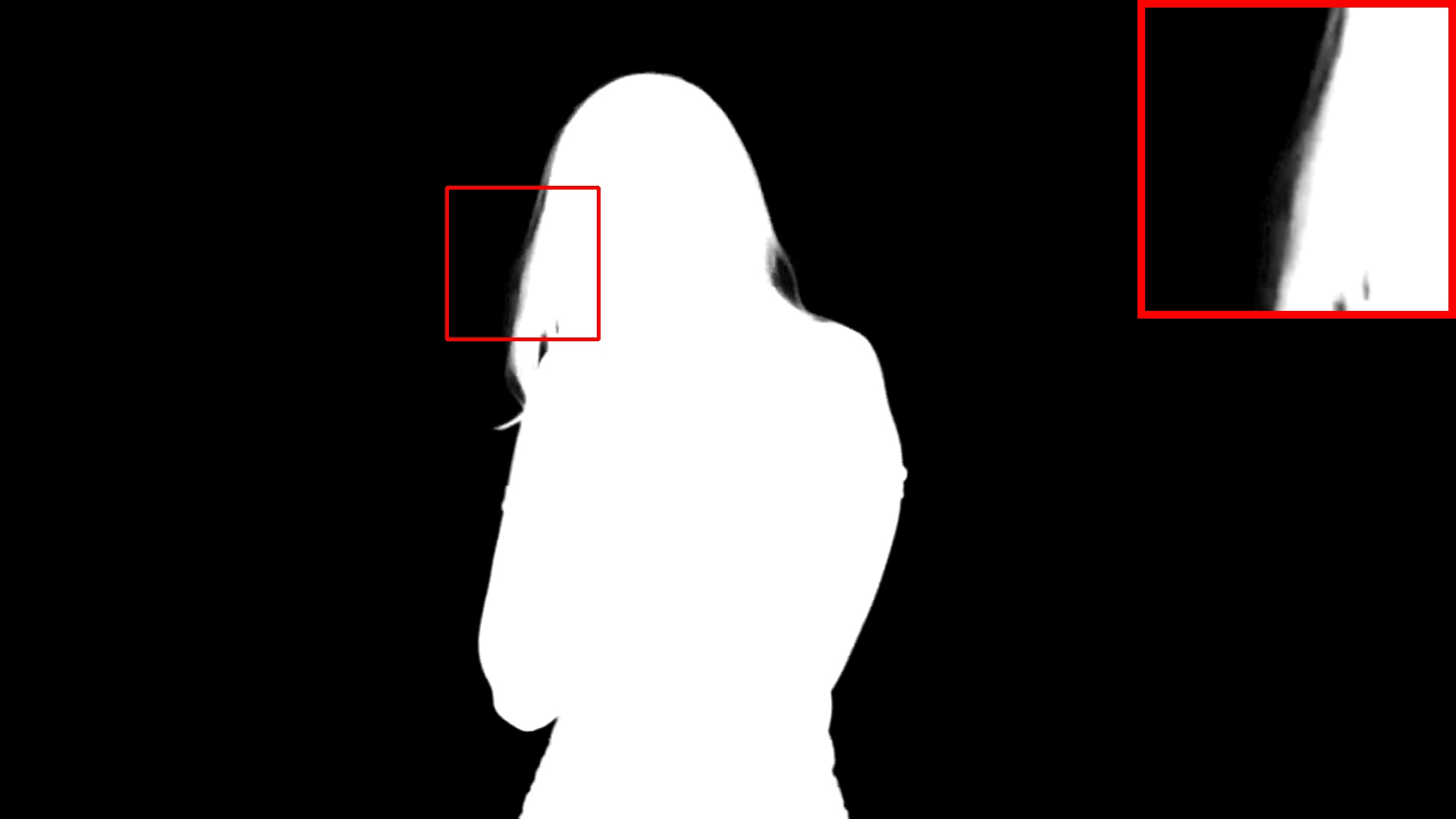} &
        \includegraphics[width=0.15\textwidth]{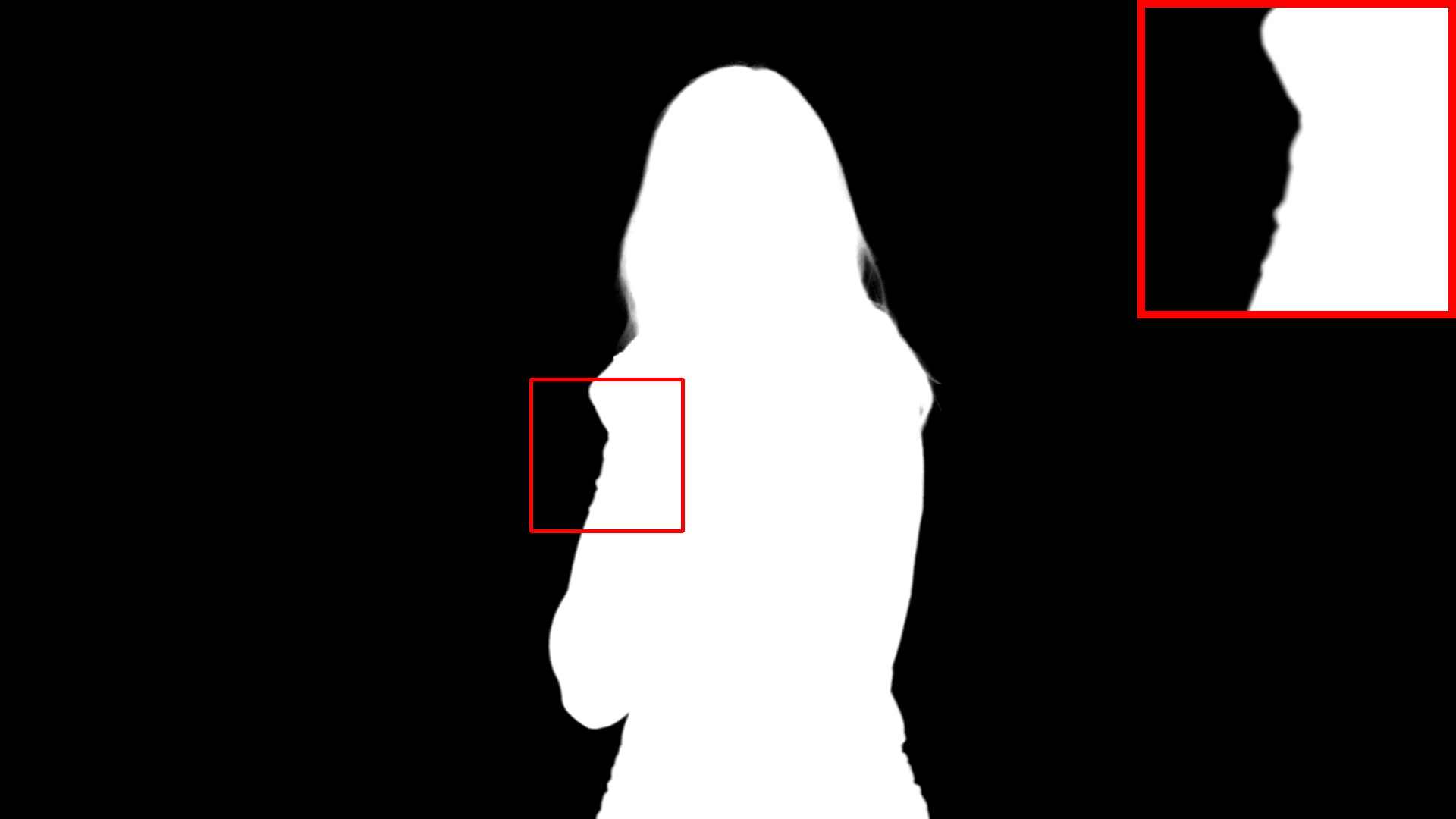} &
		\rotatebox{90}{\scriptsize{GT}} &
		\includegraphics[width=0.15\textwidth]{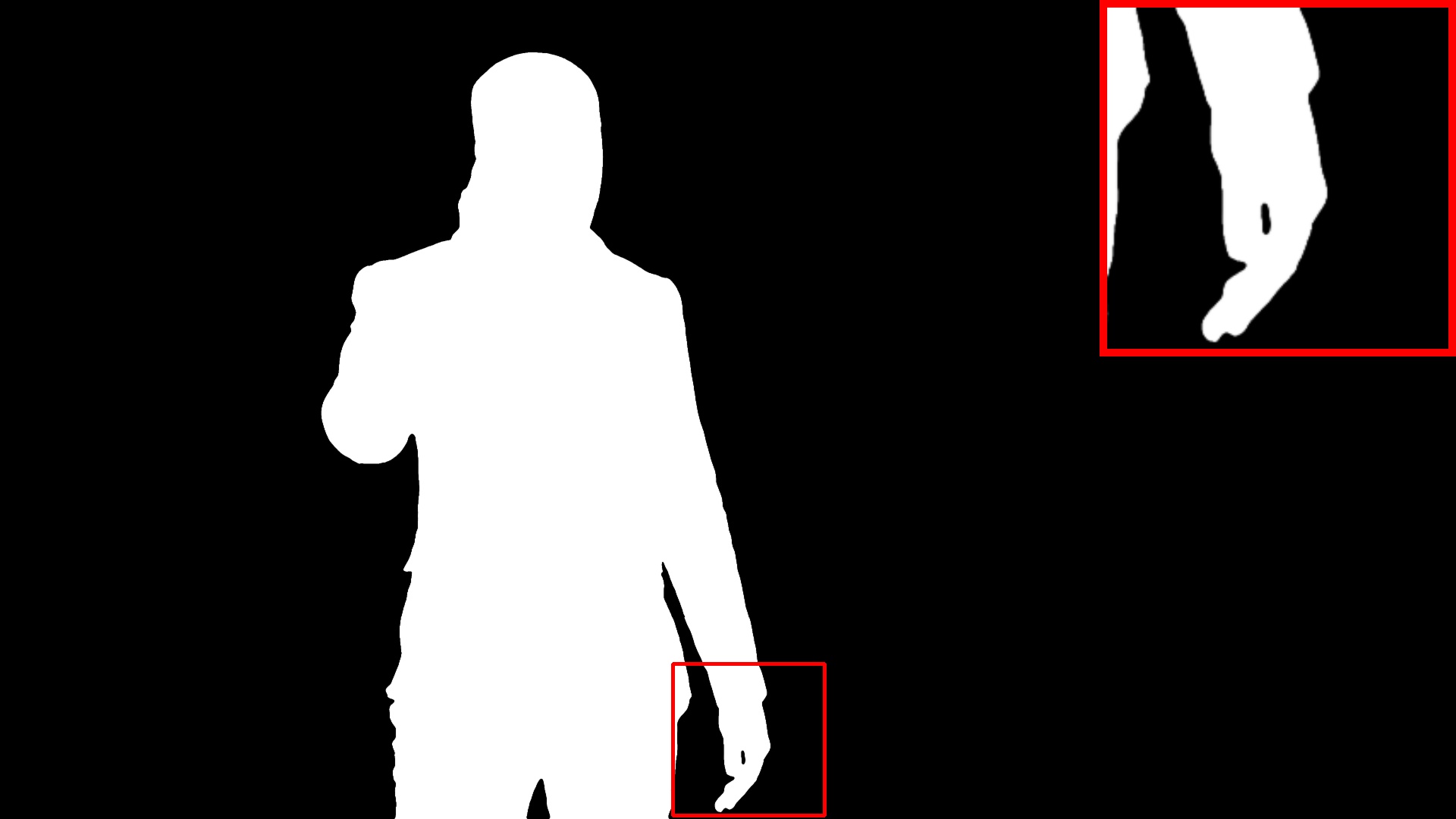} &
        \includegraphics[width=0.15\textwidth]{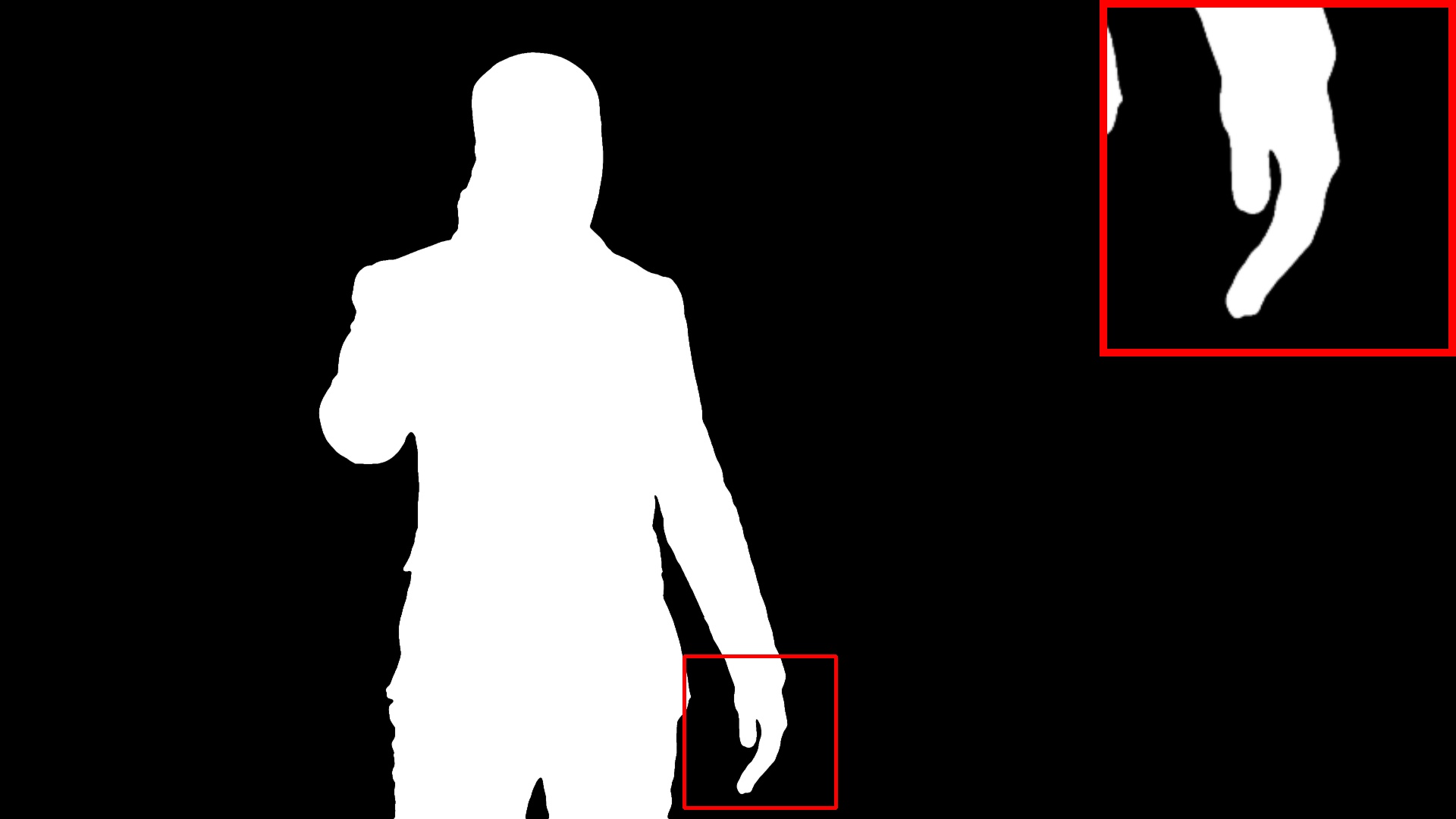} \\
        
        \rotatebox{90}{\scriptsize{GCA~\cite{li2020natural}}} & 
		\includegraphics[width=0.15\textwidth]{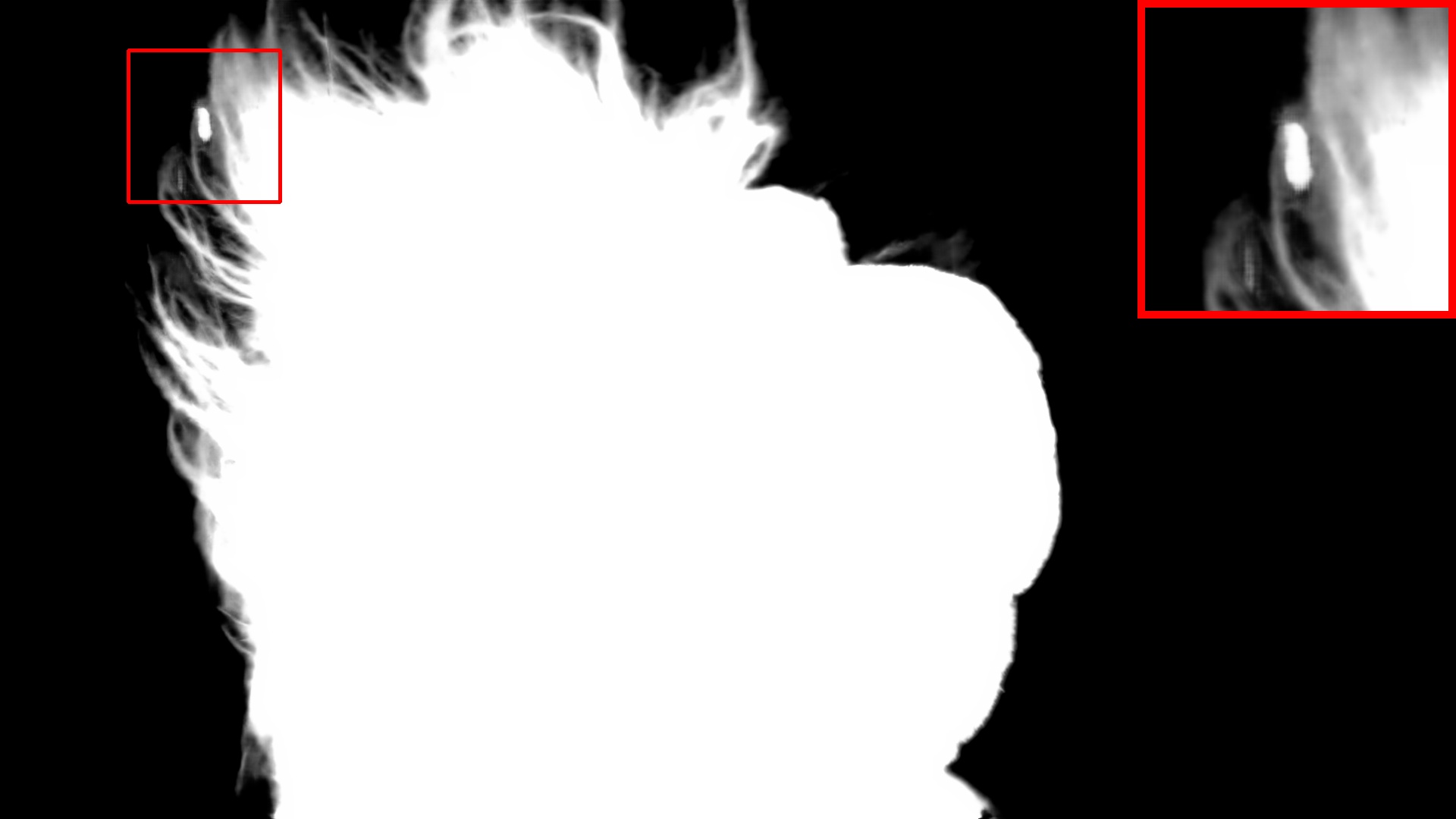} &
        \includegraphics[width=0.15\textwidth]{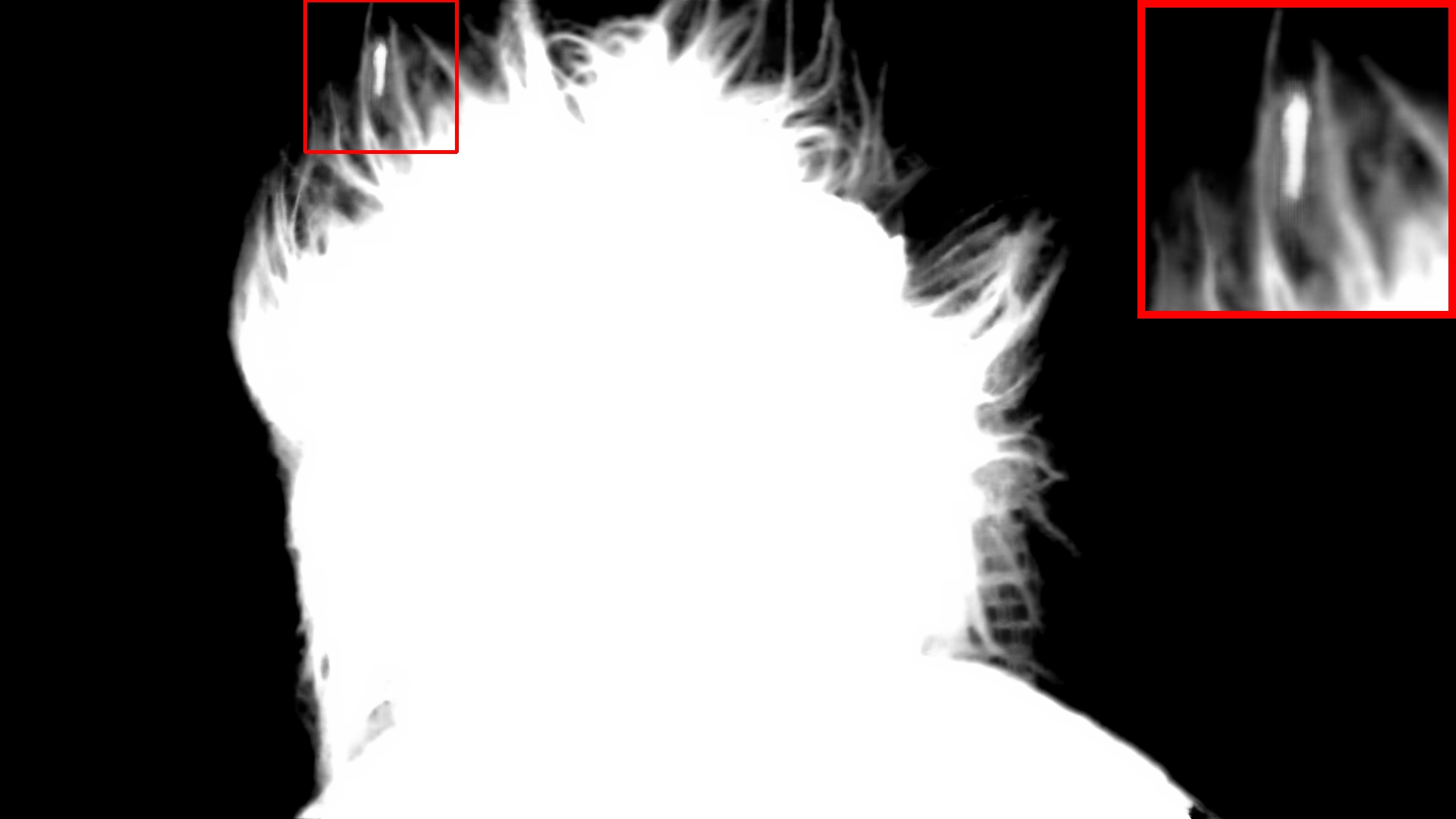} &
        \rotatebox{90}{\scriptsize{Index~\cite{Lu_2019_ICCV}}} &
        \includegraphics[width=0.15\textwidth]{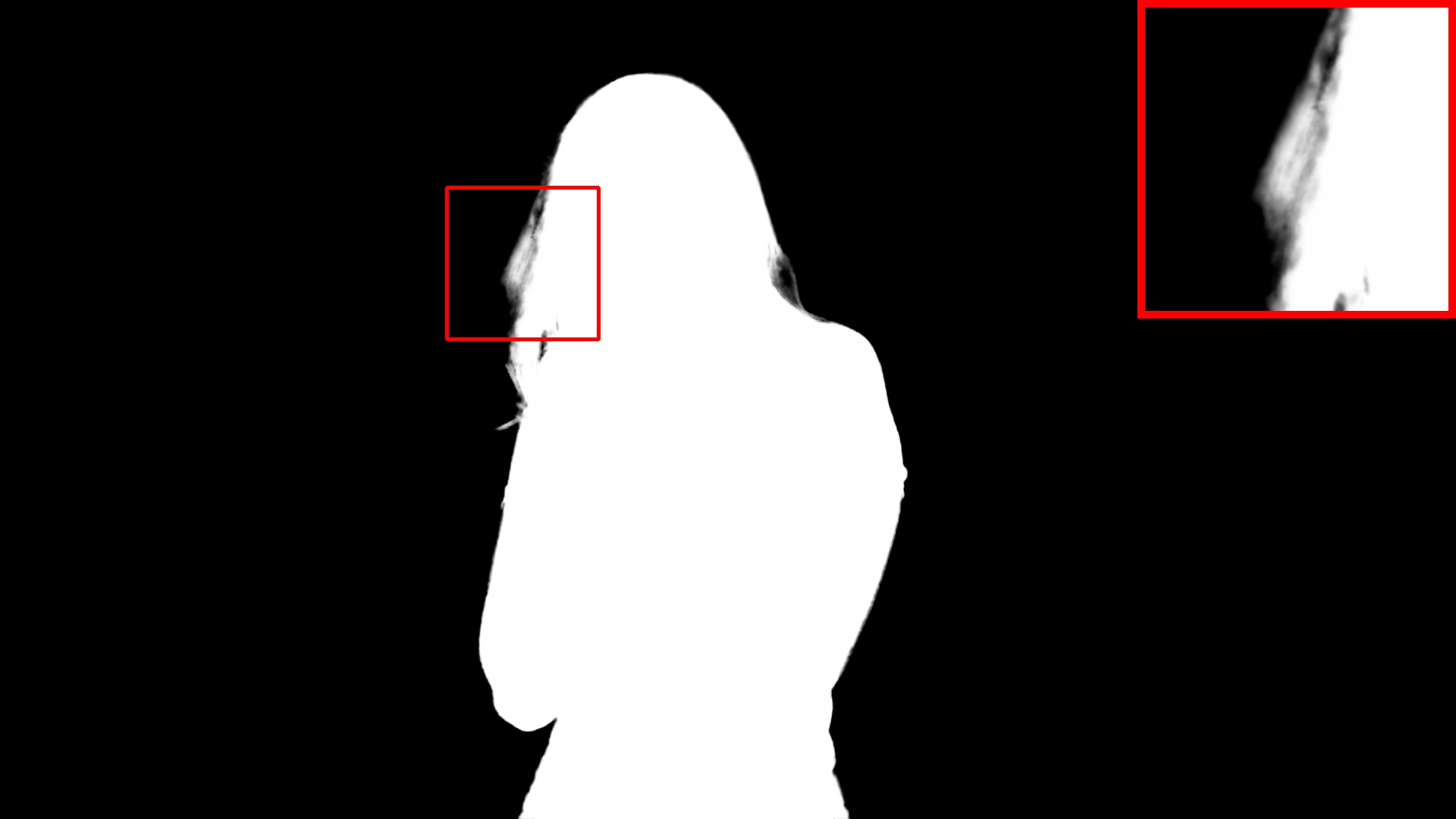} &
        \includegraphics[width=0.15\textwidth]{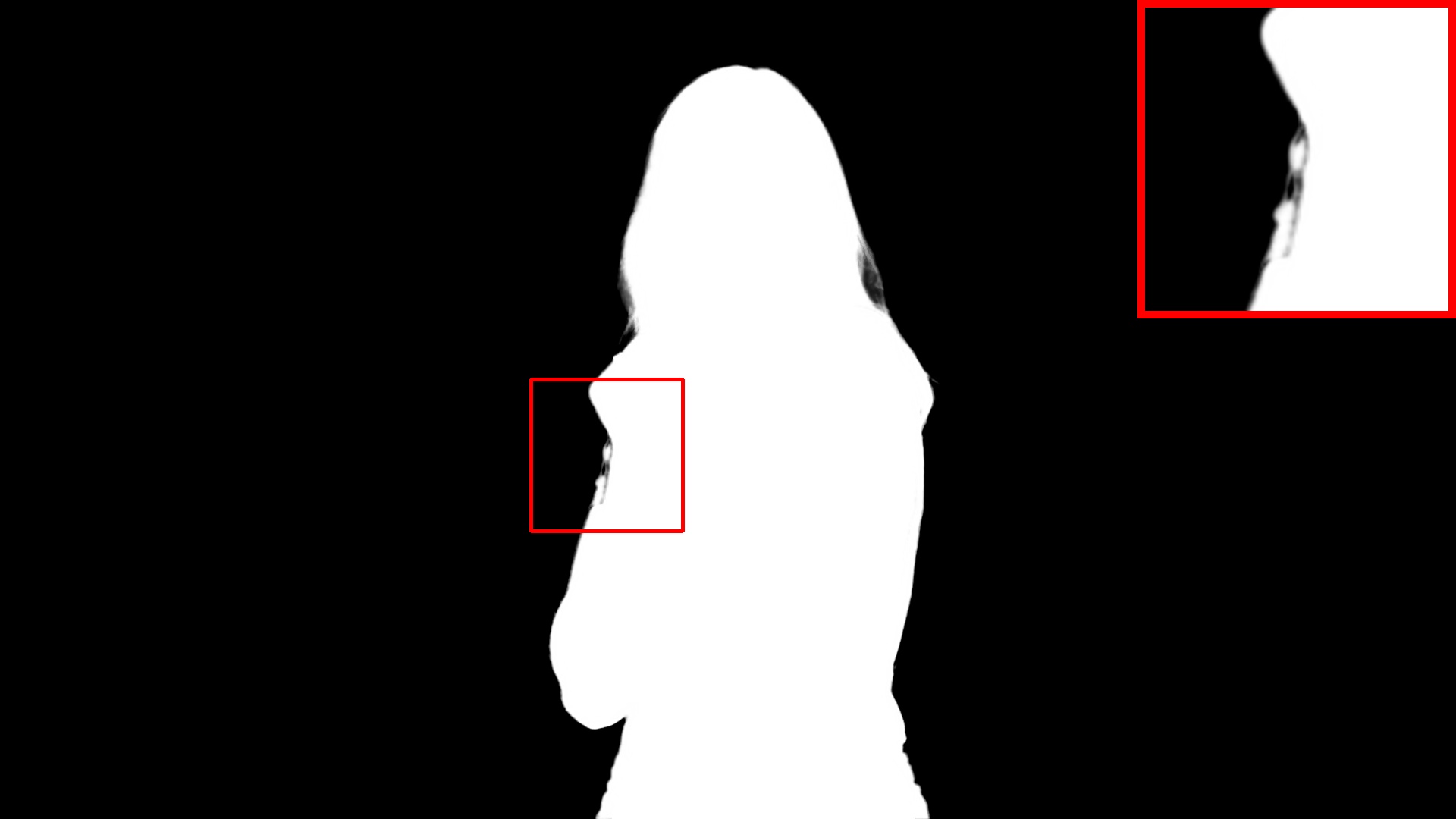} &
		\rotatebox{90}{\scriptsize{DIM~\cite{xu2017deep}}} & 
		\includegraphics[width=0.15\textwidth]{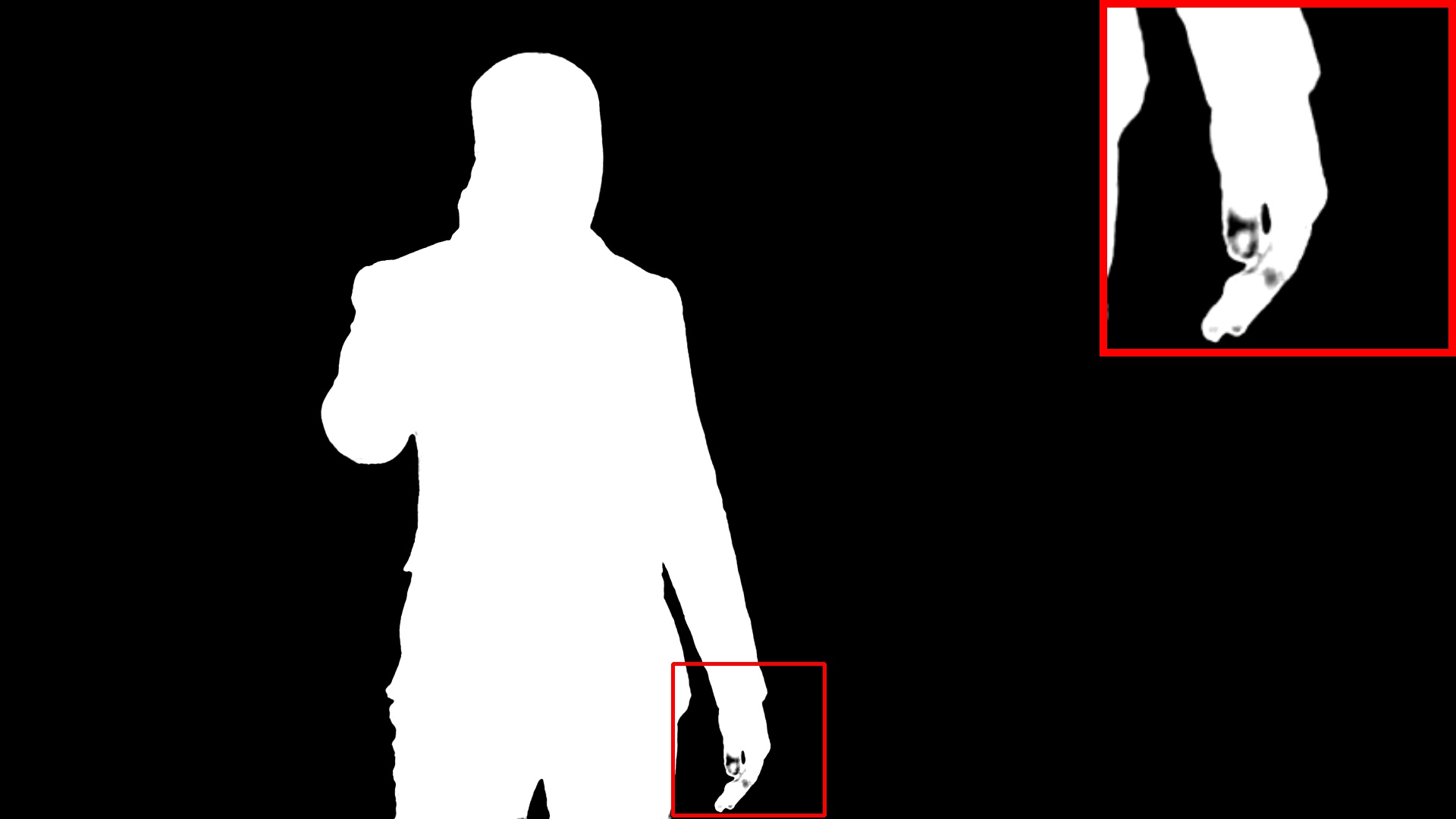} &
        \includegraphics[width=0.15\textwidth]{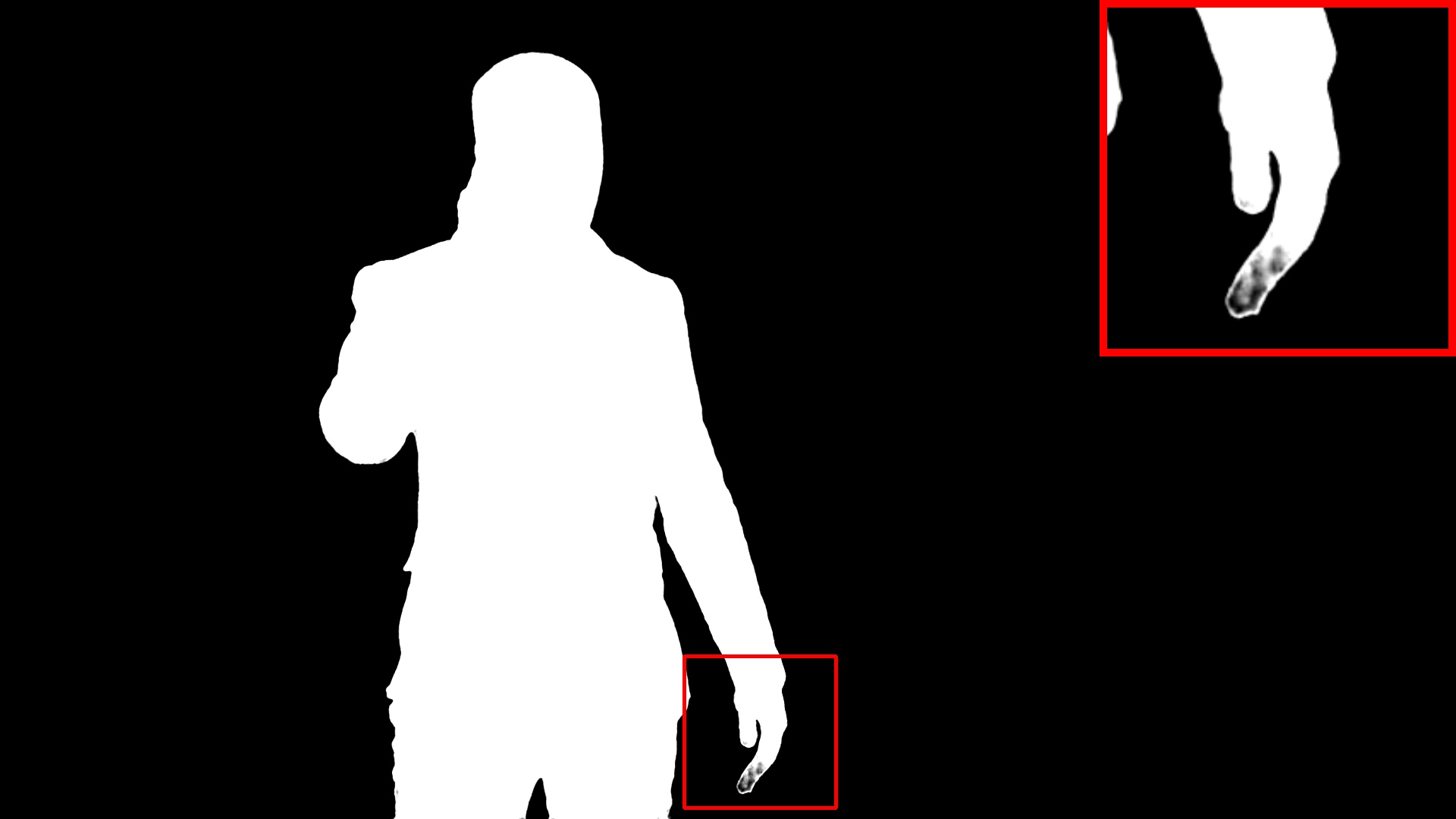} \\
        \rotatebox{90}{\scriptsize{GCA+TAM}} & 
        
		\includegraphics[width=0.15\textwidth]{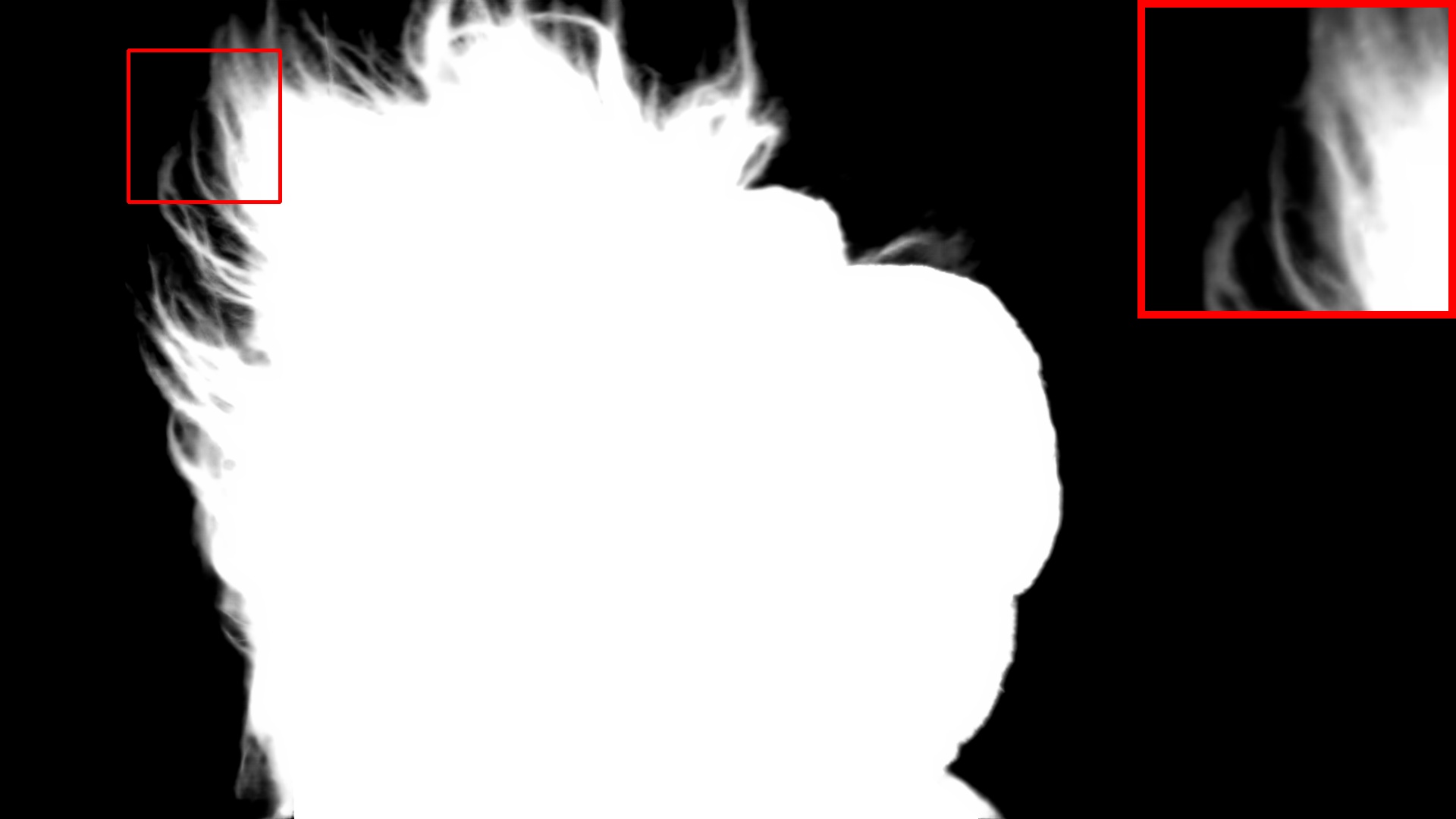} &
        \includegraphics[width=0.15\textwidth]{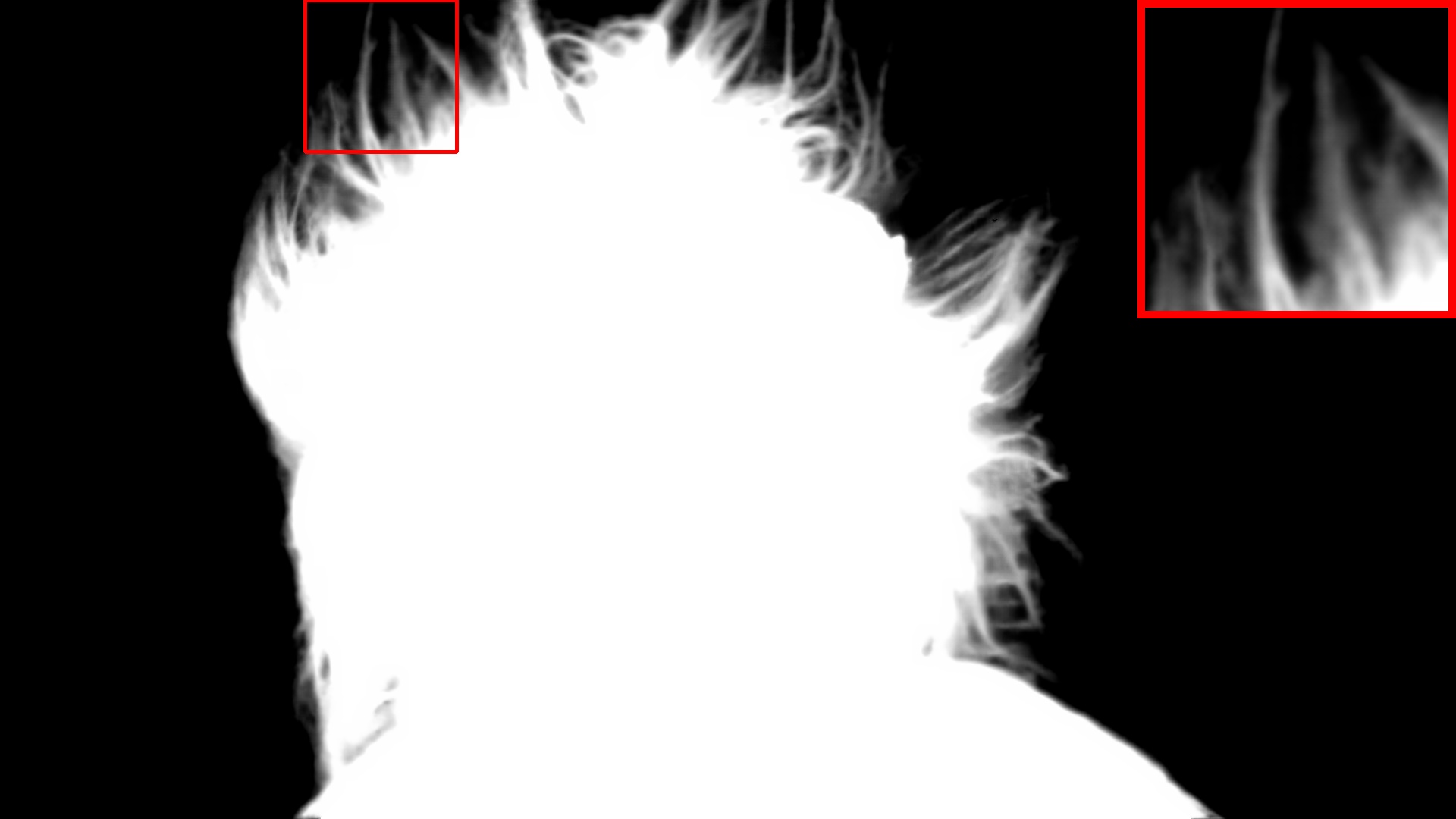} &
        \rotatebox{90}{\scriptsize{Index+TAM}} &
        \includegraphics[width=0.15\textwidth]{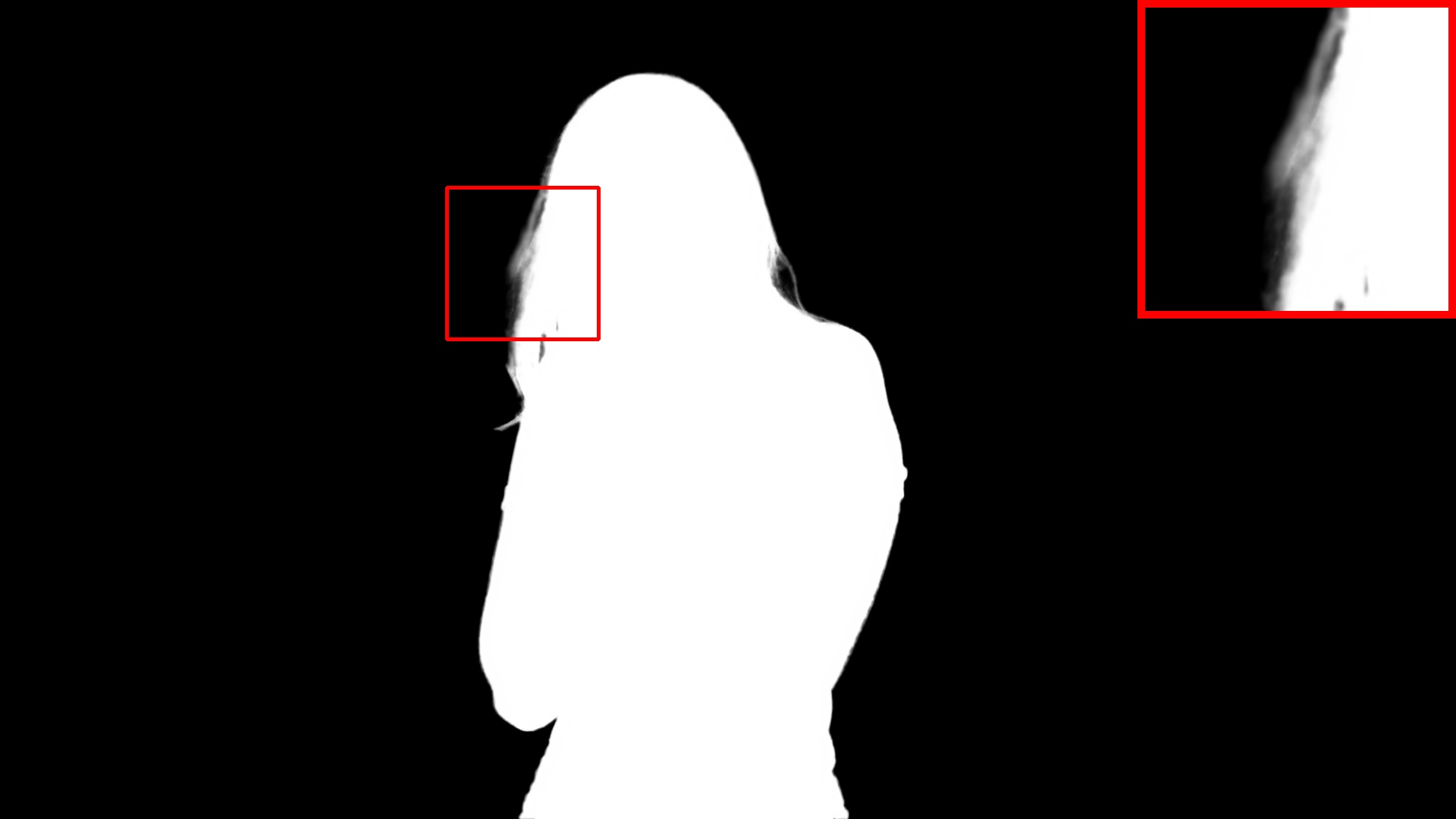} &
        \includegraphics[width=0.15\textwidth]{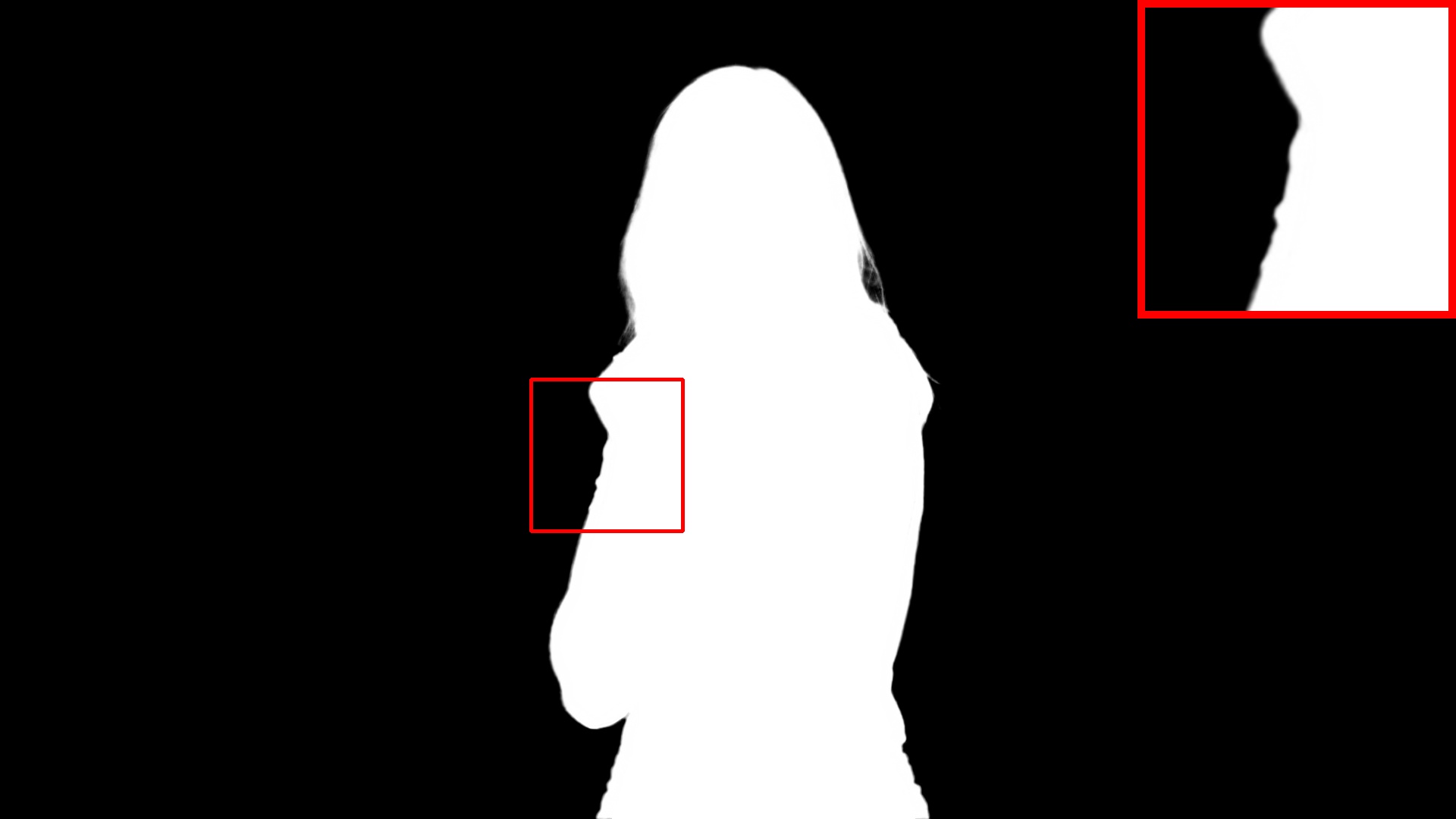} &
		\rotatebox{90}{\scriptsize{DIM+TAM}} & 
		\includegraphics[width=0.15\textwidth]{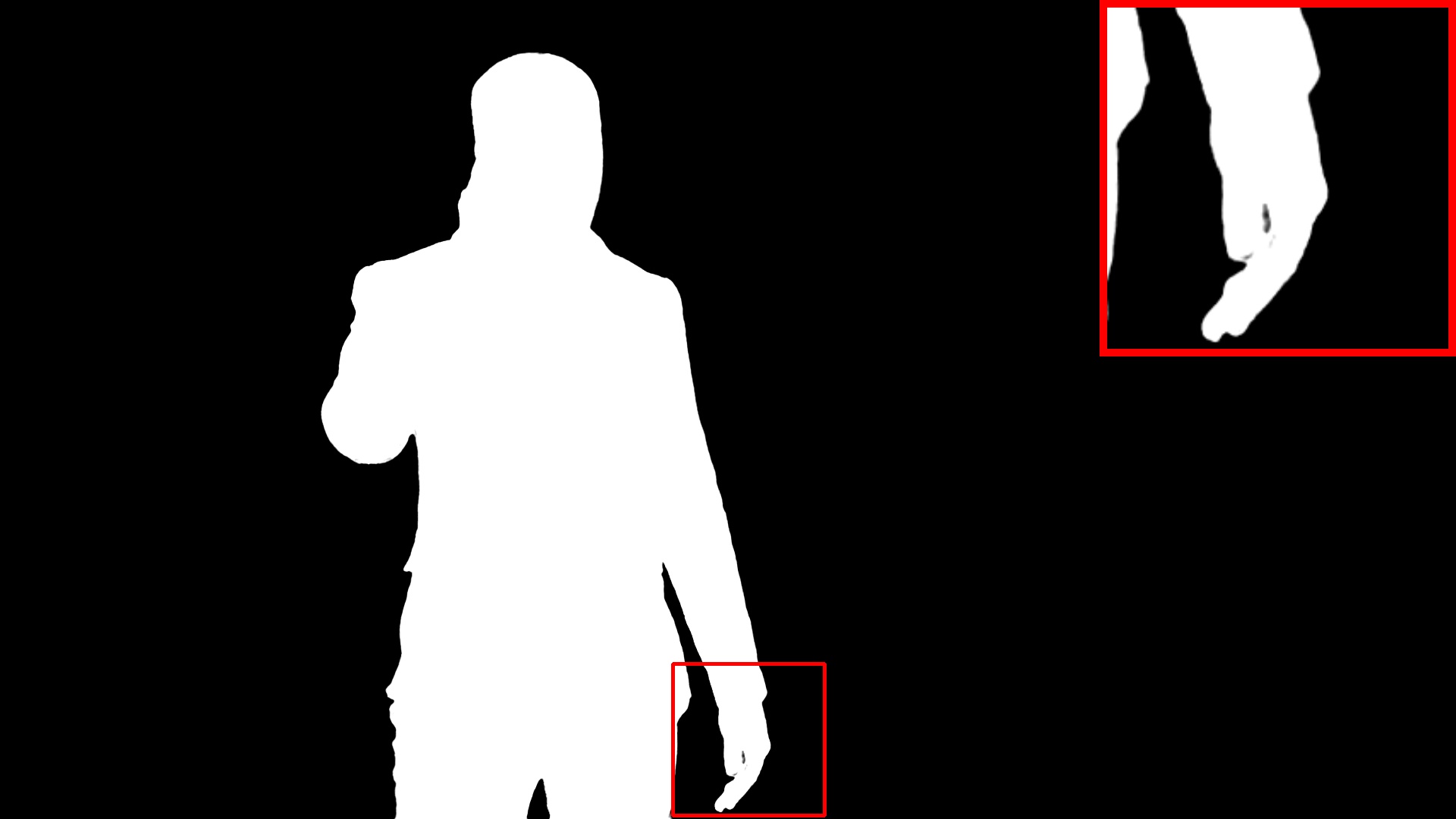} &
        \includegraphics[width=0.15\textwidth]{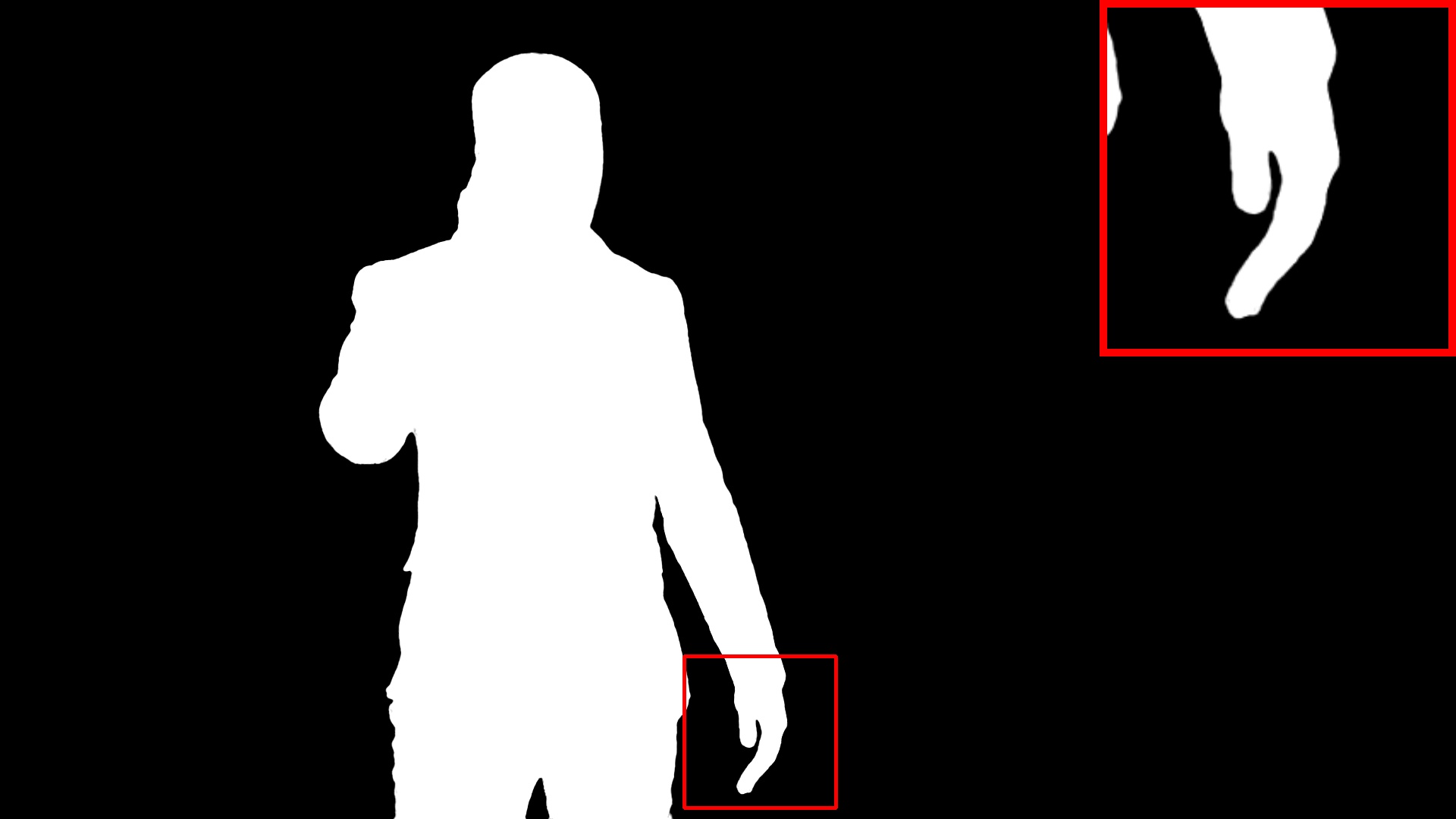} \\
        & \#149 & \#334 & & \#257 & \#382 & & \#380 & \#382 \\
        & \multicolumn{2}{c}{(a) lion} & & \multicolumn{2}{c}{(b) dancing woman} & & \multicolumn{2}{c}{(c) standing man} \\
        \end{tabular}
    }
\end{center}
\vspace{-10pt}
\caption{Qualitative evaluations that illustrate the effectiveness of our TAM. Blowups are used to show the details of the alpha matte. These three video clips are from VideoMatting108 validation set, and we use the ``medium''  ground-truth trimaps to obtain the results. Please see the supplementary video for the complete results.} 
\label{fig:val_matting}
\vspace{-10pt}
\end{figure*}

\begin{figure*}[ht]
\begin{center}
    \setlength{\tabcolsep}{1pt}
    \resizebox{0.925\linewidth}{!}{
        \begin{tabular}{ccccccc}
        \rotatebox{90}{\scriptsize{Frames}} &
        \includegraphics[width=0.15\textwidth]{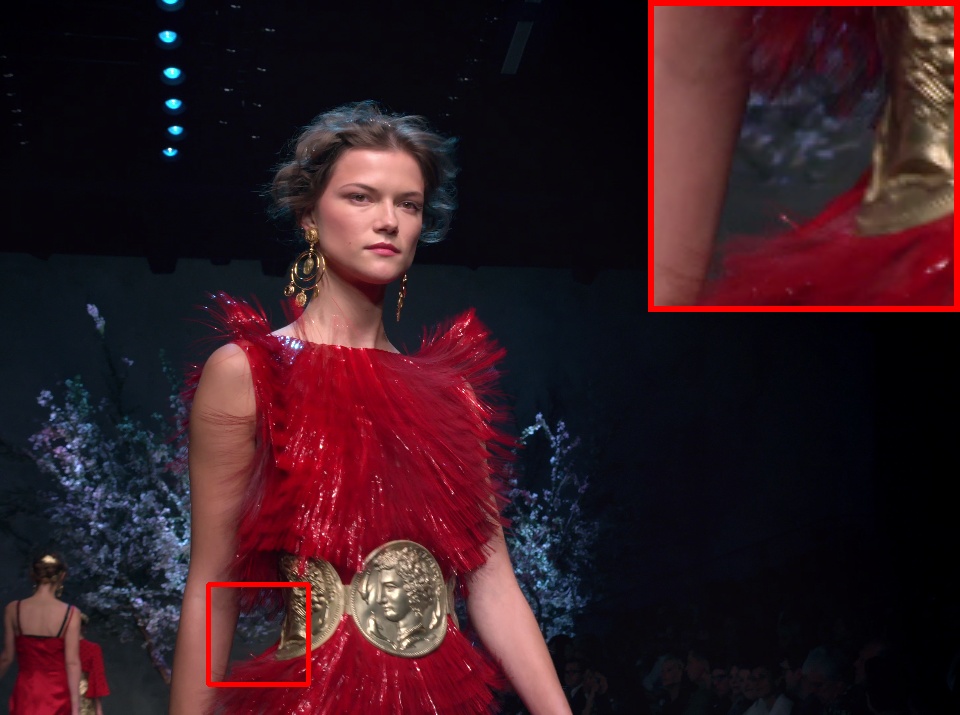} &
        \includegraphics[width=0.15\textwidth]{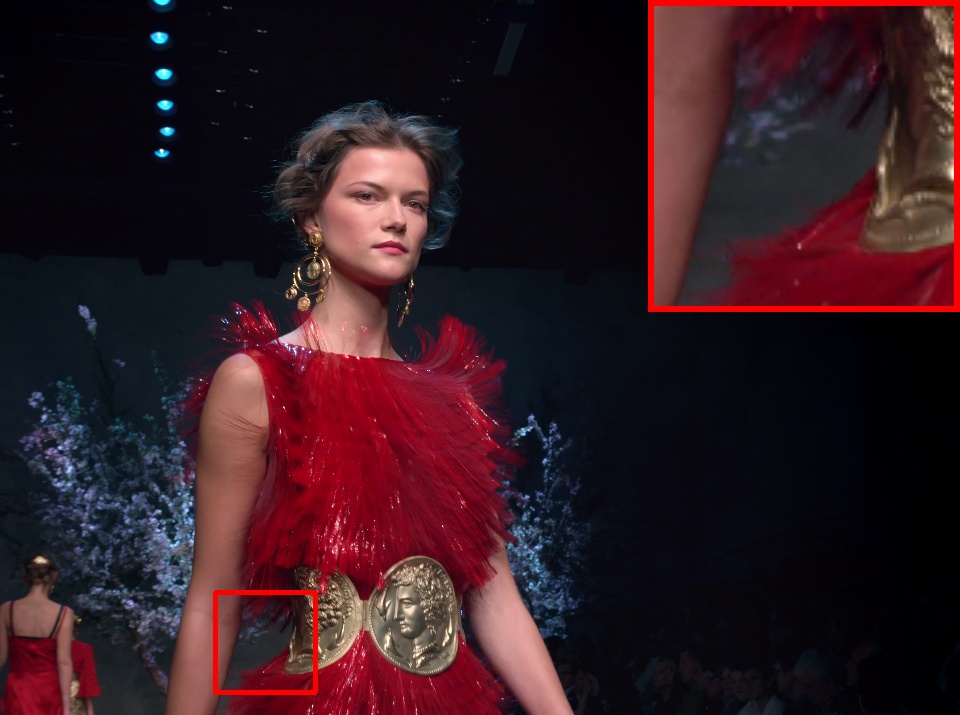} &
        \includegraphics[width=0.122\textwidth]{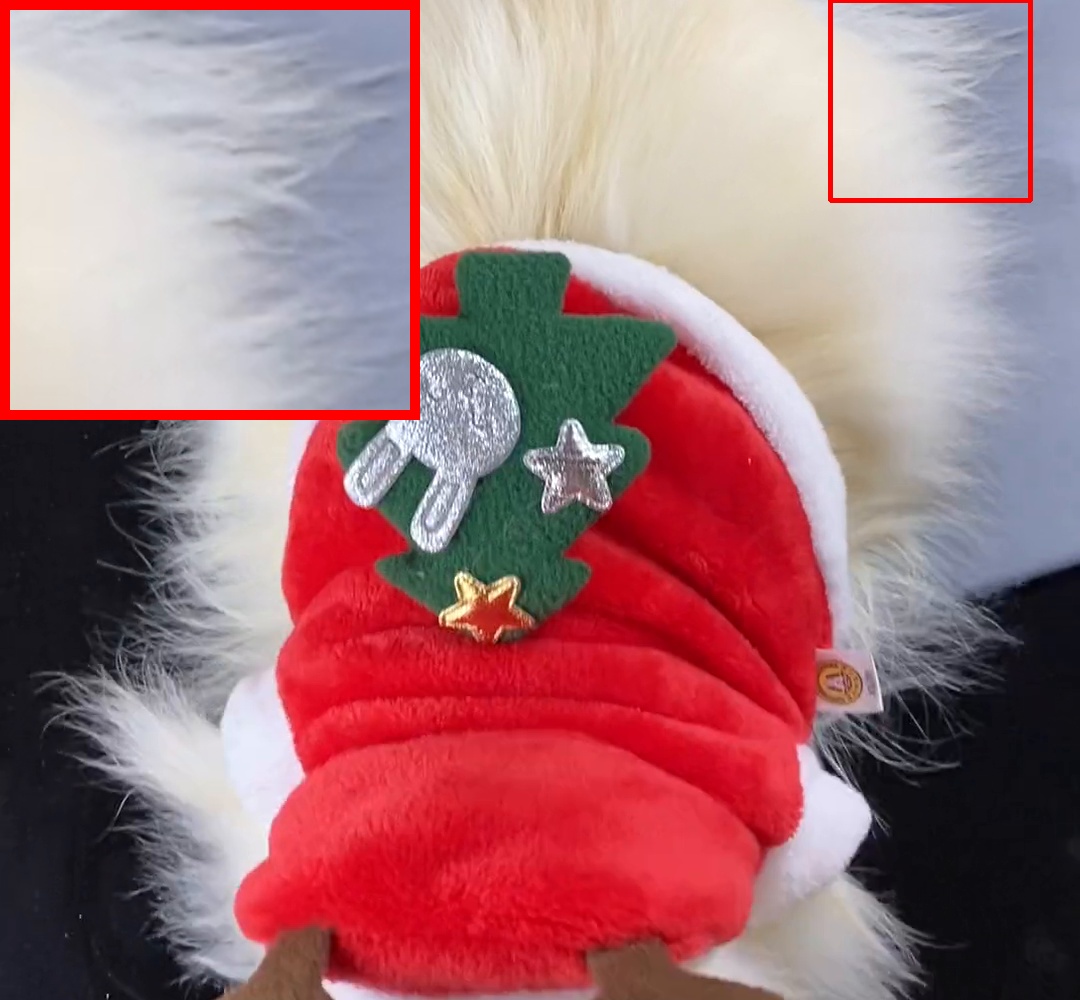} &
        \includegraphics[width=0.122\textwidth]{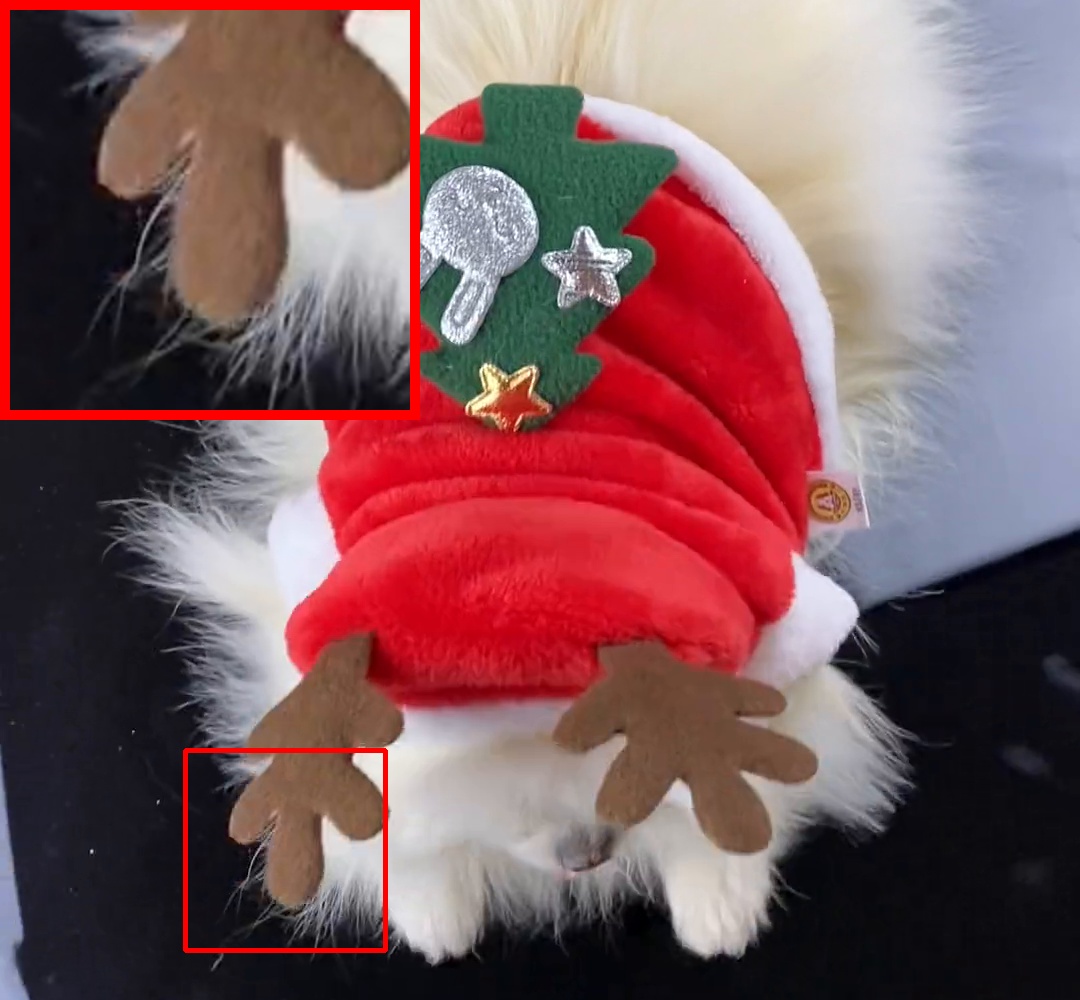} &
        \includegraphics[width=0.15\textwidth]{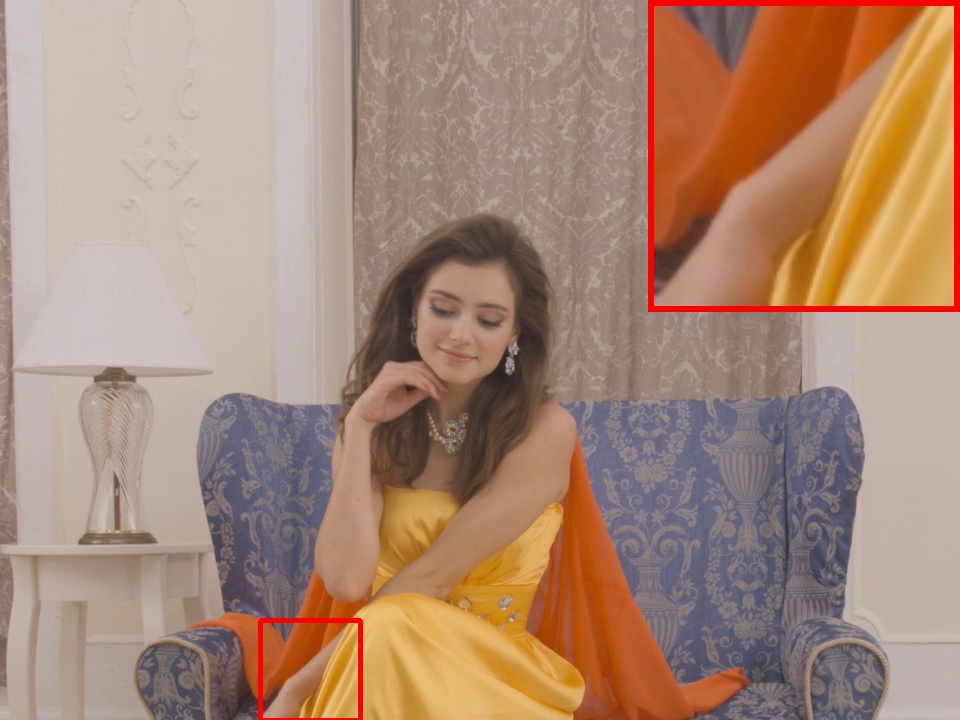} &
        \includegraphics[width=0.15\textwidth]{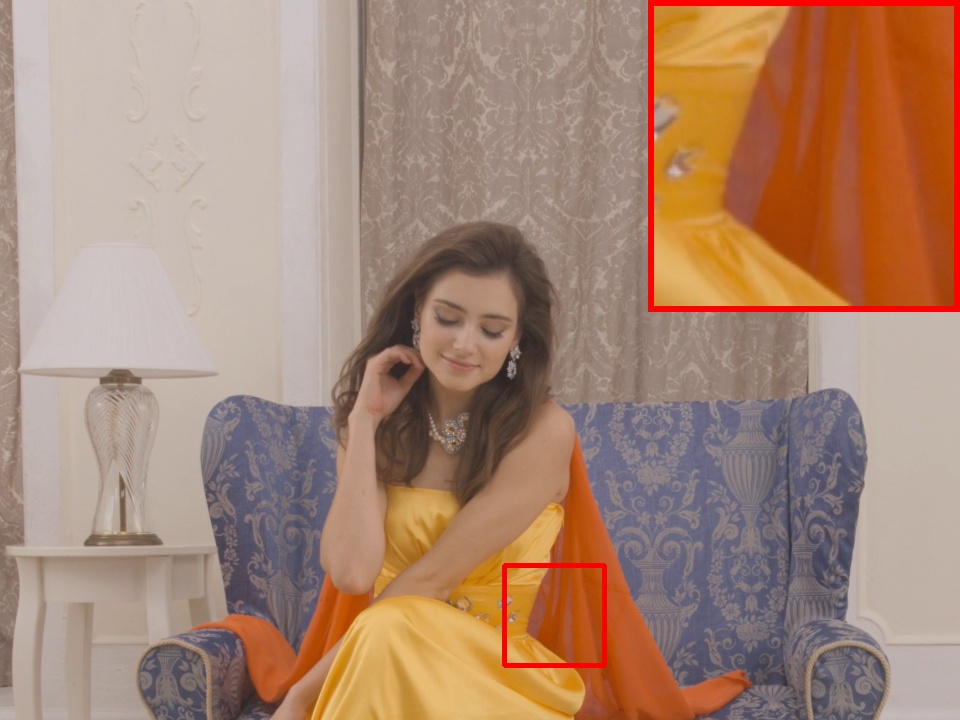} \\
        \rotatebox{90}{\scriptsize{Trimap}} &
        \includegraphics[width=0.15\textwidth]{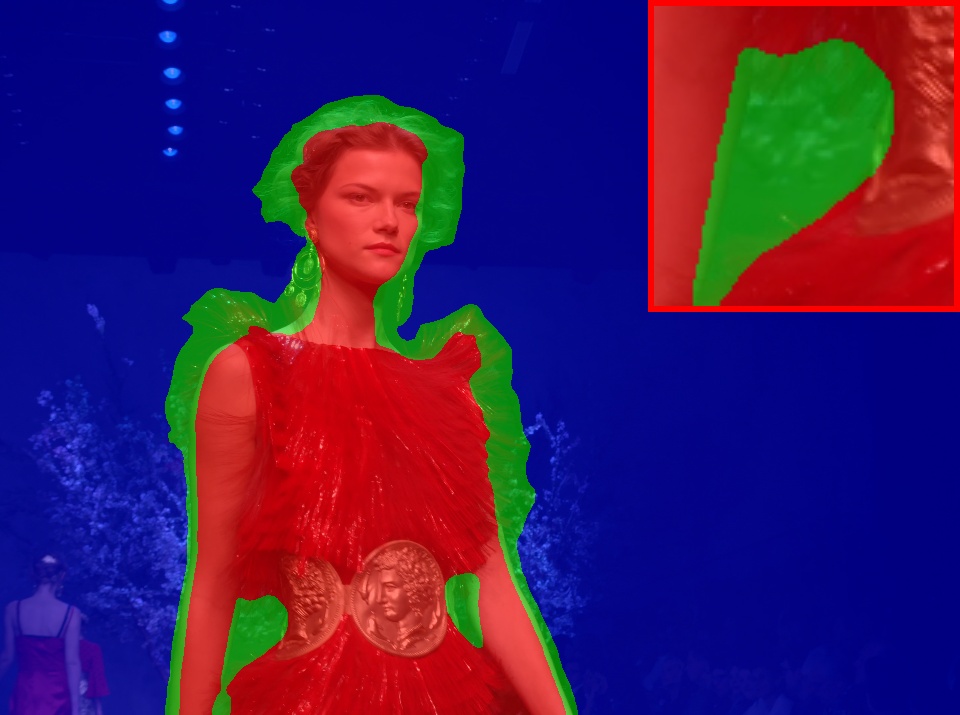} &
        \includegraphics[width=0.15\textwidth]{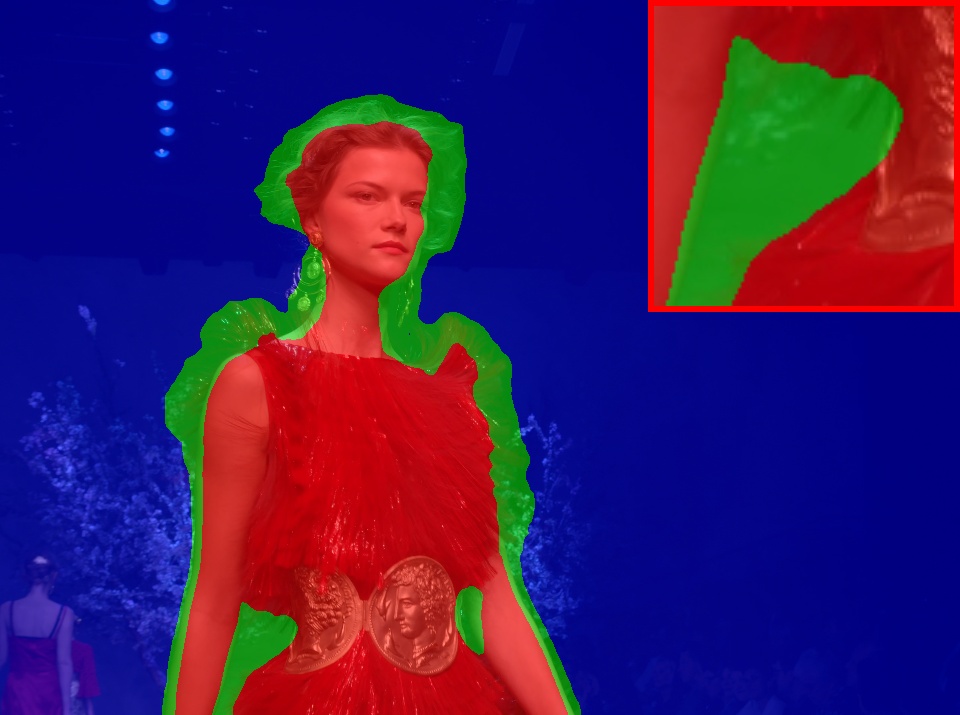} &
        \includegraphics[width=0.121\textwidth]{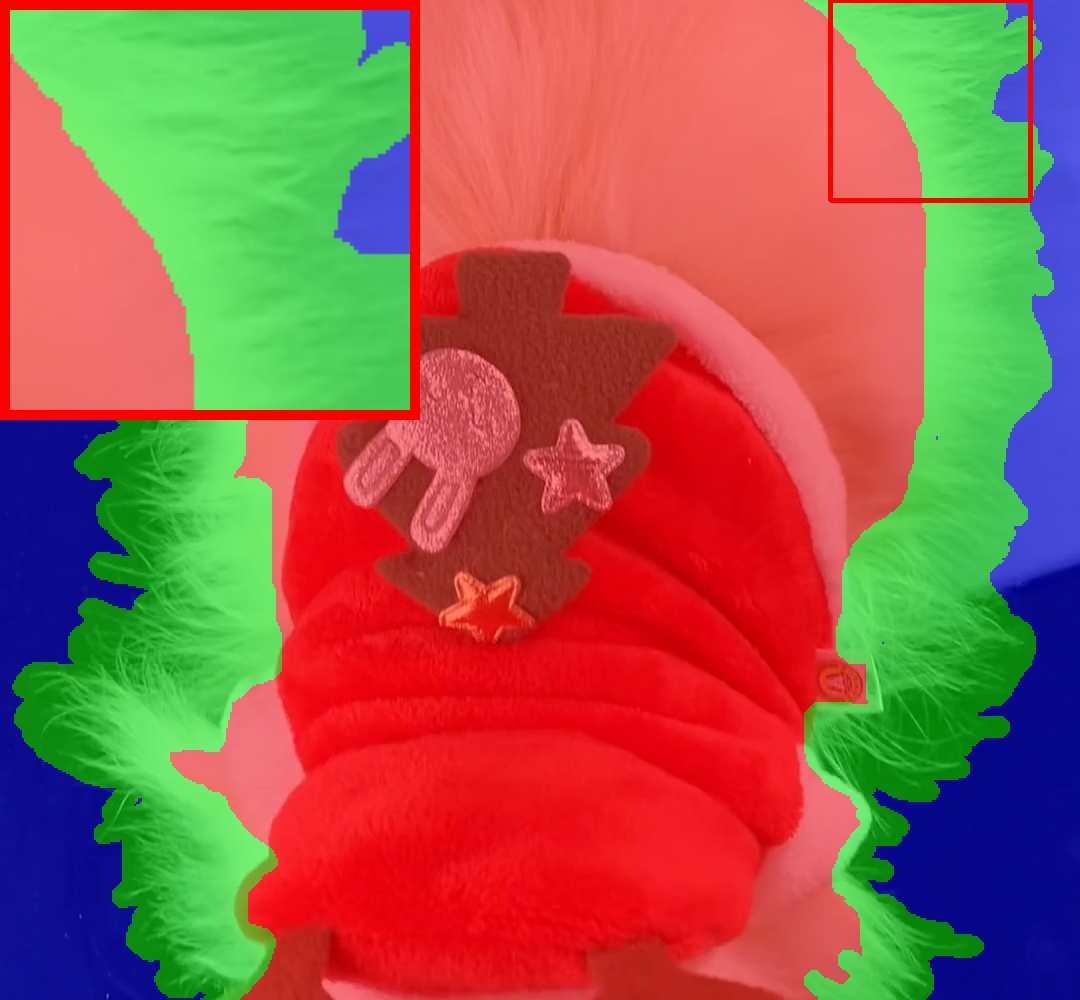} &
        \includegraphics[width=0.121\textwidth]{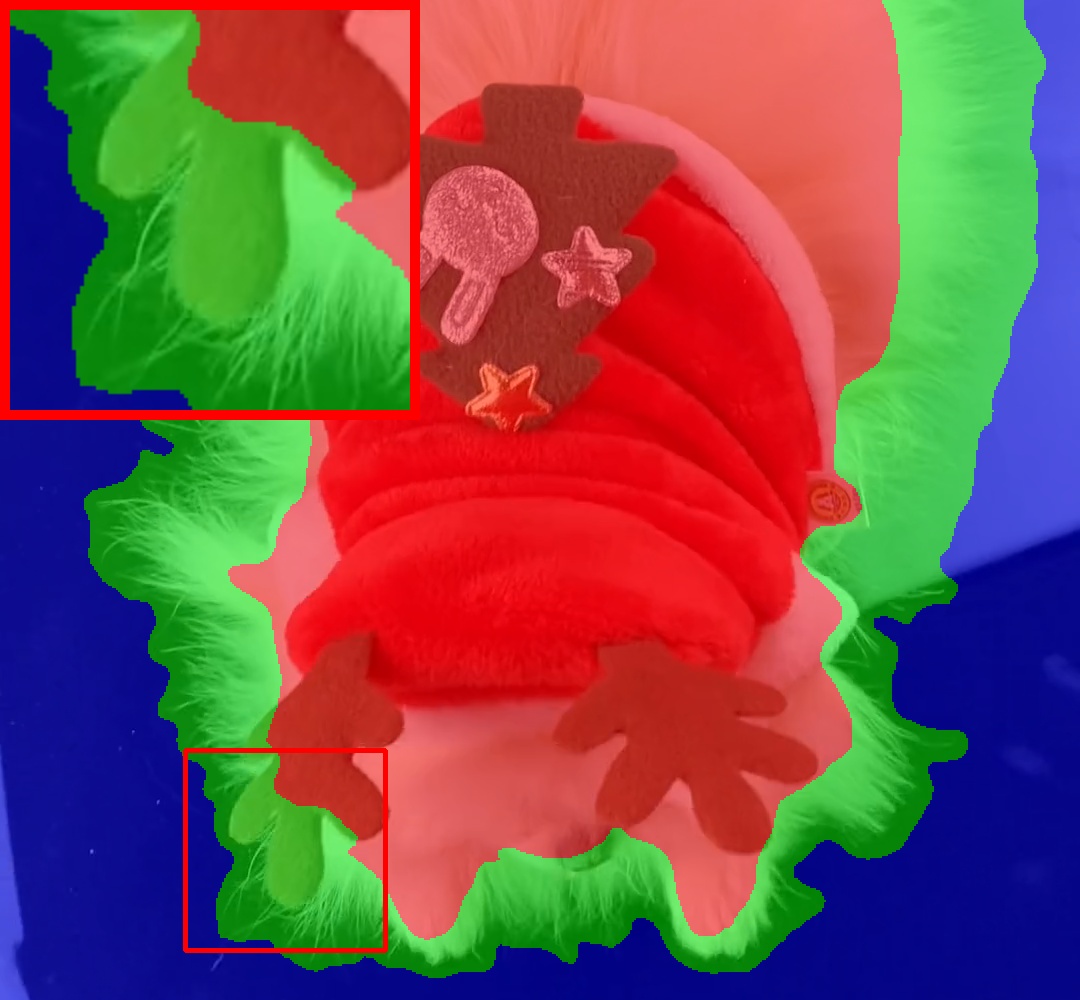} &
        \includegraphics[width=0.15\textwidth]{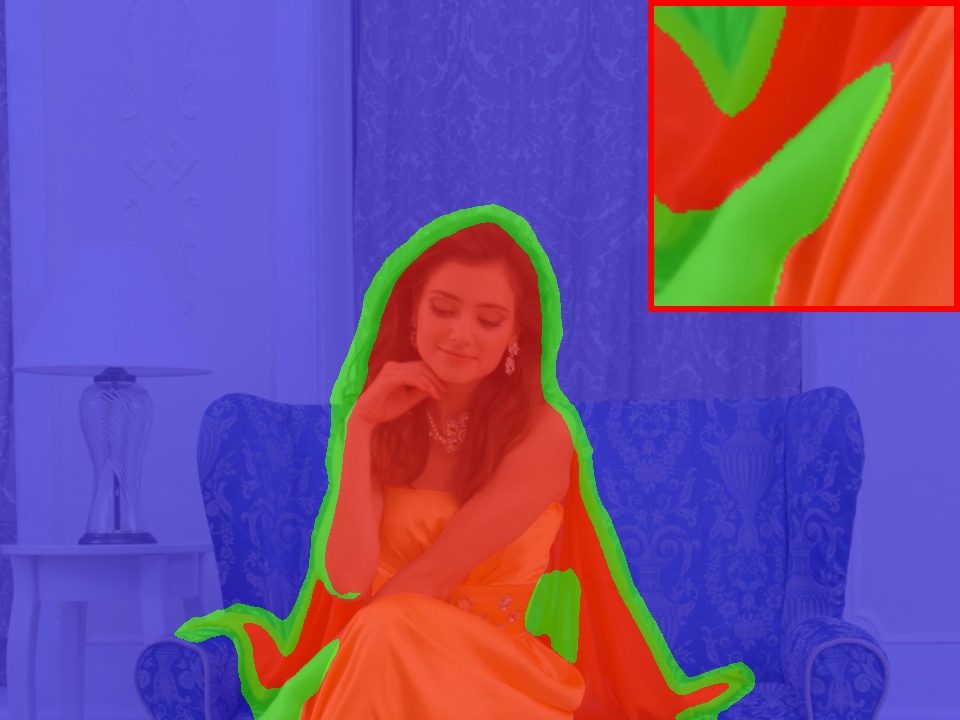} &
        \includegraphics[width=0.15\textwidth]{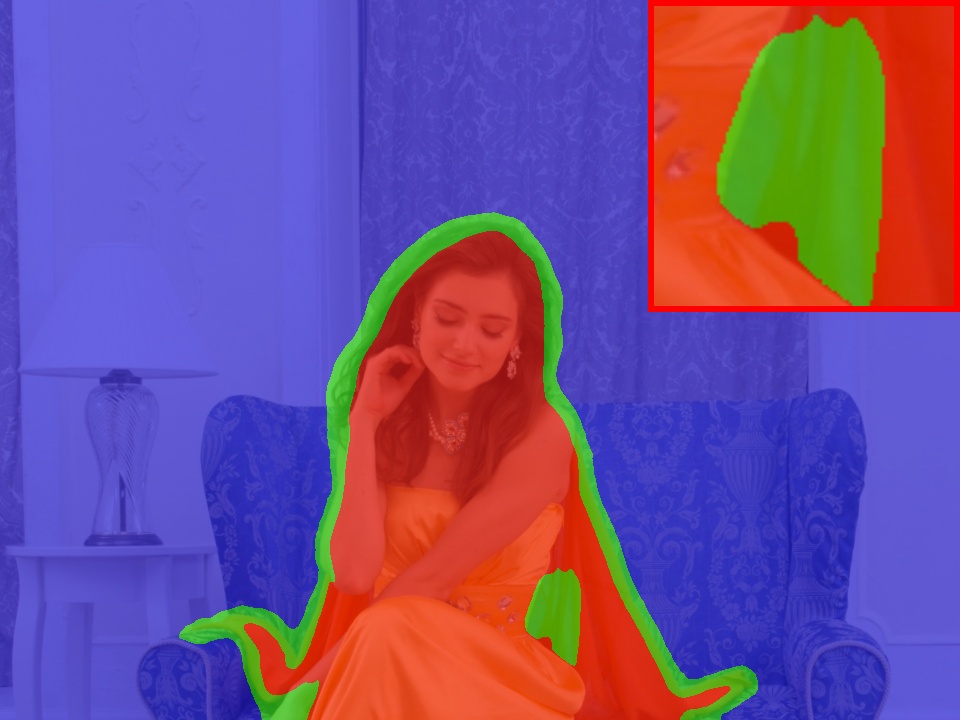} \\
        \rotatebox{90}{\scriptsize{GCA~\cite{li2020natural}}} & 
        \includegraphics[width=0.15\textwidth]{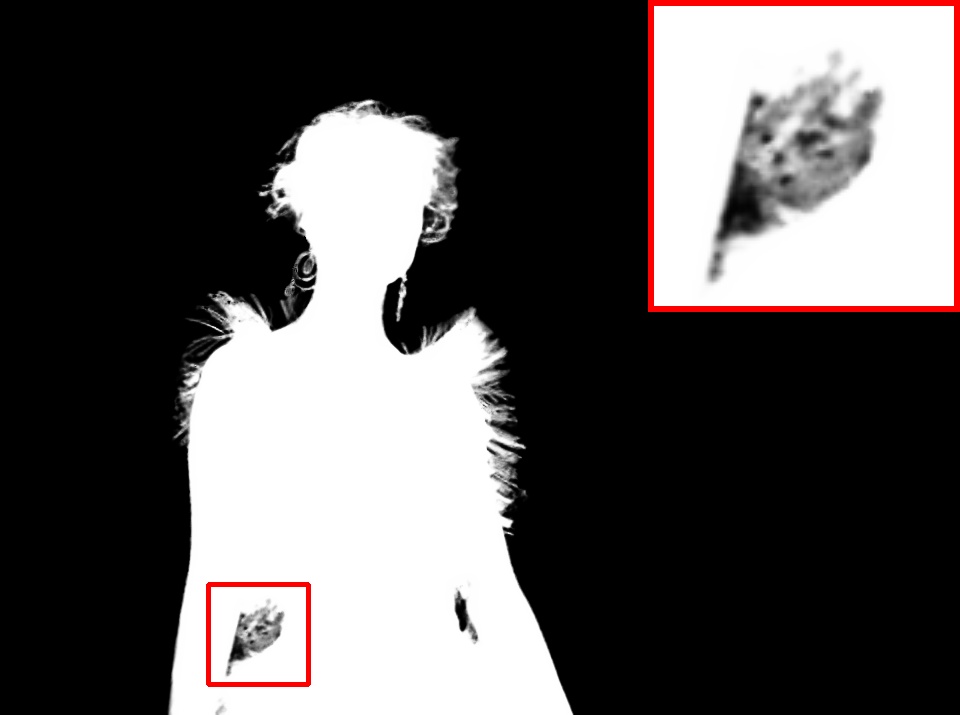} &
        \includegraphics[width=0.15\textwidth]{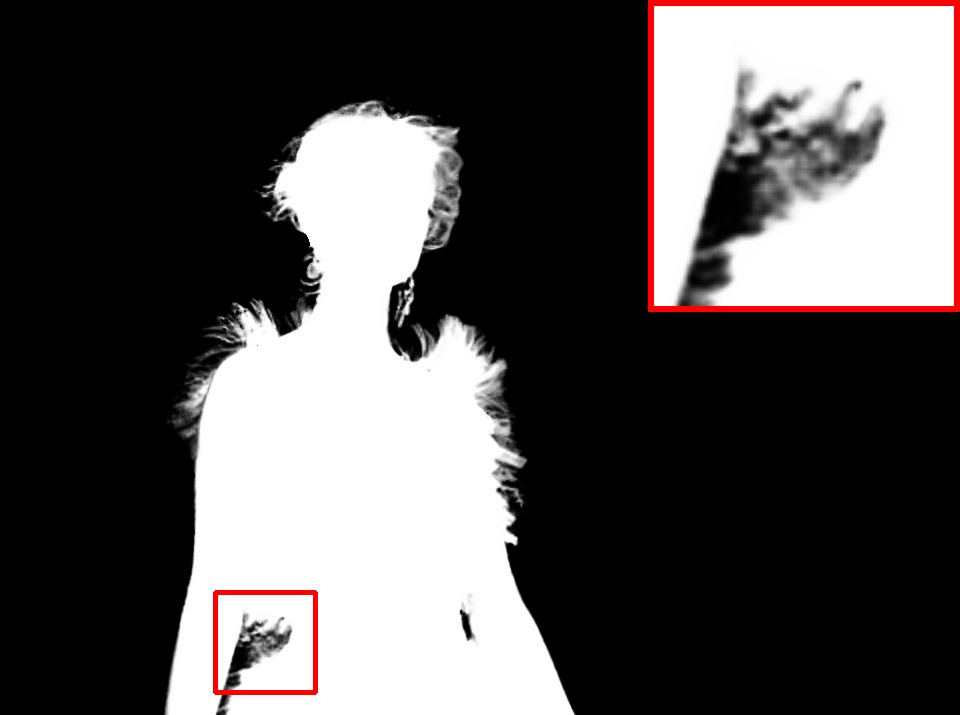} &
        \includegraphics[width=0.121\textwidth]{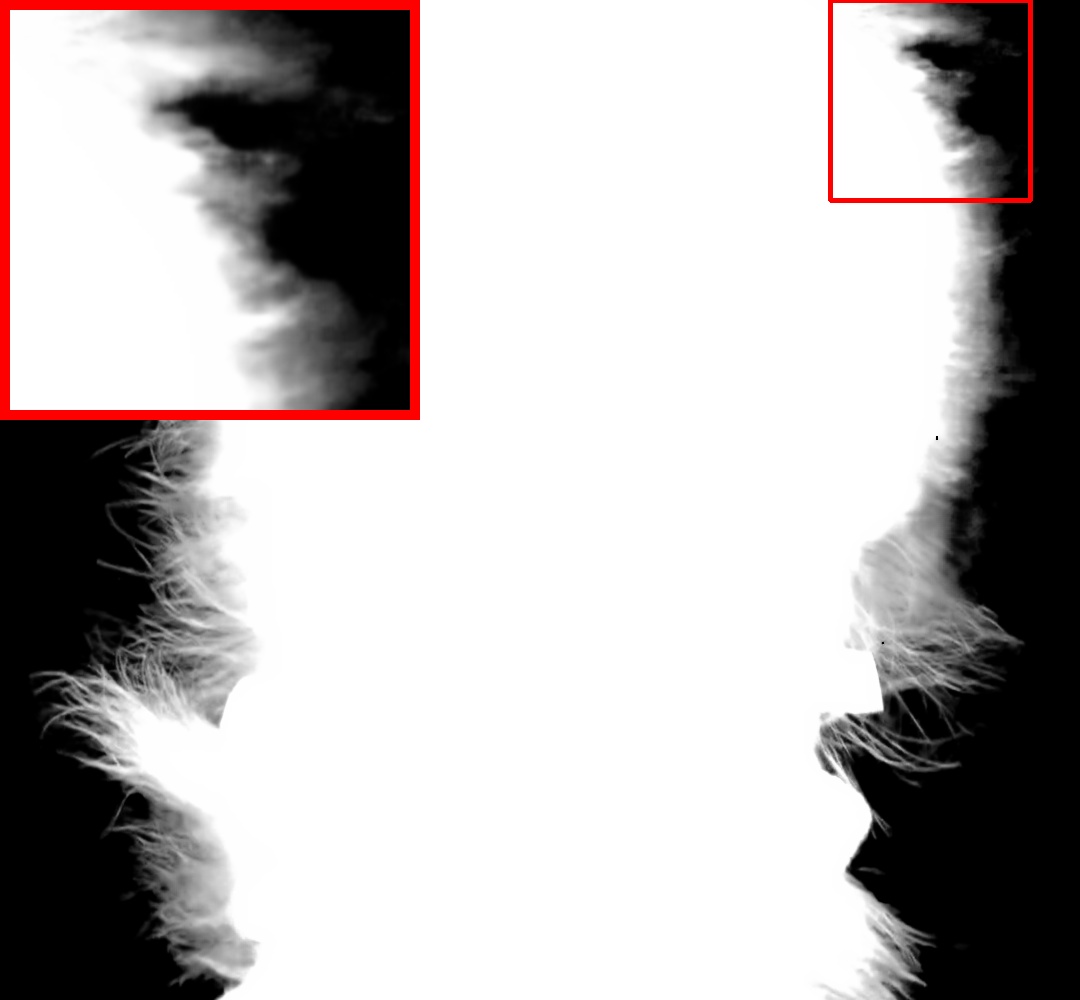} &
        \includegraphics[width=0.121\textwidth]{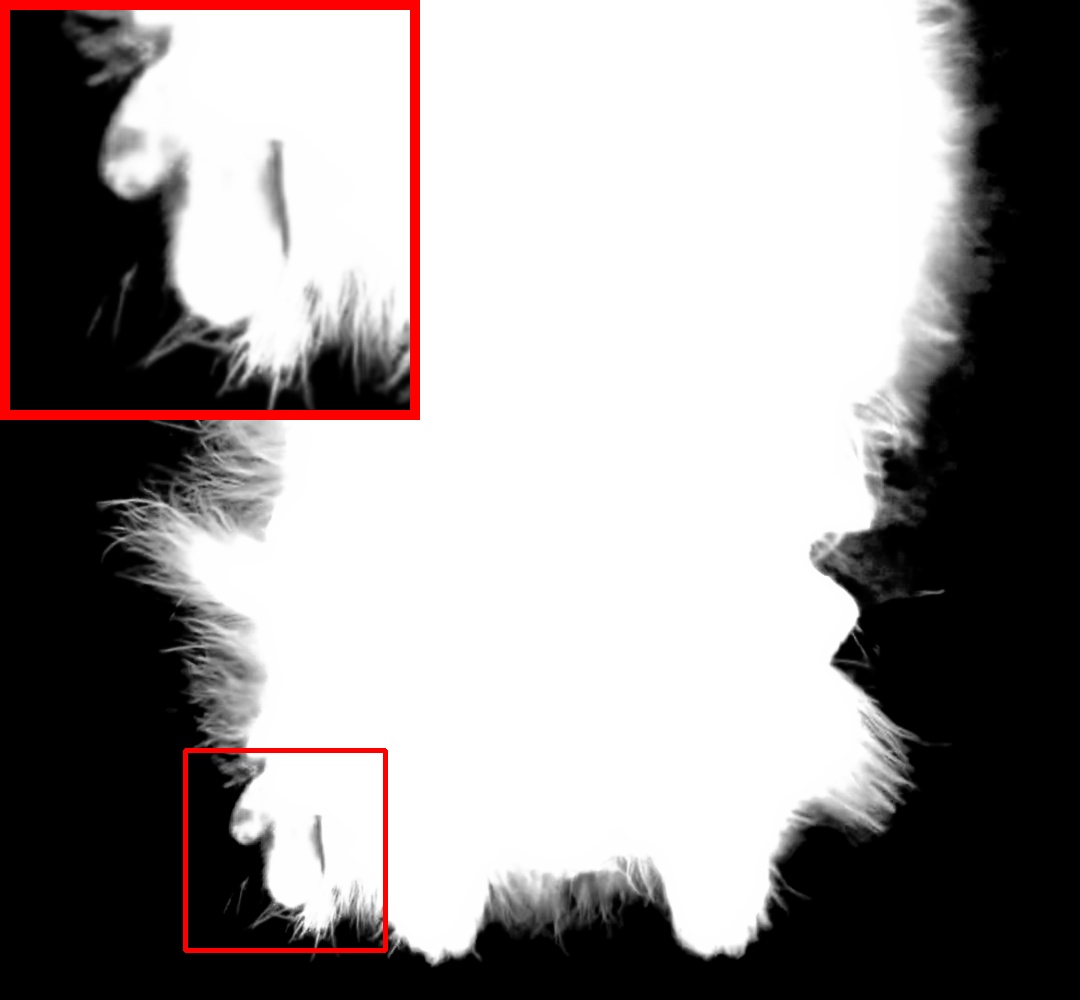} &
        \includegraphics[width=0.15\textwidth]{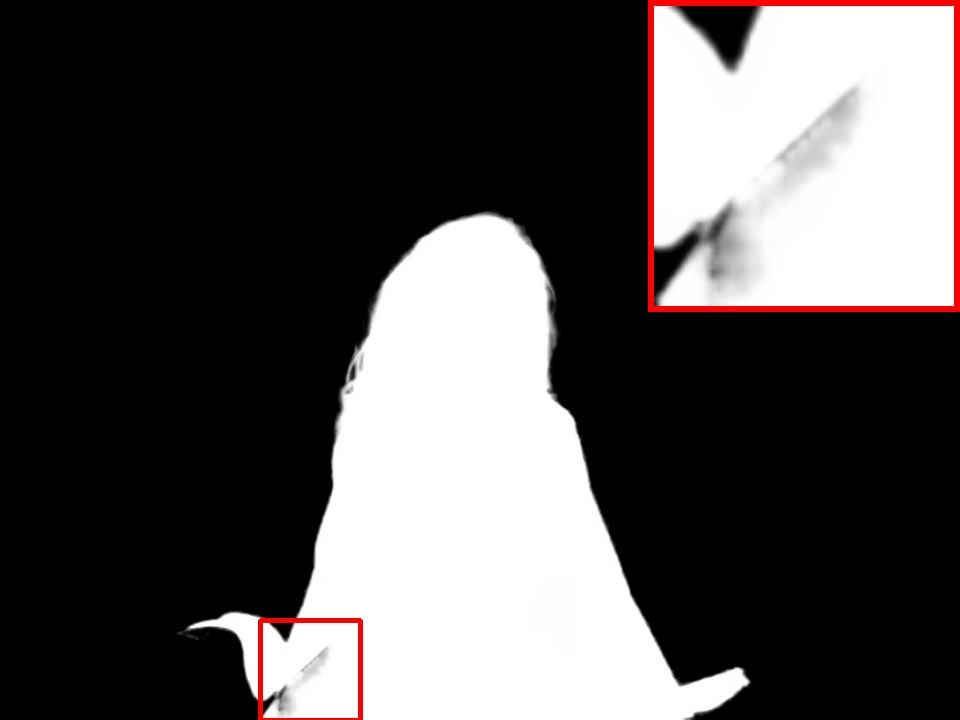} &
        \includegraphics[width=0.15\textwidth]{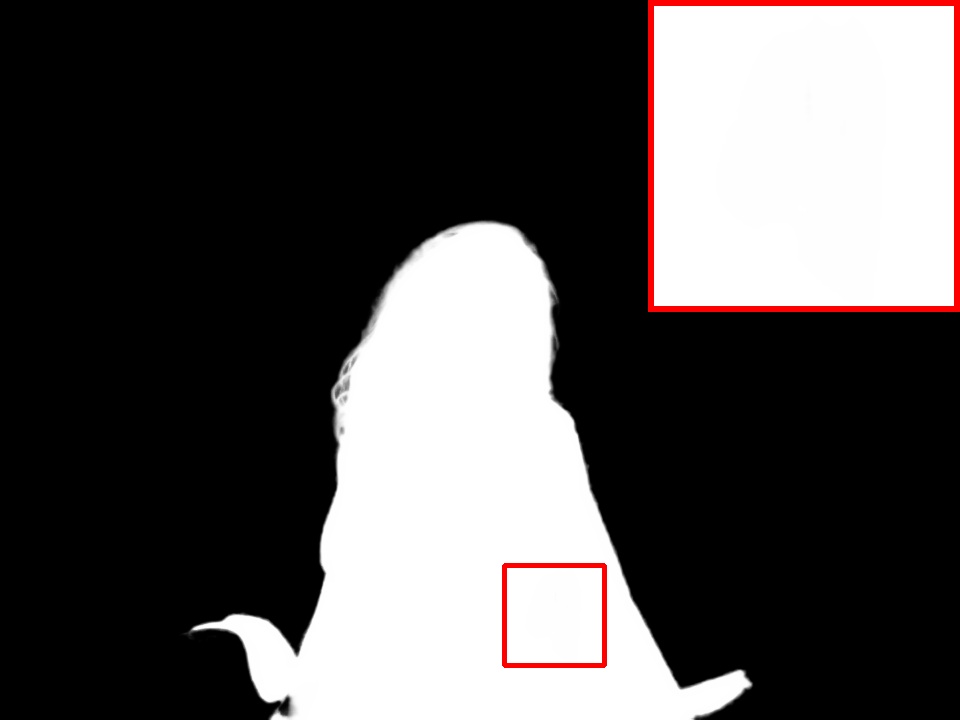} \\
        \rotatebox{90}{\scriptsize{GCA+TAM}} & 
        \includegraphics[width=0.15\textwidth]{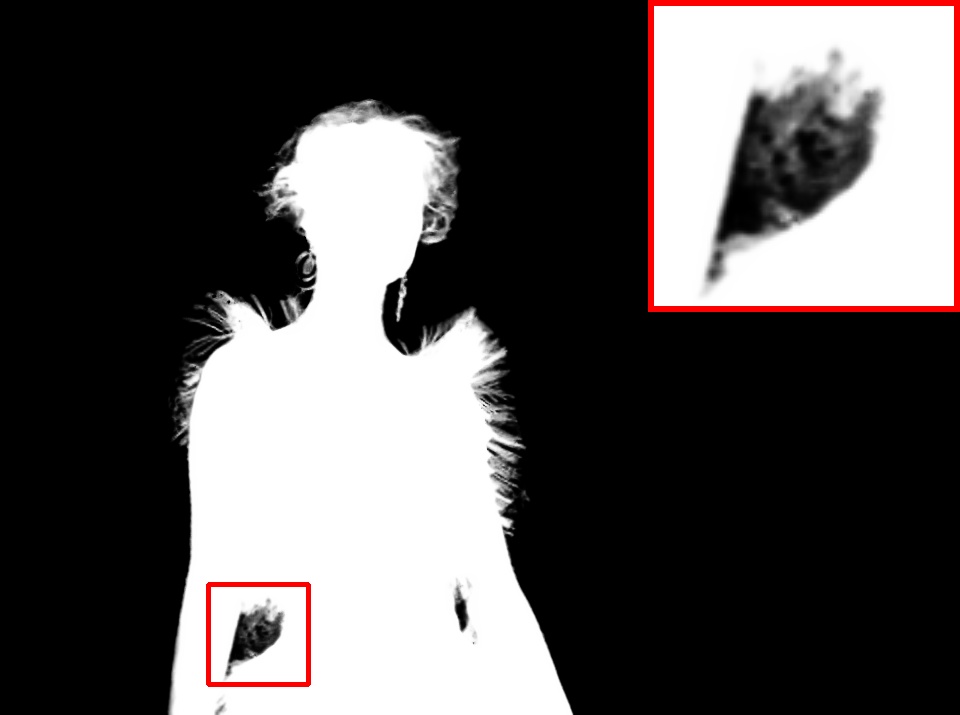} &
        \includegraphics[width=0.15\textwidth]{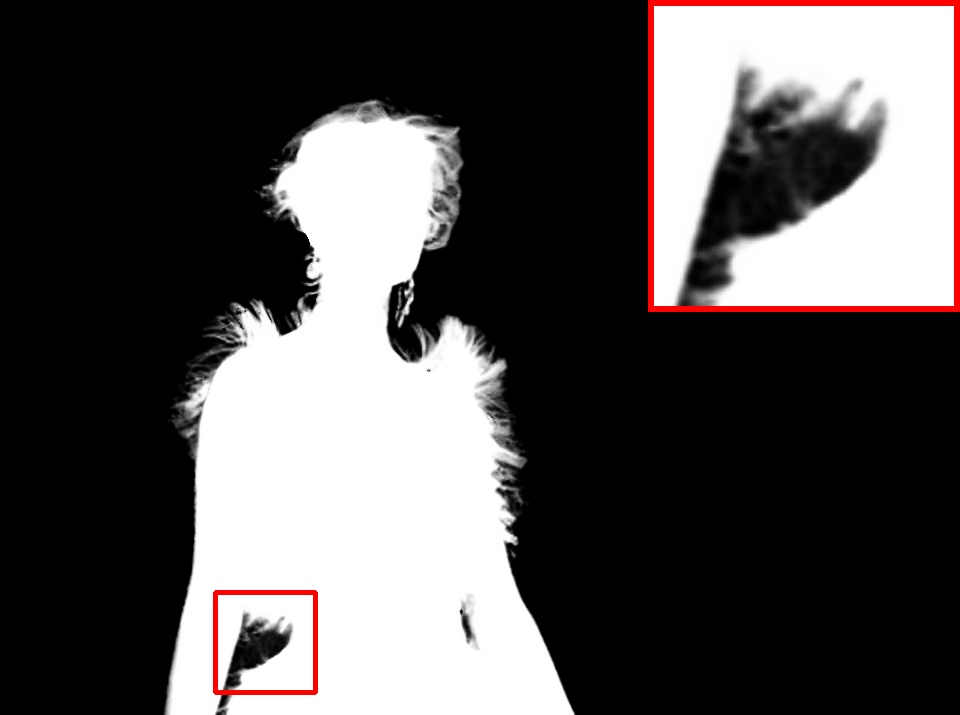} &
        \includegraphics[width=0.121\textwidth]{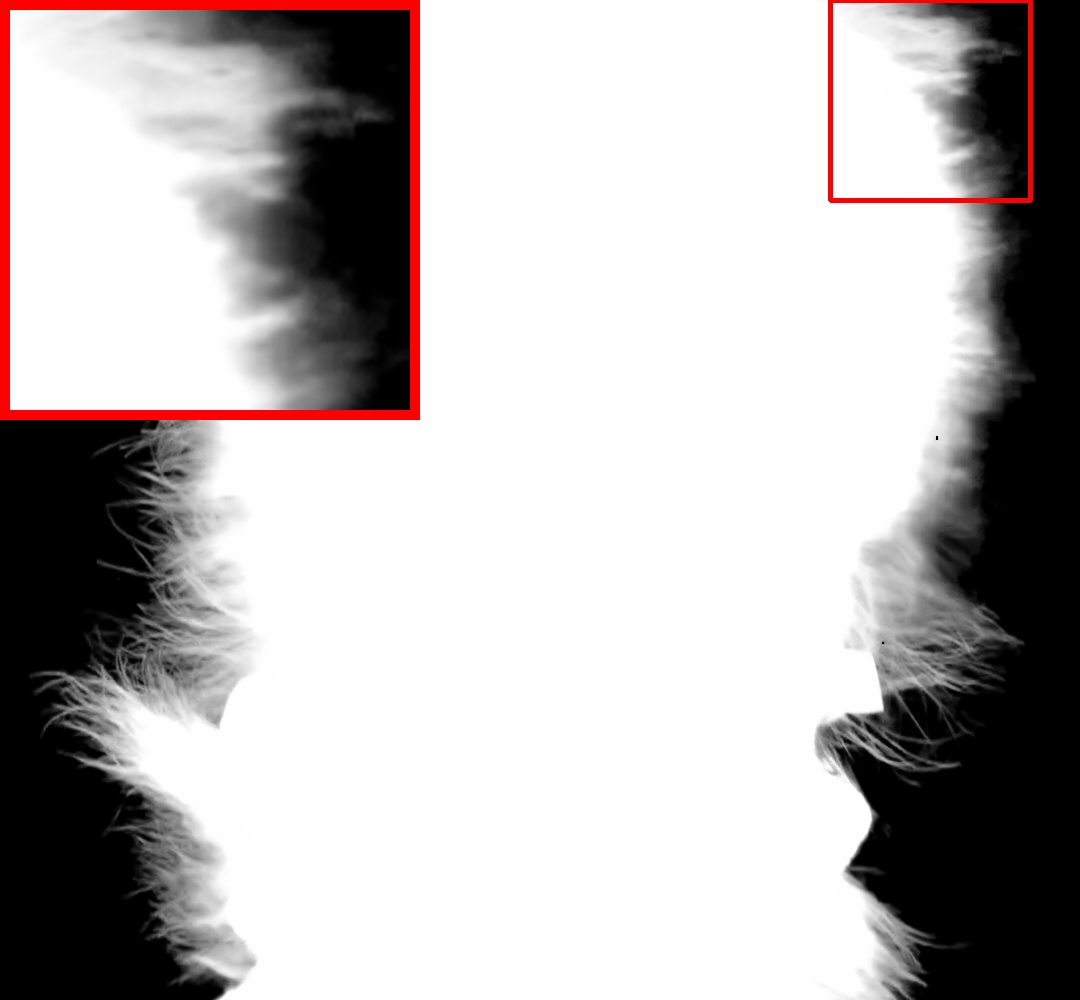} &
        \includegraphics[width=0.121\textwidth]{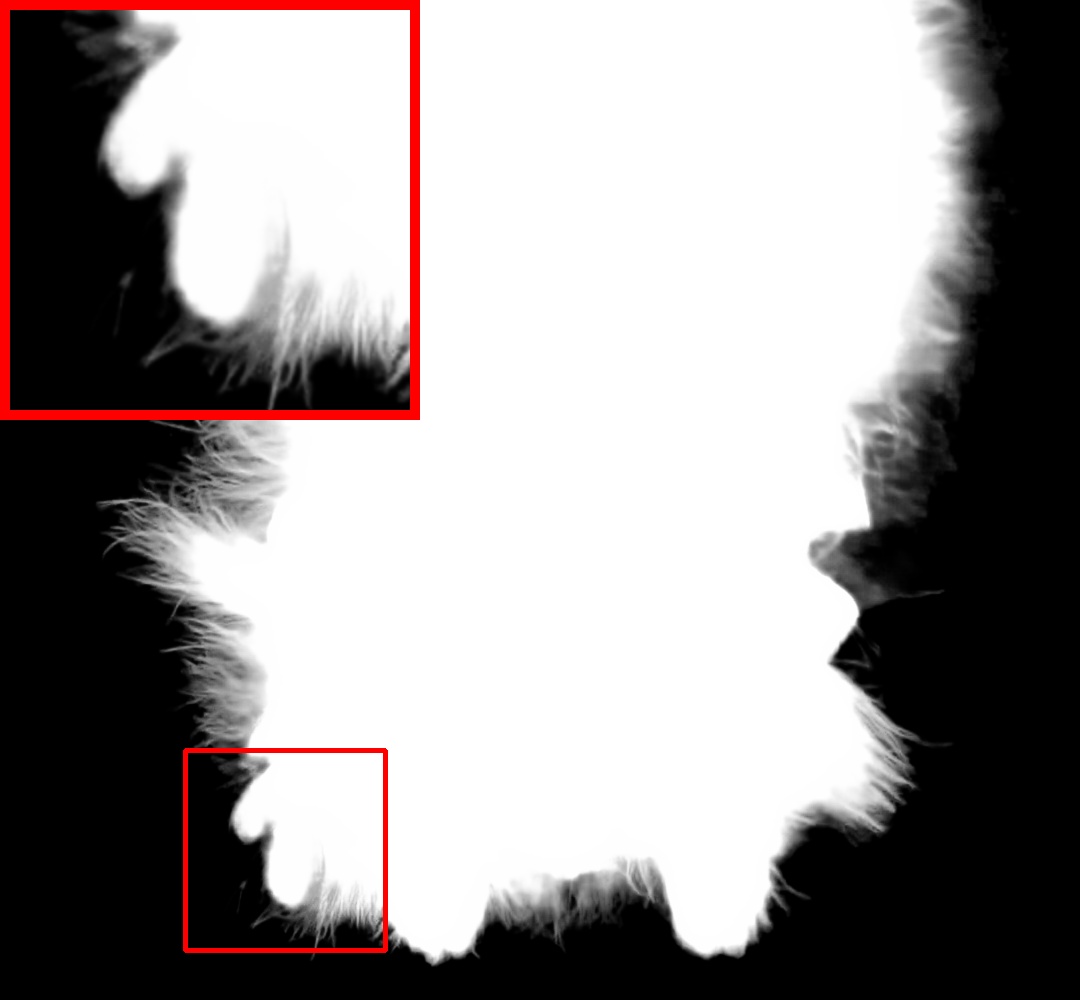} &
        \includegraphics[width=0.15\textwidth]{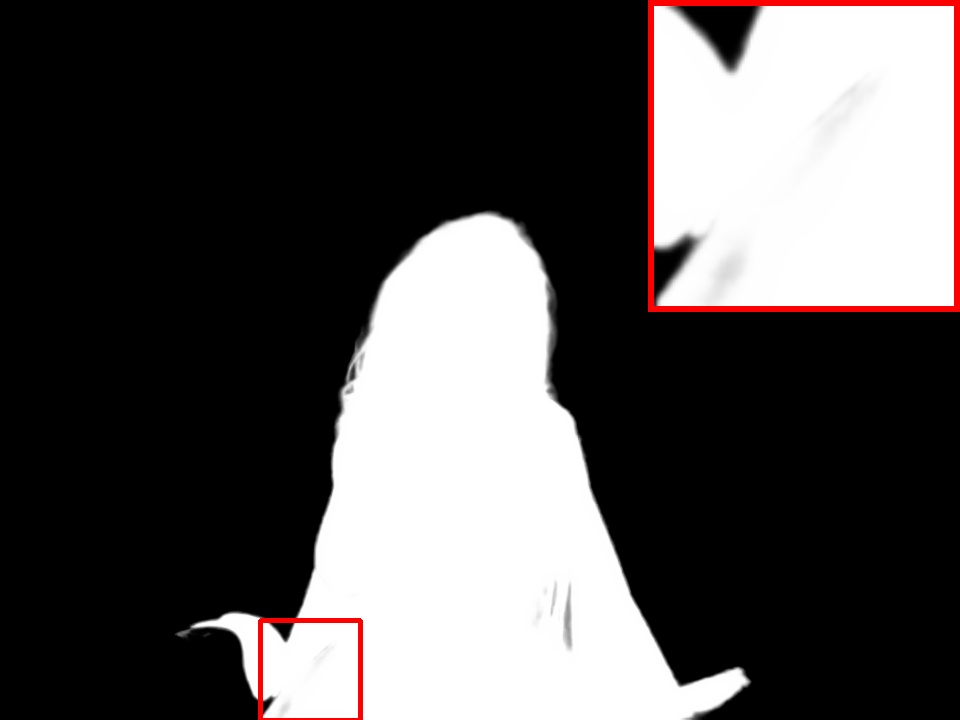} &
        \includegraphics[width=0.15\textwidth]{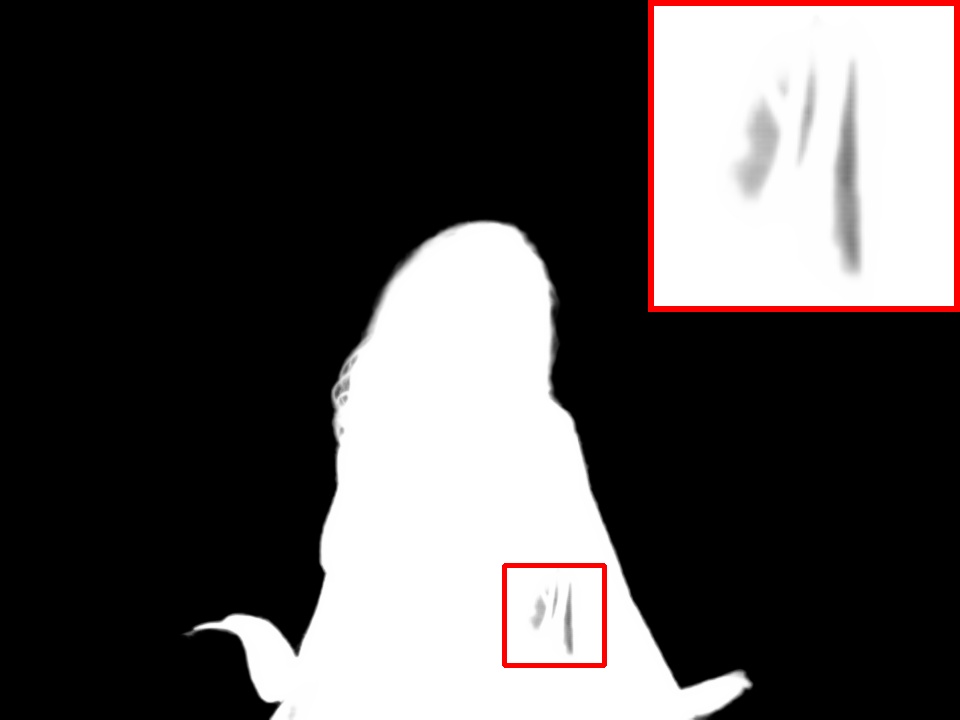} \\
        & \#77 & \#79 & \#9 & \#22 & \#23 & \#92 \\
        & \multicolumn{2}{c}{(a) fashion (k=4)} & \multicolumn{2}{c}{(b) xmas-dog (k=5)} & \multicolumn{2}{c}{(c) sofa (k=3)}
        \end{tabular}
    }
\end{center}
\vspace{-10pt}
\caption{Comparing our GCA+TAM with GCA~\cite{li2020natural} on Internet videos. ``k'' indicates the number of the annotated keyframe trimaps. Please see the supplementary video for the complete results.} 
\label{fig:test_matting}
\vspace{-10pt}
\end{figure*}

\noindent \textbf{Quantitative comparisons}. Table~\ref{tab:vm108_val} shows the quantitative comparisons between our video matting networks and single image matting networks on the Videomatting108 validation set. For fair comparisons, we fine-tune single image matting networks on VideoMatting108 using each video frame as the image matting training data. The learning rate and input resolution are kept the same as we train our video matting network. The results are averaged over all 28 test video clips. It can be seen that our method (denoted ``+TAM'' with ``$L_{im}+L_{tc}+L_{af}$'' in the table) consistently outperform the baseline image matting networks (denoted ``GCA+F'', ``Index+F'' and ``DIM+F'') on all metrics. More comparison results with other methods, such as KNN Video Matting~\cite{Li_2013_ICCV} and DVM~\cite{Sun_2021_CVPR}, can be found in our supplementary material.

Furthermore, in Table~\ref{tab:videomatting.com}, the GCA+TAM network also outperforms GCA on the test dataset from videomatting.com~\cite{Erofeev2015}. This verifies the ability of the proposed TAM that is designed to aggregate temporal information for better video object matting results. Since GCA+TAM achieves much lower metric numbers comparing to DIM+TAM and Index+TAM, we use GCA+TAM as the default for evaluating video matting, if not mentioned otherwise.

\begin{table}[t]
    \caption{Ablation study on the temporal window size $W$ and the number of aggregated frames used for GCA+TAM. ``nF'' denotes $n$-frames are aggregated in TAM. 2F: only uses $\mathbf{I}^{t-1}$; 3F: uses $\mathbf{I}^{t-1}$ and $\mathbf{I}^{t+1}$; 5F: uses $\mathbf{I}^{t-2}$, $\mathbf{I}^{t-1}$, $\mathbf{I}^{t+1}$ and $\mathbf{I}^{t+2}$.}
    \label{tab:abl_window}
    \vspace{-10pt}
    \small
    \centering
    \begin{tabularx}{\linewidth}{l|l|YYY}
        \hline
        $W$ & F & SSDA & dtSSD & MESSDdt \\
        \hline
        $W=5$ & 3F & 49.75 & 24.08 & 1.30 \\
        $W=7$ & 3F & 47.59 & 23.53 & 1.19 \\
        $W=9$ & 3F & 52.35 & 24.29 & 1.38 \\
        $W=7$ & 2F & 48.51 & 23.81 & 1.30 \\
        $W=7$ & 5F & 51.25 & 24.79 & 1.32 \\
        \hline
    \end{tabularx}
\vspace{-15pt}
\end{table}

\begin{table}[t]
    \caption{Ablation study for trimap generation on VideoMatting108 validation set using the mIoU metric. All metrics are averaged with a per-video basis. ``nFT'' denotes how many keyframe trimaps are used in fine-tuning. 1FT: first frame as keyframe. 2FT: first+last frame as keyframes. 3FT: first+last+100-th frame as keyframes.}
    \label{tab:abl_trimap}
    \vspace{-10pt}
    \small
    \centering
    \begin{tabularx}{\linewidth}{l|YYYY}
        \hline
        Method & FR & BR & UR & Average  \\
        \hline
        STM~\cite{oh2019video}     & 81.43 & 95.58 & 81.63 & 86.21 \\
        STM+1FT & 85.92 & 96.62 & 82.75 & 88.43 \\
        STM+2FT & 87.73 & 97.91 & 84.90 & 90.18 \\
        STM+3FT  & 87.72 & 97.93 & 85.04 & 90.23 \\
        \hline
    \end{tabularx}
\vspace{-15pt}
\end{table}

\noindent \textbf{Ablation studies}. We first investigate the influence of the weight sharing in TAM (see the second to the fourth row in Table~\ref{tab:vm108_val}). Different from the widely utilized ``KQV'' structure without any weight sharing, in our case we empirically found out that sharing ``query'' and ``value'' weights achieves the best result compared with other weight sharing configurations. Note that the conventional ``KQV'' structure without weight sharing performs worse than the baseline method without TAM on the validation set. We speculate that separating all weights causes over-fitting during the training, since it does achieve lower training loss compared with our design.

We proceed to assess the influence of different loss terms (see the fifth to the seventh row in Table~\ref{tab:vm108_val}). By only adding the temporal coherence term $L_{tc}$, we obtain performance boost across all metrics, except pixel-wise alpha value accuracy in ``narrow''  trimaps. Since the UR of a narrow trimap mainly consists of transparent pixels, we speculate that the motion ambiguities at these pixels are the main reason for the drop of alpha value accuracy. By only adding the target affinity term $L_{af}$, the temporal metrics ``dtSSD'' and ``MESSDdt'' are slightly improved while the alpha value accuracy metrics are comparable to the baseline. By combining both terms, the network achieved much improved results. For the ``narrow'' trimaps in particular, the direct supervision in $L_{af}$ could suppress the erroneous affinity values, thus improving performance.

In Table~\ref{tab:abl_window}, we analyze the influence of the temporal window size (Sec.~\ref{sec:tam}) parameter $W$ and the number of frames used in TAM. To reduce the computational cost, we conduct experiments on half of the VideoMatting108 training and validation set with ``medium'' ground-truth trimaps. We can see that the network performance achieves the best balance when $W=7$. In contrast, the network performance degrades when $W=9$. The reason may come from the difficulties of suppressing a large number of unrelated features from adjacent frames through attention. Thus, we choose $W=7$ in all of our experiments. To verify the bi-directional design, we conduct experiments on the number of aggregated frames (fourth and fifth rows). As seen in Table~\ref{tab:abl_window}, our bi-directional design (second row, 3F) outperforms all other configurations.

\begin{figure}[t]
\begin{center}
    \setlength{\tabcolsep}{1pt}
    \resizebox{0.925\linewidth}{!}{
        \begin{tabular}{cccc}
        \rotatebox{90}{\scriptsize{Frames}} &
        \includegraphics[width=0.3\columnwidth]{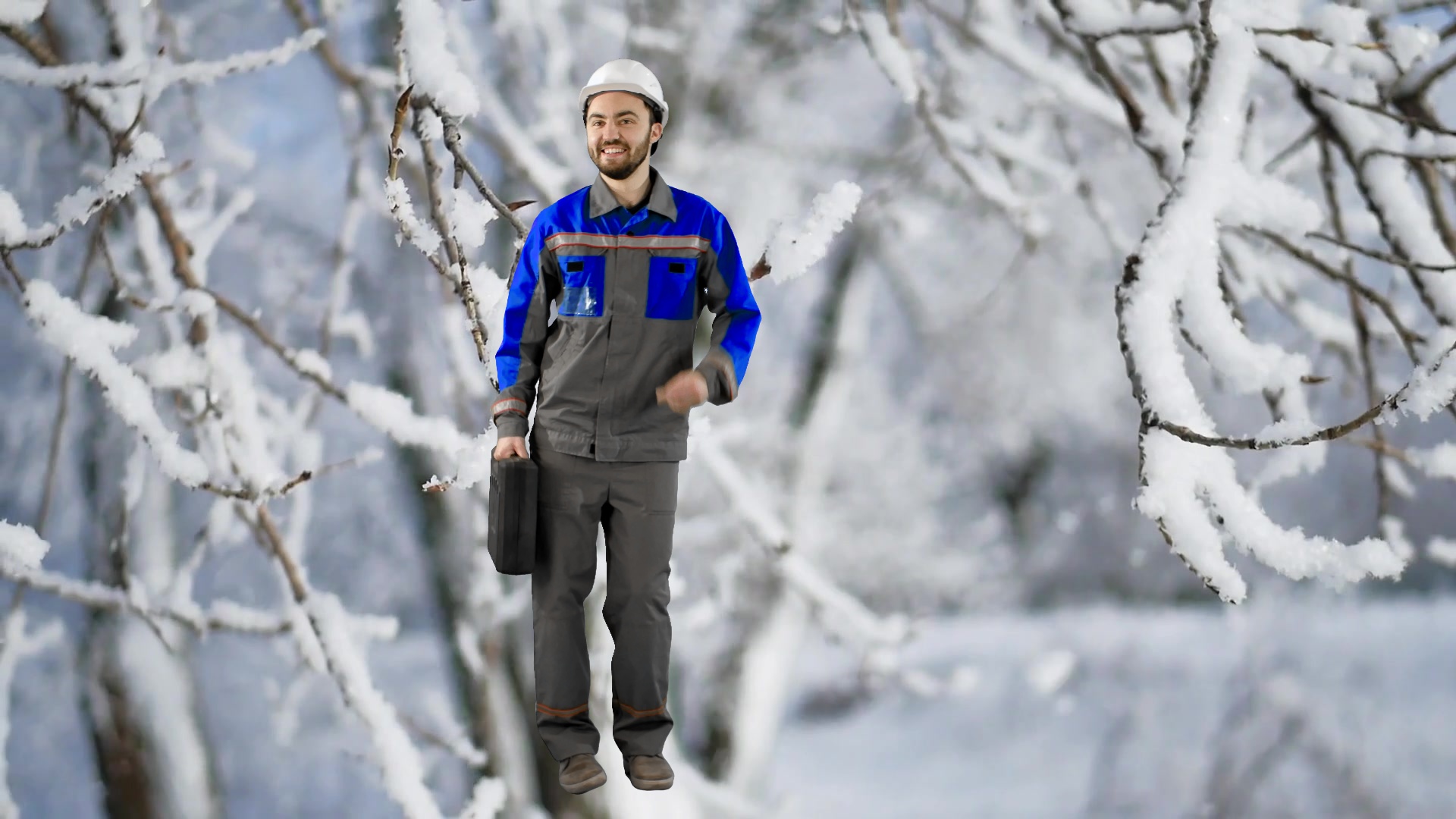} &
        \includegraphics[width=0.3\columnwidth]{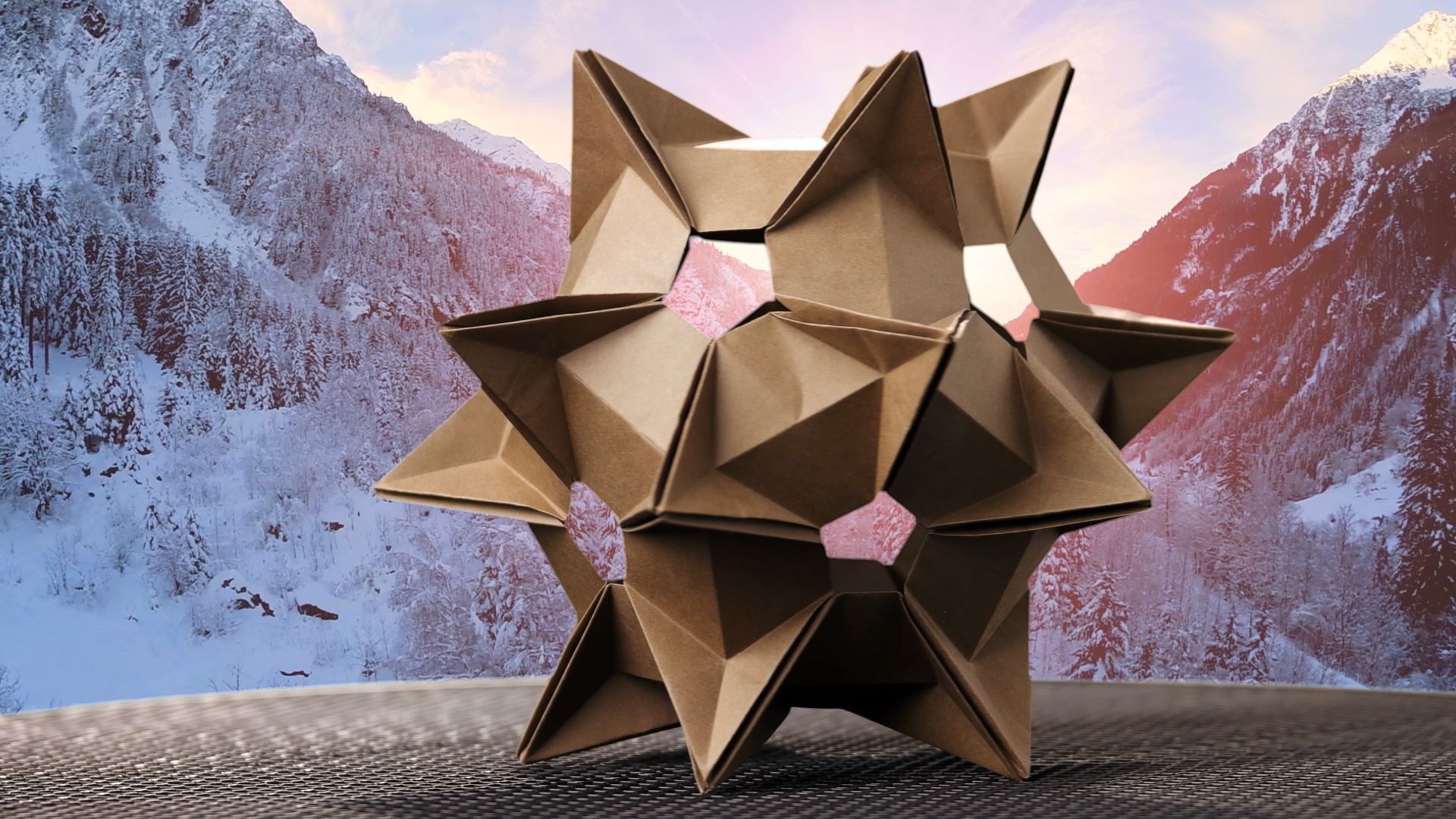} &
		\includegraphics[width=0.3\columnwidth]{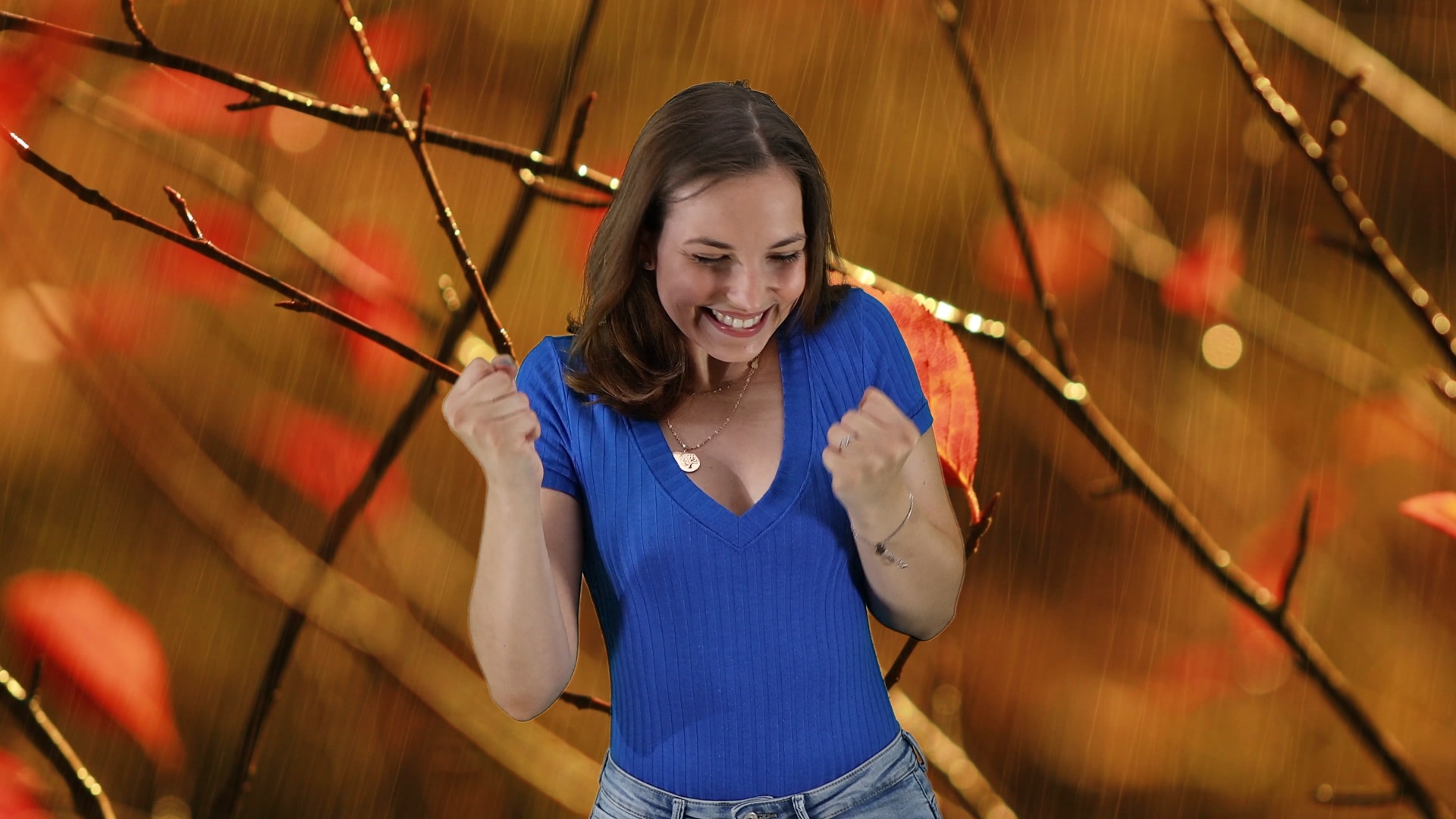} \\
        \rotatebox{90}{\scriptsize{STM~\cite{oh2019video}}} &
        \includegraphics[width=0.3\columnwidth]{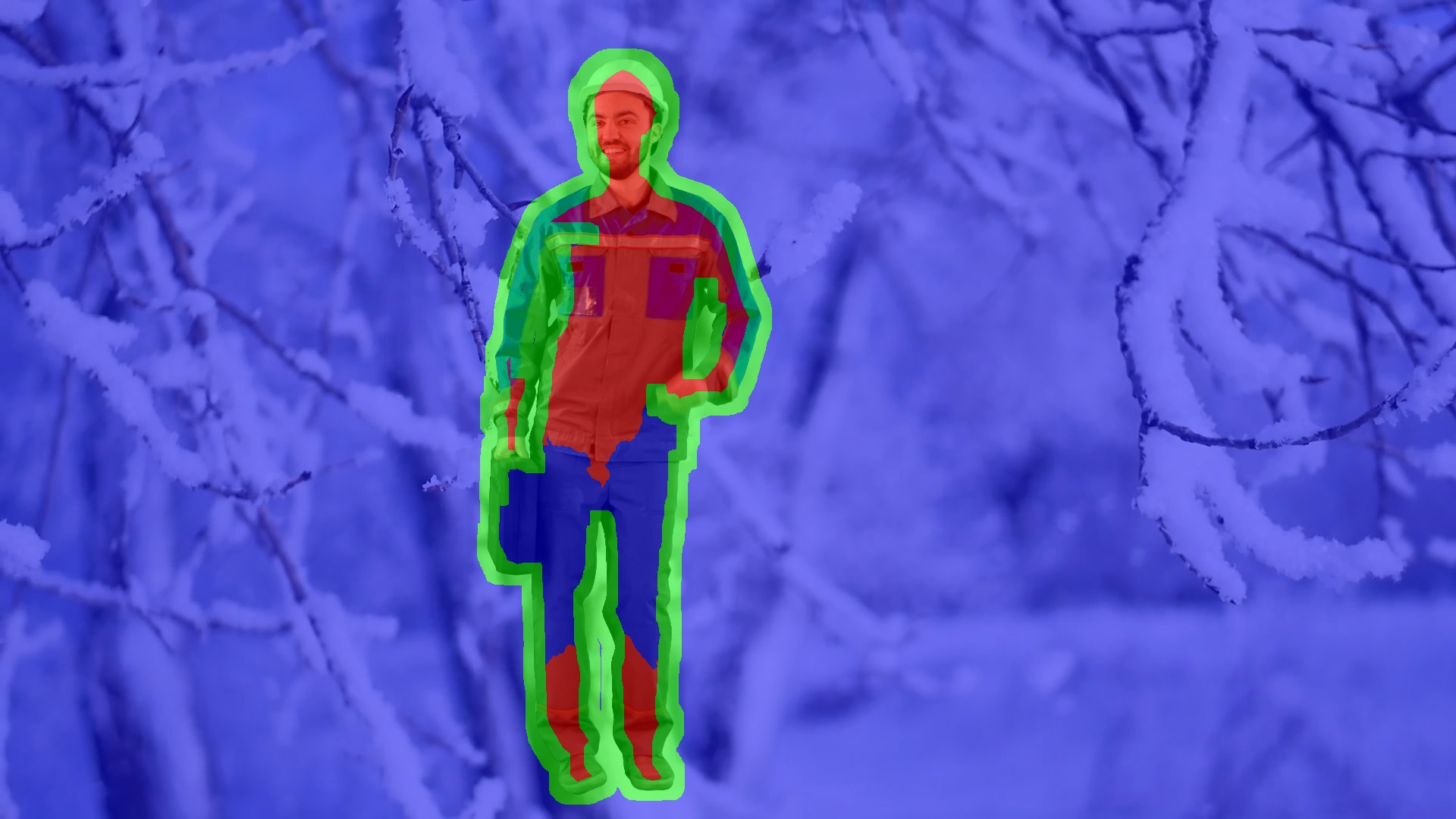} &        
        \includegraphics[width=0.3\columnwidth]{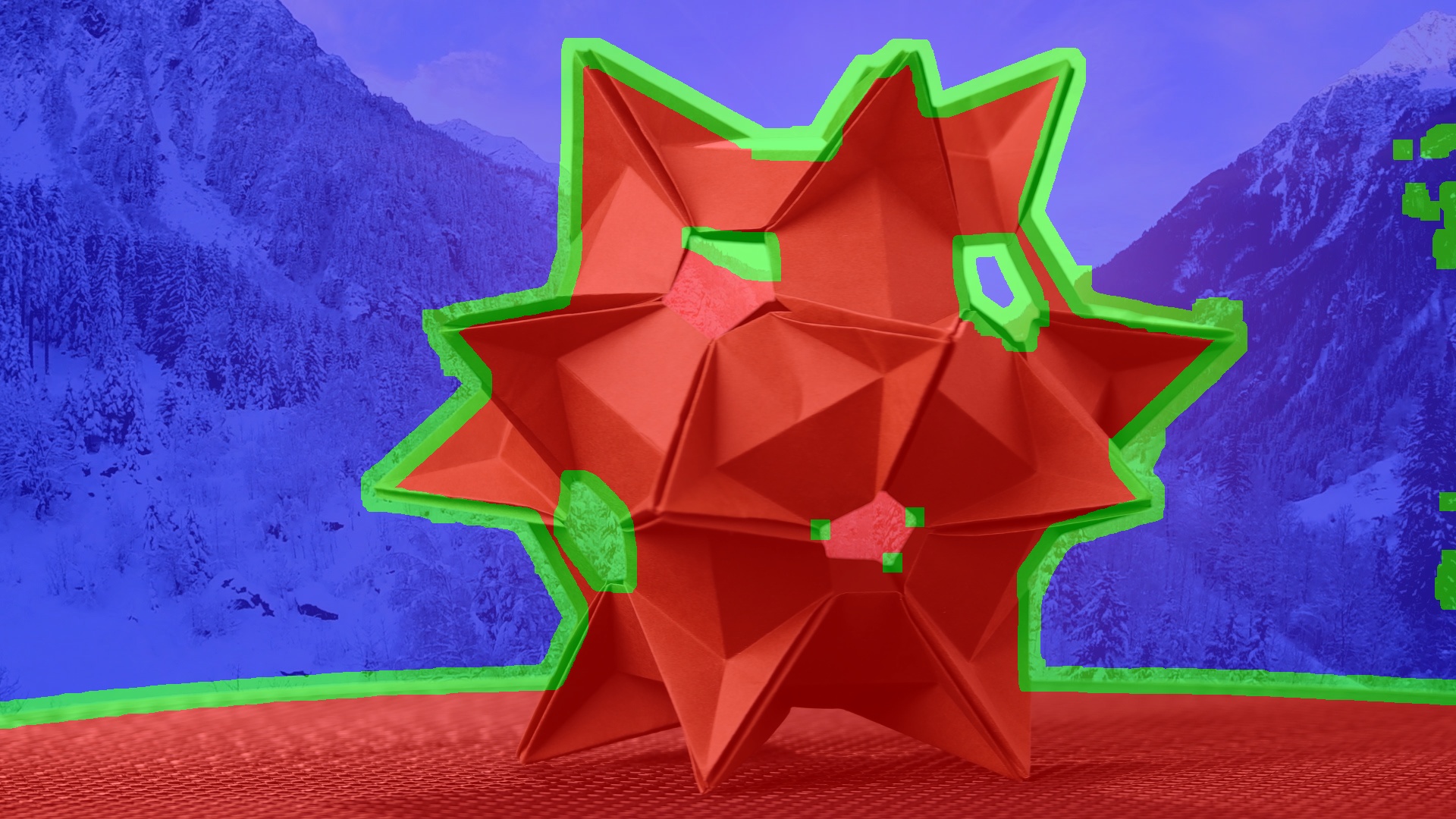} &
		\includegraphics[width=0.3\columnwidth]{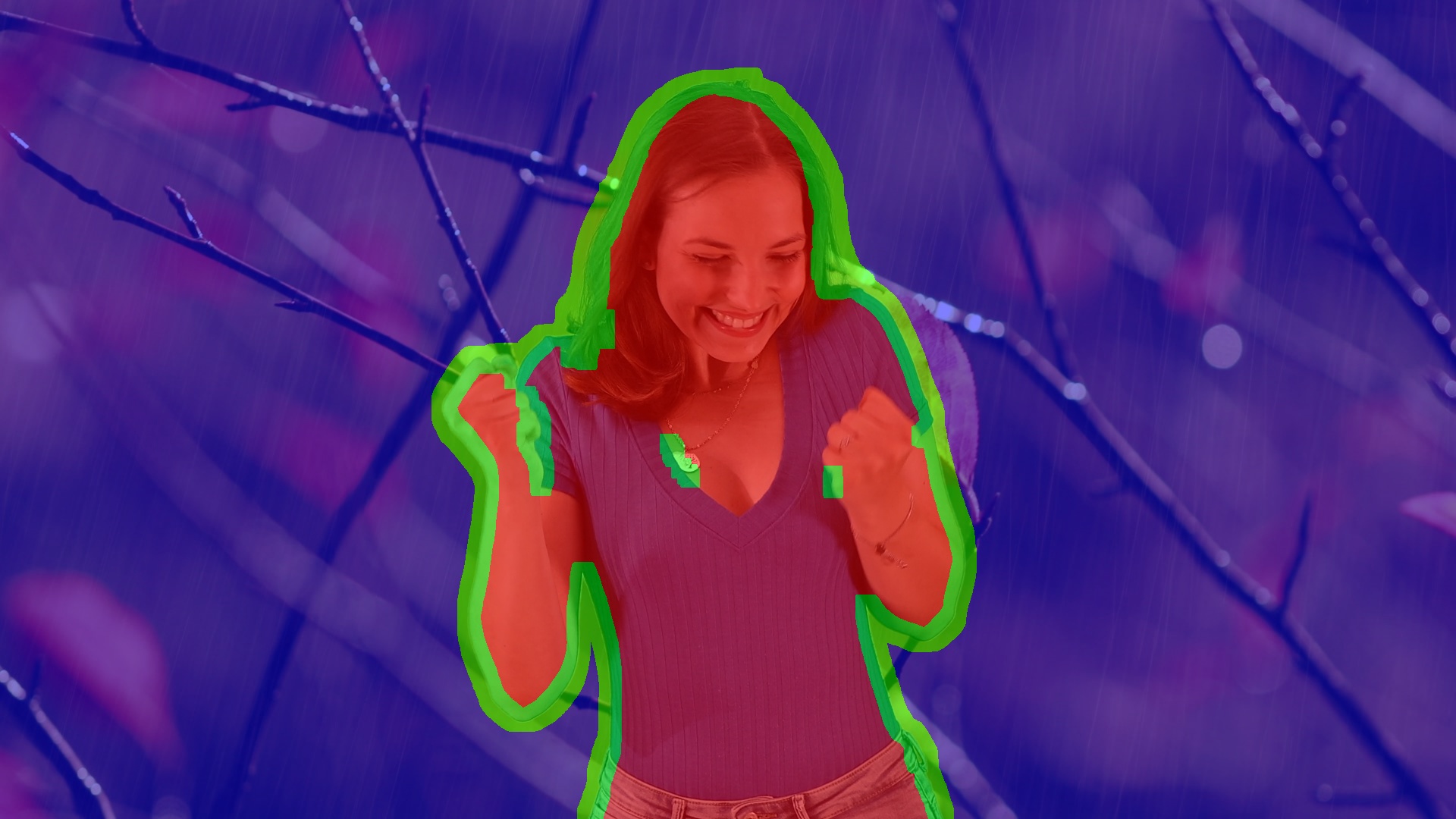} \\
        \rotatebox{90}{\scriptsize{STM+3FT}} & 
        \includegraphics[width=0.3\columnwidth]{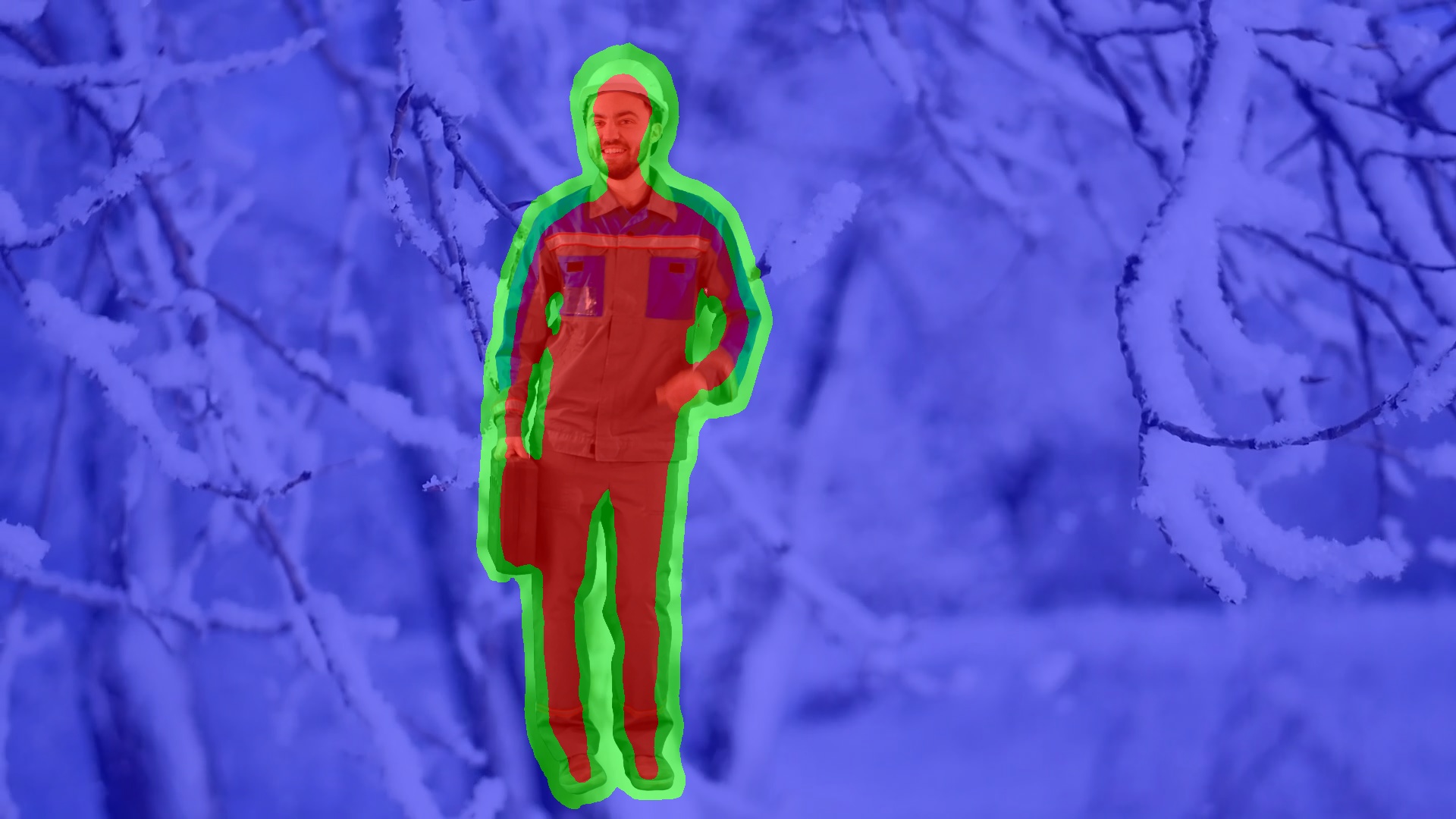} &
        \includegraphics[width=0.3\columnwidth]{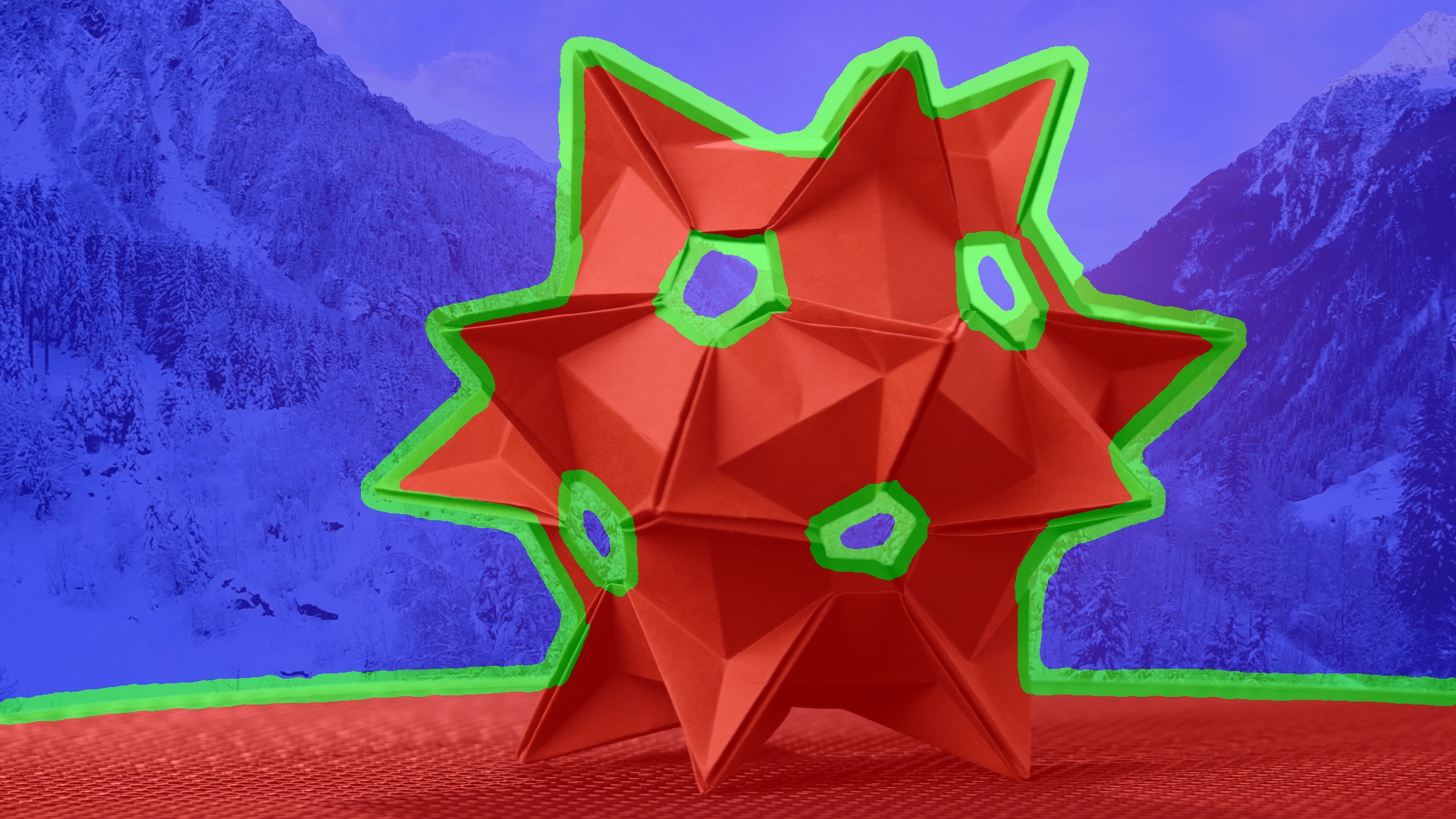} &
		\includegraphics[width=0.3\columnwidth]{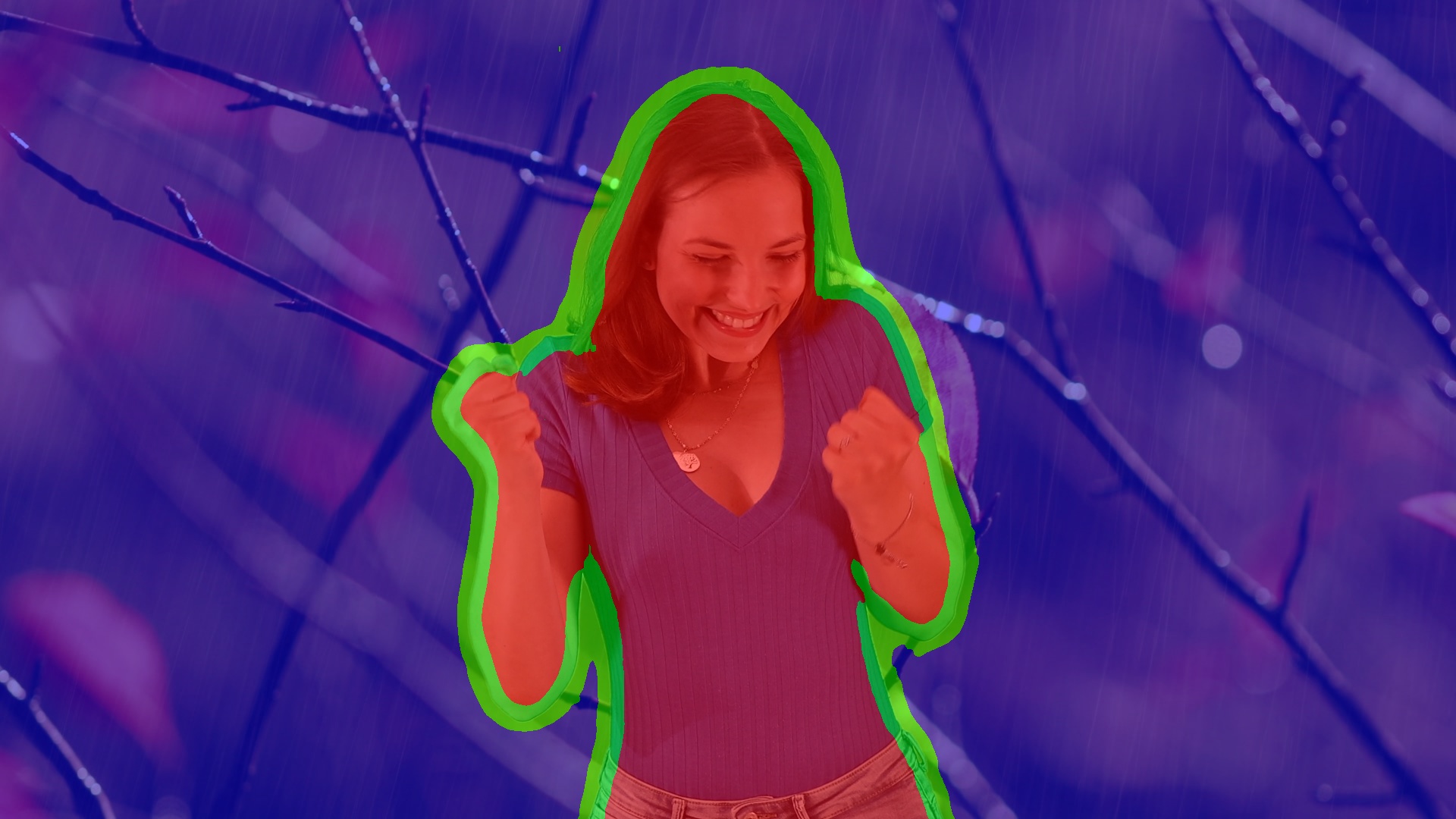} \\
        \rotatebox{90}{\scriptsize{Matte STM}} & 
        \includegraphics[width=0.3\columnwidth]{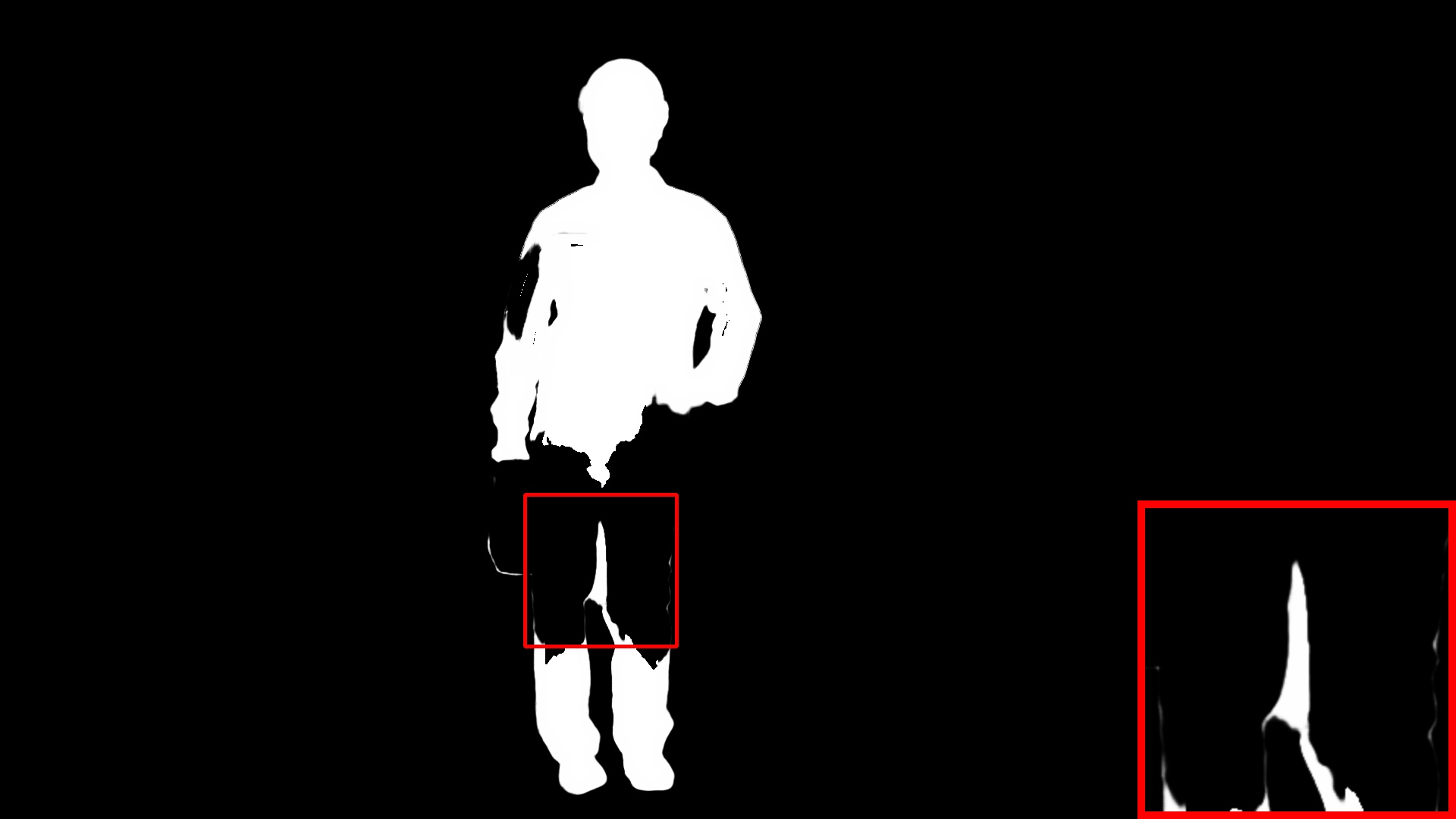} &
        \includegraphics[width=0.3\columnwidth]{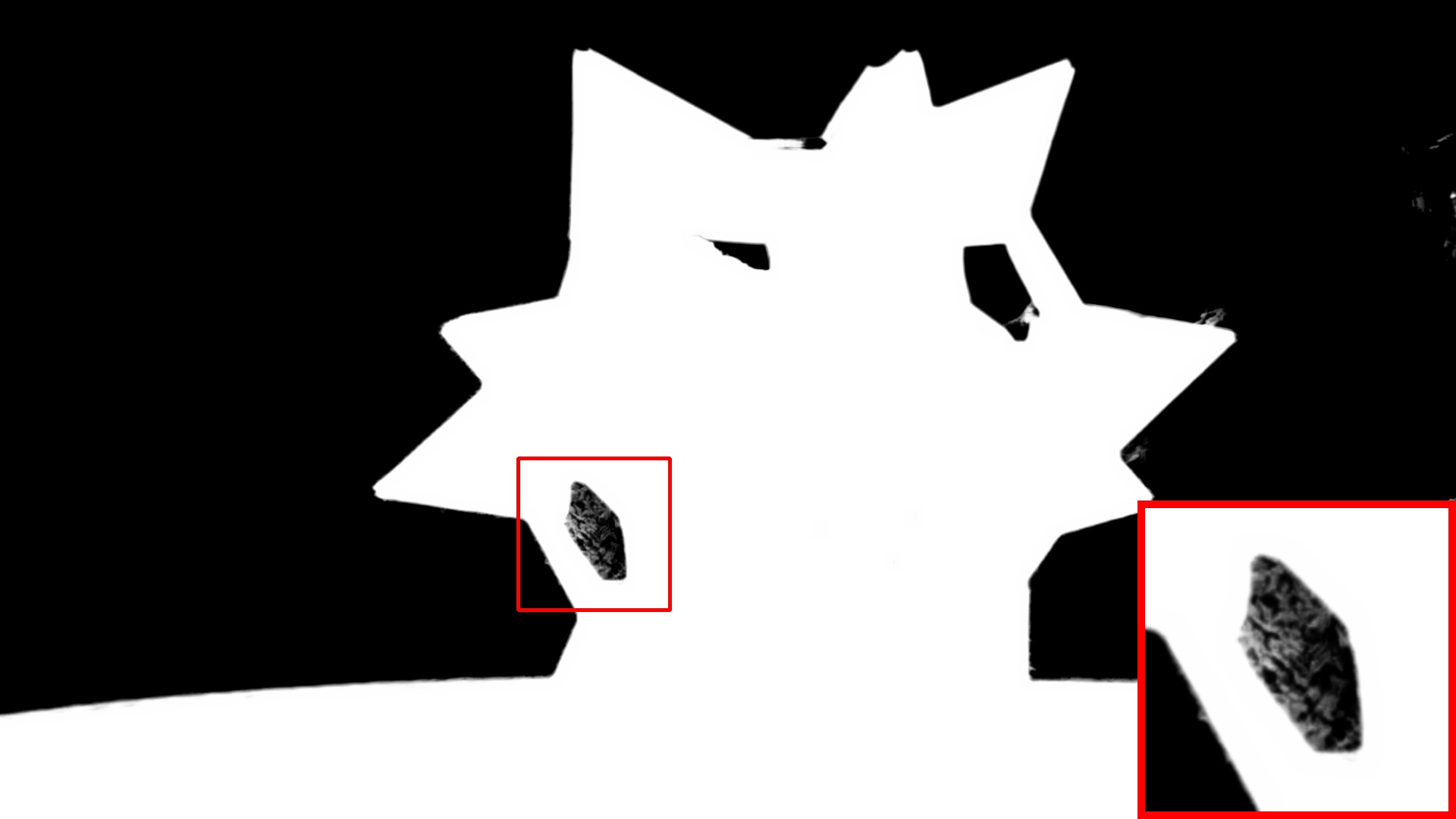} &
		\includegraphics[width=0.3\columnwidth]{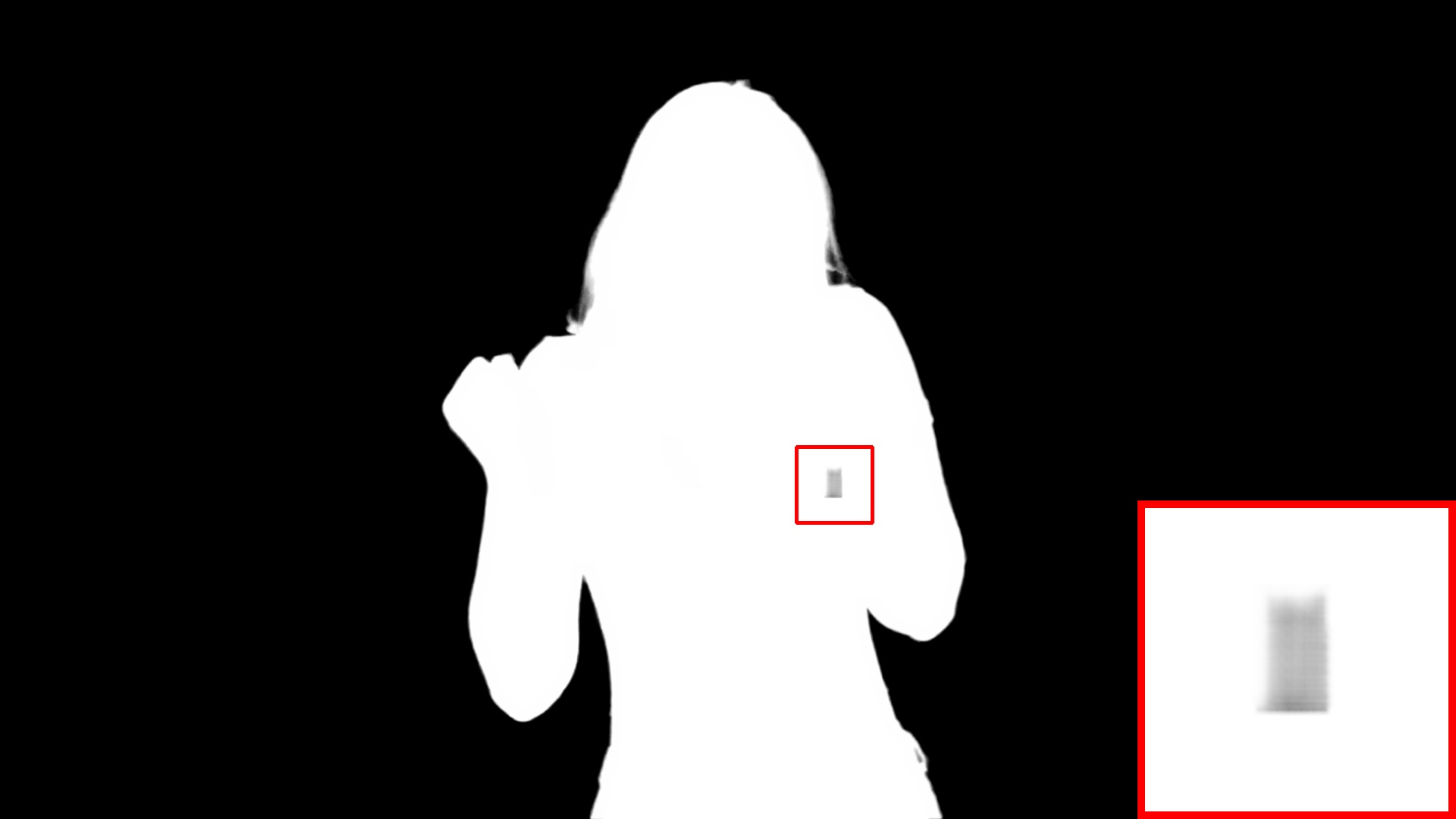} \\
        \rotatebox{90}{\scriptsize{\makecell[l]{Matte \\ STM+3FT}}} & 
        \includegraphics[width=0.3\columnwidth]{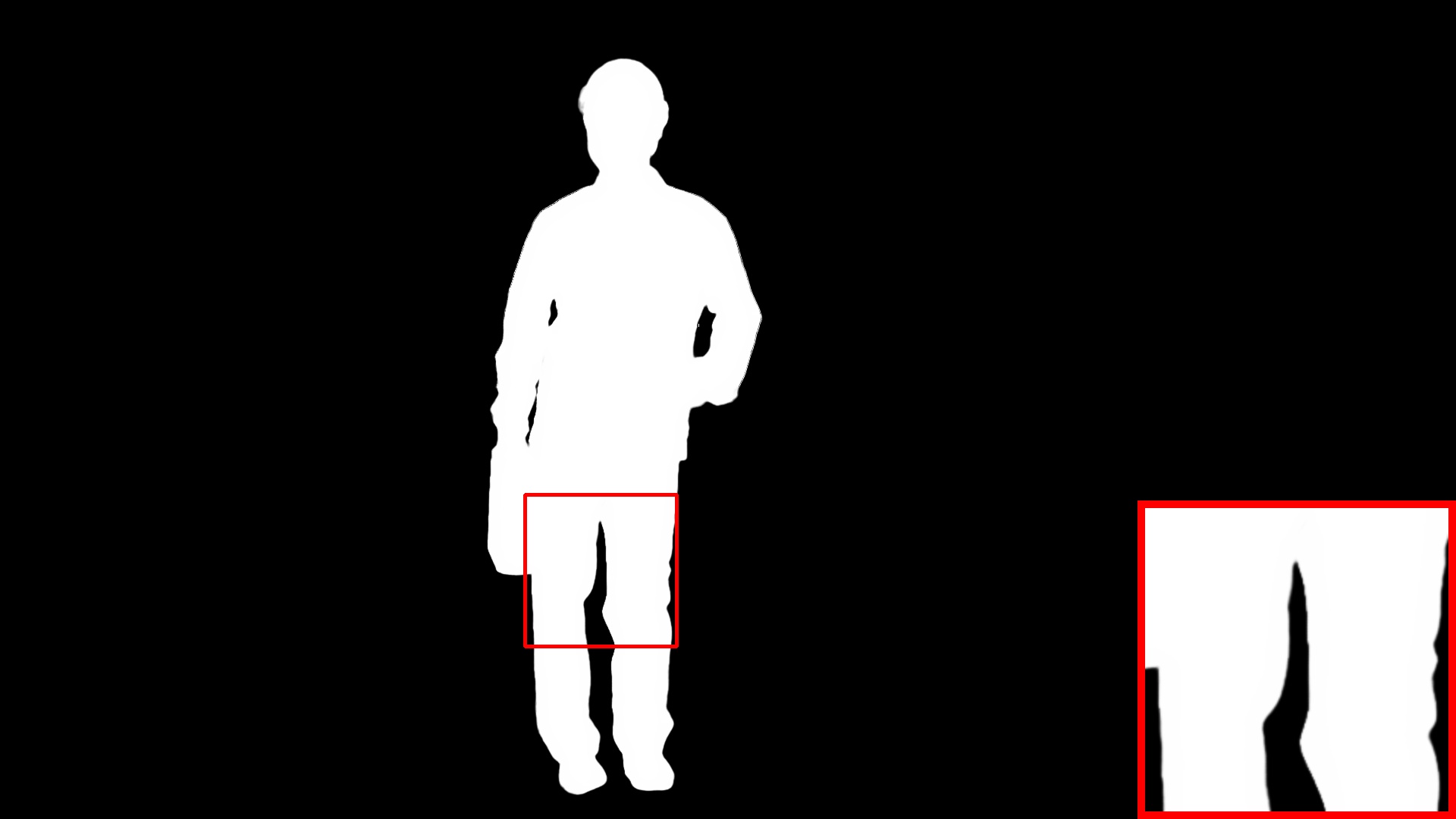} &
        \includegraphics[width=0.3\columnwidth]{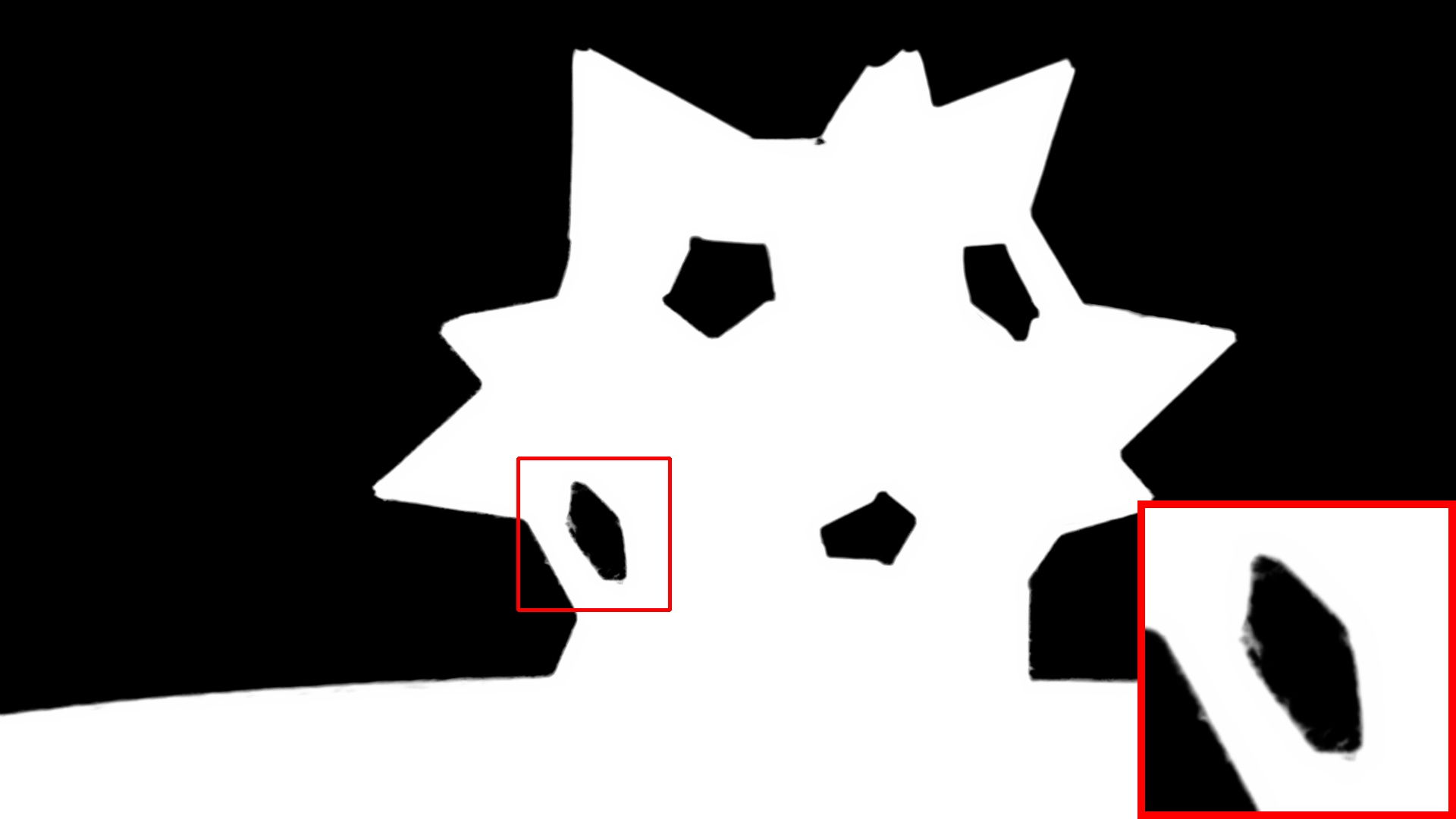} &
		\includegraphics[width=0.3\columnwidth]{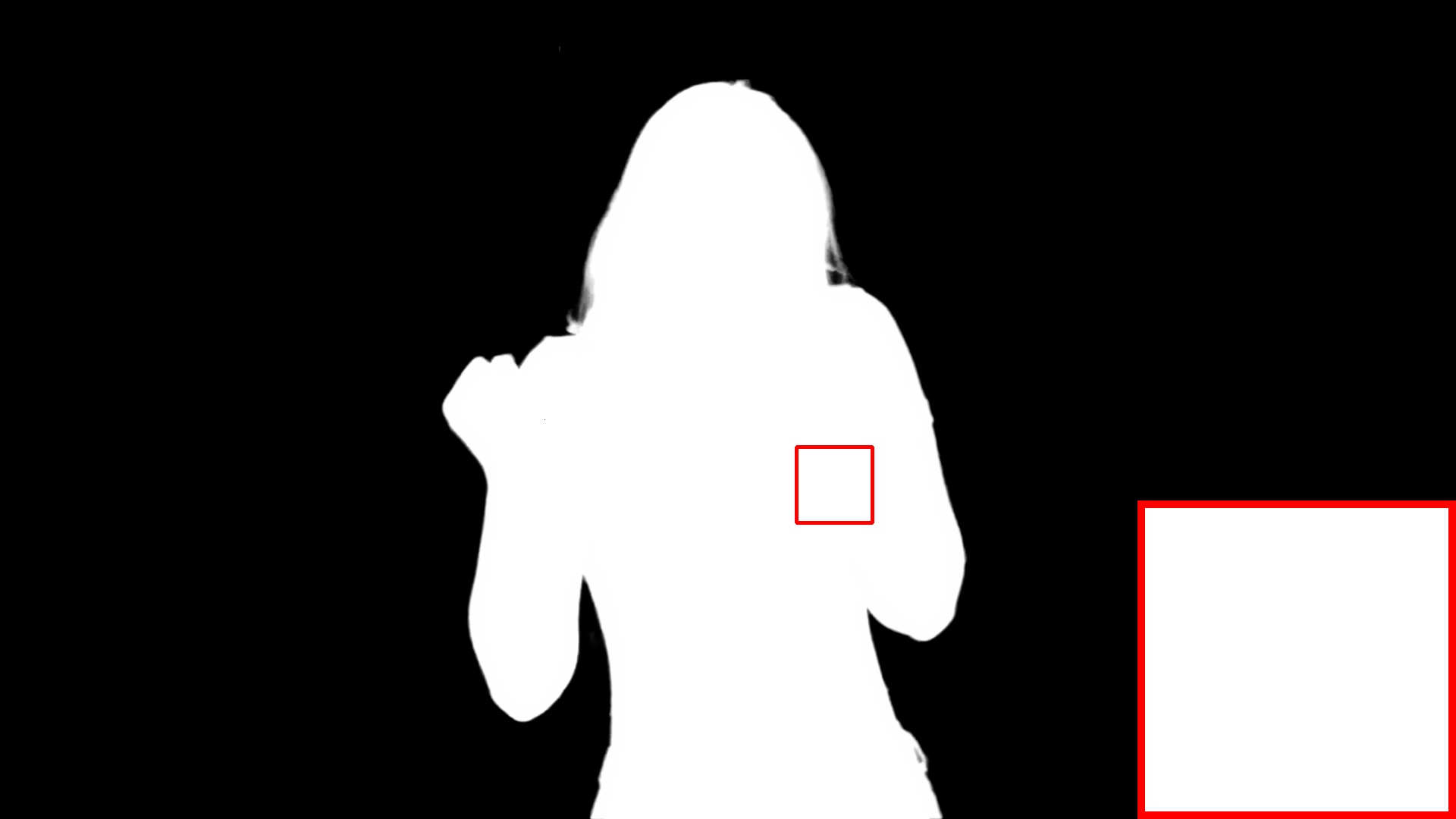} \\
		& (a) & (b) & (c)
        \end{tabular}
    }
\end{center}
\vspace{-10pt}
\caption{Qualitative evaluation of trimap generation. Red, blue and green corresponds to FR, BR and UR respectively. ``3FT'' denotes that we use three keyframes during fine-tuning.} 
\label{fig:abl_trimap}
\vspace{-15pt}
\end{figure}

\noindent \textbf{STM-based trimap generation}. We use the medium trimaps from VideoMatting108 as the ground truth to evaluate the performance of STM~\cite{oh2019video} on trimap generation using the mIoU metric. As the video sequences in VideoMatting108 are long, we only use the first 200 frames in this experiment. Specifically, we choose the first, the last, and the 100-th frame as keyframes during online fine-tuning. We also gradually add their corresponding ground-truth trimaps to show their impact on online fine-tuning. The input resolution is set to $768 \times 768$, and we fine-tune the network for 500 iterations. The average time for fine-tuning STM is 7.5 minutes on one GPU. As shown in Table~\ref{tab:abl_trimap}, online fine-tuning with the first or other frames improved the result by a large margin, especially in FR. Adding more ground-truth trimaps improves the results further. In conclusion, fine-tuning is necessary to adapt the network to user-specified keyframe trimaps, which improve the quality of generated trimaps. Table~\ref{tab:abl_trimap} also shows that adding more keyframes improves the mIoU score marginally, which indicates that a sparse set of annotated keyframe trimaps is enough for video trimap generation. We use around three frames in all our experiments.
On the other hand, inaccurate FR/BR heavily affects the matting result. As shown in the third row of Figure~\ref{fig:abl_trimap}(a), the pixels inside the gap between the two legs have erroneous high alpha values, despite the UR being roughly the same in the trimaps. In (b) and (c), the excessive URs lead to artifacts in the final alpha matte.

\noindent \textbf{Qualitative evaluations}. In Figure~\ref{fig:teaser} and Figure~\ref{fig:val_matting}, we show that our method could improve the temporal coherence comparing to single image matting networks. In the ``plant'' clip, GCA~\cite{li2020natural} produces flickering in the alpha matte although the foreground object is nearly static. As seen in the ``lion'' clip, GCA produces an erroneous white blob between the fur in the blowup. In contrast, our method could mitigate this discontinuity by aggregating temporal features. The same effect can be seen in the ``dancing woman'' clip and ``standing man'' clip when using Index~\cite{Lu_2019_ICCV} and DIM~\cite{xu2017deep} as the base network, respectively. All examples validate the effectiveness of our TAM.
In Figure~\ref{fig:test_matting}, we qualitatively compare GCA+TAM with GCA on three Internet video clips. The trimaps are generated by the fine-tuned STM~\cite{oh2019video}. In the ``fashion'' clip, our method can alleviate the artifacts in the gap between the model's arm and body. In the ``xmas-dog'' and the ``sofa'' clip, not only our method can produce more detailed result (\#9 of ``xmas-dog'' and \#92 of ``sofa''), it is also more robust to inaccurate URs in the trimap (\#22 of ``xmas-dog'' and \#23 of ``sofa''). The lengths of these three clips are 89, 100 and 93 frames respectively. More results can be found in the supplementary materials.

%% file: 05_discussion.tex

%% file: 06_conclusion.tex
\vspace{-5pt}
\section{Conclusion}
We have developed a deep video object matting method to achieve temporally coherent video matting results. Its key feature is an attention-based temporal aggregation module to compute the temporal affinity values in feature space, which are robust to appearance changes, fast motions, and occlusions. The temporal aggregation module can be easily integrated into image matting networks to enhance video object matting performance. We constructed a video matting dataset to enable the training of video object matting and trimap generation networks. This dataset has 80 training and 28 validation foreground video sequences with ground truth alpha mattes. 
In the future, we plan to investigate weakly supervised video object matting methods to reduce the workload of creating high-quality video matting training data.